\documentclass[twoside, 10pt]{article} 

\usepackage[preprint]{tmlr}

\usepackage[pdftex]{graphicx}

\DeclareUnicodeCharacter{0301}{\'{e}}

\usepackage{amsthm} 

\usepackage[
backend=biber,
style=numeric,
]{biblatex} 

\usepackage{float}
\usepackage{adjustbox}
\usepackage{multicol}
\usepackage{tcolorbox}
\usepackage{hyperref}
\hypersetup{
    colorlinks=true,
    linkcolor=blue,
    filecolor=blue,      
    urlcolor=blue,
    citecolor=black, 
    pdfpagemode=FullScreen,
    }

\usepackage{enumitem}

\usepackage{url}
\usepackage{booktabs, makecell, longtable}

\usepackage{amsmath}

\usepackage{adjustbox}
\usepackage{tikz}
\usetikzlibrary{arrows}
\usepackage{amsfonts}
\usepackage{bm}
\usepackage{amssymb}
\usepackage{braket}

\usepackage{tikz-cd}
\usepackage{wrapfig}
 \usepackage{xcolor}
\usepackage{xcolor,colortbl}
\usepackage{lipsum}

\usepackage{booktabs}       
\usepackage{amsfonts}       
\usepackage{nicefrac}       
\usepackage{microtype}      

\usepackage{blkarray}
\usepackage{multirow}
\usepackage{tikz-cd}
\usepackage{yfonts}

\usepackage{xspace}

\usepackage{cleveref}

\newtheorem{definition}{Definition}
\newtheorem{theorem}{Theorem}[section]

\newtheorem{proposition}[theorem]{Proposition}
\newtheorem{remark}[theorem]{Remark}
\newtheorem{example}[theorem]{Example}

\AtEveryBibitem{%
    \clearfield{url}
    \clearfield{isbn}
    \clearfield{issn}
    \clearfield{doi}
    \clearfield{note}
    \clearfield{eprint}
}

\DeclareRobustCommand\onedot{\futurelet\@let@token\@onedot}
\def\@onedot{\ifx\@let@token.\else.\null\fi\xspace}

\definecolor{buff}{rgb}{0.94, 0.86, 0.51}
\definecolor{burgundy}{rgb}{0.5, 0.0, 0.13}
\definecolor{iris}{rgb}{0.35, 0.31, 0.81}

\crefname{section}{section}{§§}
\crefname{section}{Section}{§§}
\crefname{thm}{Theorem}{Thm}
\crefname{eq}{Equation}{Eq}
\crefname{figure}{Fig.}{Figure}
\crefname{table}{Tab.}{Table}
\crefname{dfn}{Definition}{Dfn}

\def\eqref#1{equation~\ref{#1}}

\def\1{\bm{1}}

\DeclareMathOperator{\CC}{CC}
\newcommand{\CCHOP}[1]{\CC_{{#1}\text{-hop}}} 
\newcommand{\CCX}{\mathcal{X}}
\newcommand{\CCY}{\mathcal{Y}}
\newcommand{\CCZ}{\mathcal{Z}}

\newcommand{\Sym}{\mathrm{Sym}}

\newcommand{\N}{\mathbb{N}} 
\newcommand{\Znon}{\mathbb{Z}_{\ge 0}} 

\DeclareMathOperator{\HB}{HB}

\DeclareMathOperator{\proj}{Proj}

\DeclareMathOperator{\rk}{rk}

\DeclareMathOperator{\Int}{int}

\DeclareMathOperator{\CCN}{CCNN}

\begin{document}

\title{Topological Deep Learning: Going Beyond Graph Data}



\author{\name Mustafa Hajij \email mhajij@usfca.edu 
      \AND
      \name Ghada Zamzmi$^*$ \email ghadh@mail.usf.edu 
      \AND
      \name Theodore Papamarkou$^*$ \email theo.papamarkou@manchester.ac.uk 
      \AND
      \name Nina Miolane \email ninamiolane@ucsb.edu 
      \AND
      \name Aldo Guzm\'an-S\'aenz \email aldo.guzman.saenz@ibm.com 
      \AND
      \name Karthikeyan Natesan Ramamurthy \email knatesa@us.ibm.com 
     \AND
      \name Tolga Birdal \email tbirdal@imperial.ac.uk 
      \AND
      \name Tamal K. Dey \email tamaldey@purdue.edu 
      \AND
      \name Soham Mukherjee \email mukher26@purdue.edu 
      \AND
      \name Shreyas N. Samaga \email ssamaga@purdue.edu 
      \AND      
      \name Neal Livesay \email n.livesay@northeastern.edu 
        \AND
      \name Robin Walters \email r.walters@northeastern.edu 
      \AND      
      \name Paul Rosen \email prosen@sci.utah.edu 
      \AND
      \name Michael~T. Schaub \email schaub@cs.rwth-aachen.de 
}



\maketitle

\def\thefootnote{*}\footnotetext{These authors contributed equally to this work.}\def\thefootnote{\arabic{footnote}}

\tableofcontents

\newpage

\begin{abstract}
Topological deep learning is a rapidly growing field that pertains to the development of deep learning models for data supported on topological domains such as simplicial complexes, cell complexes, and hypergraphs, which generalize many domains encountered in scientific computations. In this paper, we present a unifying deep learning framework built upon a richer data structure that includes widely adopted topological domains.

Specifically, we first introduce \emph{combinatorial complexes}, a novel type of topological domain. Combinatorial complexes can be seen as generalizations of graphs that maintain certain desirable properties. Similar to hypergraphs, combinatorial complexes impose no constraints on the set of relations. In addition, combinatorial complexes permit the construction of hierarchical higher-order relations, analogous to those found in simplicial and cell complexes. Thus, combinatorial complexes generalize and combine useful traits of both hypergraphs and cell complexes, which have emerged as two promising abstractions that facilitate the generalization of graph neural networks to topological spaces.

Second, building upon combinatorial complexes and their rich combinatorial and algebraic structure, we develop a general class of message-passing \emph{combinatorial complex neural networks (CCNNs)}, focusing primarily on attention-based CCNNs. We characterize permutation and orientation equivariances of CCNNs, and discuss pooling and unpooling operations within CCNNs in detail.

Third, we evaluate the performance of CCNNs on tasks related to mesh shape analysis and graph learning. Our experiments demonstrate that CCNNs have competitive performance as compared to state-of-the-art deep learning models specifically tailored to the same tasks. Our findings demonstrate the advantages of incorporating higher-order relations into deep learning models in different applications.
\end{abstract}

\begin{figure}[!t]
\begin{center}
\includegraphics[scale = 0.99, keepaspectratio = 1]{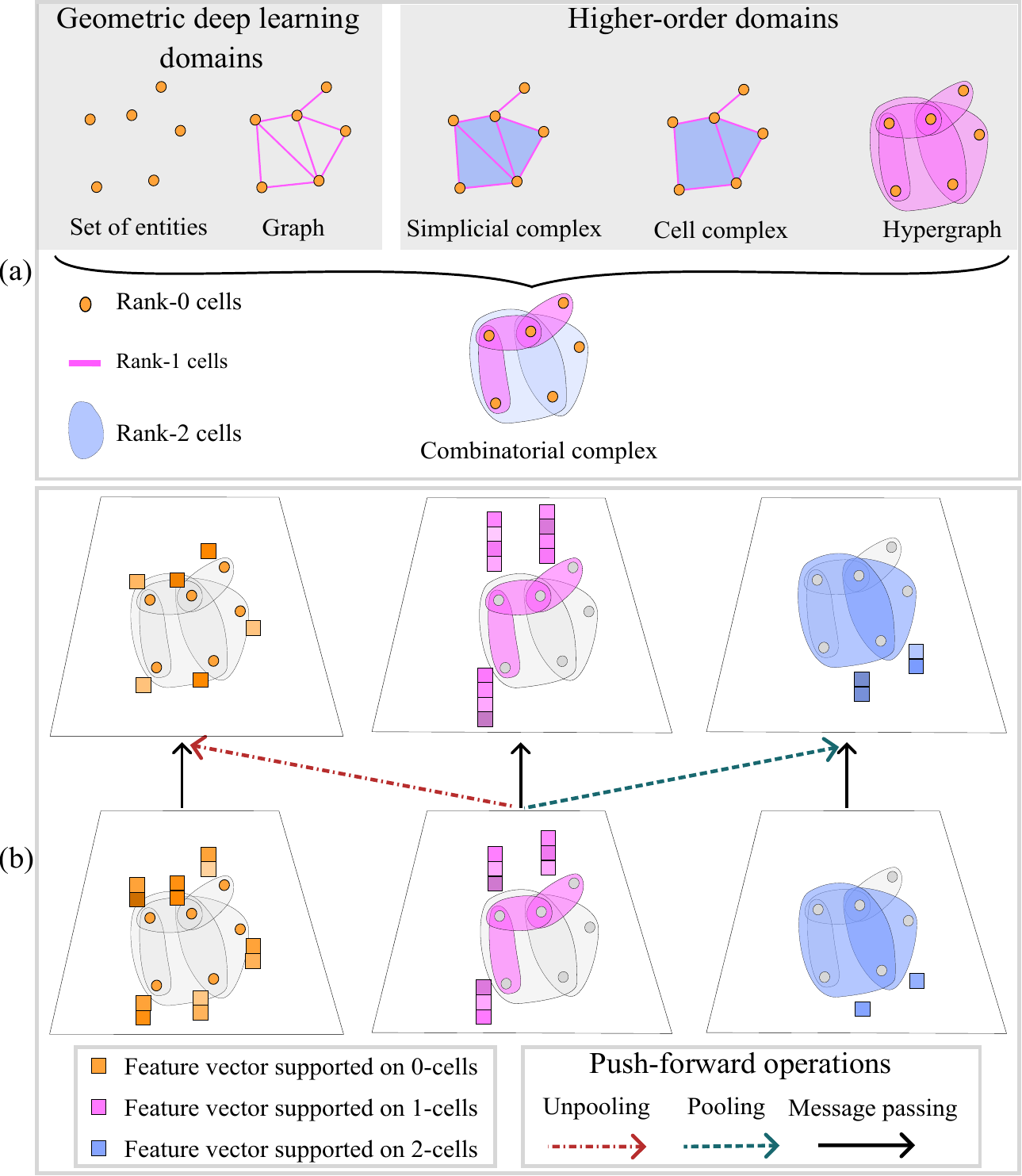}
\end{center}
\caption{A graphical abstract that visualizes our main contributions. (a): Different mathematical structures can be used to represent relations between abstract entities. Sets have entities with no connections, graphs encode binary relations between vertices, simplicial and cell complexes model hierarchical higher-order relations, and hypergraphs accommodate arbitrary set-type relations with no hierarchy.
We introduce combinatorial complexes (CCs), which generalize graphs, simplicial and cell complexes, and hypergraphs. CCs are equipped with set-type relations as well as with a hierarchy of these relations. (b): By utilizing the hierarchical and topological structure of CCs, we introduce the push-forward operation, a fundamental building block for higher-order message-passing protocols and for (un)pooling operations on CCs. Our push-forward operations on CCs enable us to construct combinatorial complex neural networks (CCNNs),
which provide a general conceptual framework for topological deep learning
on higher-order domains.\vspace{-3mm}}
\label{fig:main}
\end{figure}

\section{Introduction}
\label{sec:intro}

In recent years, there has been an exponential growth in the amount of data available for computational analysis, including scientific data as well as common data types such as text, images, and audio. This abundance of data has empowered various fields, including physics, chemistry, computational social sciences, and biology, to make significant progress using machine learning techniques, primarily deep neural networks. As deep neural networks can effectively summarize and extract patterns from large data sets, they are suitable for many complex tasks. Initially, deep neural networks have been developed to learn from data supported on regular (Euclidean) domains,
such as grids in images, sequences of text and time-series. These models, including convolutional neural networks (CNNs)~\cite{krizhevsky2012imagenet,lecun1998gradient,simonyan2014very}, recurrent neural networks (RNNs)~\cite{sutskever2014sequence,bahdanau2014neural} and transformers \cite{vaswani2017attention}, have proven highly effective in processing such Euclidean data~\cite{goodfellow2016deep}, resulting in unprecedented performance in various applications, most recently in chat-bots (e.g., ChatGPT~\cite{adesso2022gpt4}) and text-controlled image synthesis~\cite{rombach2022high}.


However, scientific data in various fields are often structured differently and are not supported on regular Euclidean domains. As a result, adapting deep neural networks to process this type of data has been a challenge. Against this backdrop, geometric deep learning (GDL)~\cite{bronstein2021geometric,zhou2020graph,wu2020comprehensive} has emerged as an extension of deep learning models to non-Euclidean domains. To achieve this, GDL restricts the performed computations via principles of geometric regularity, such as symmetries, invariances, and equivariances. The GDL perspective allows for appropriate inductive biases to be imposed when working with arbitrary data domains, including sets~\cite{qi2017pointnet,rempe2020caspr,deng2018ppfnet,zhao20223dpointcaps++,huang2022multiway}, grids~\cite{boscaini2015learning,masci2015geodesic,boscaini2016learning,kokkinos2012intrinsic,shuman2016vertex,wu20153d,monti2017geometric}, manifolds~\cite{boscaini2015learning,masci2015geodesic,boscaini2016learning,kokkinos2012intrinsic,shuman2016vertex,wu20153d,monti2017geometric}, and graphs~\cite{scarselli2008graph,gallicchio2010graph,zhou2020graph,wu2020comprehensive,boscaini2016learning,monti2017geometric,bronstein2017geometric,kipf2016semi}. 
Graphs, in particular, have garnered interest due to their applicability in numerous scientific studies and their ability to generalize conventional grids. Accordingly, the development of graph neural networks (GNNs)~\cite{bronstein2017geometric,kipf2016semi} has remarkably enhanced our ability to model and analyze several types of data in which graphs naturally appear.

Despite the success of GDL and GNNs, seeing graphs through a purely geometric viewpoint yields a solely local abstraction and falls short of capturing non-local properties and dependencies in data. \emph{Topological data}, including interactions of edges (in graphs), triangles (in meshes) or cliques, arise naturally in an array of novel applications in complex physical systems~\cite{battiston2021physics,lambiotte2019networks}, traffic forecasting~\cite{jiang2022graph}, social influence~\cite{zhu2018social}, protein interaction~\cite{murgas2022hypergraph}, molecular design~\cite{schiff2020characterizing}, visual enhancement~\cite{efthymiou2021graph}, recommendation systems \cite{la2022music}, and epidemiology~\cite{deng2020cola}.
To natively and effectively model such data, we are bound to go beyond graphs and consider qualitative spatial properties remaining unchanged under some geometric transformations. In other words, we need to consider the \emph{topology of data}~\cite{carlsson2009topology} to formulate neural network architectures capable of extracting semantic meaning from complex data.

One approach to extract more global information from data is to go beyond graph-based abstractions and consider extensions of graphs, such as simplicial complexes, cell complexes, and hypergraphs, generalizing most data domains encountered in scientific computations~\cite{bick2021higher,battiston2020networks,benson2021higher,torres2021and}.
The development of machine learning models to learn from data supported on these topological domains~\cite{feng2019hypergraph,bunch2020simplicial, roddenberry2021signal,schaub2020random,billings2019simplex2vec,hacker2020k,hajijcell,ebli2020simplicial,schaub2021signal,roddenberry2021principled,giusti2022cell,yang2023convolutional} is a rapidly growing new frontier, to which we refer hereafter as \emph{topological deep learning (TDL)}.
TDL intertwines several research areas, including topological data analysis (TDA)~\cite{edelsbrunner2010computational,carlsson2009topology,DW22,love2020topological,ghrist2014elementary}, topological signal processing~\cite{schaub2018denoising,yang2021finite,schaub2022signal,roddenberry2021signal,barbarossa2020topological,robinson2014topological,sardellitti2022topological}, network science~\cite{skardal2021higher,lambiotte2019networks,barabasi2013network,battiston2020networks,bick2021higher,bianconi2021higher,benson2016higher,de2016physics,bao2022impact,oballe2021bayesian}, and geometric deep learning~\cite{zhang2020deep,cao2020comprehensive,fey2019fast,loukas2019graph,battaglia2018relational,morris2019weisfeiler,battaglia2016interaction}.


Despite the growing interest in TDL, a broader synthesis of the foundational principles of these ideas has not been established so far. We argue that this is a deficiency that inhibits progress in TDL, as it makes it challenging to draw connections between different concepts, impedes comparisons, and makes it difficult for researchers of other fields to find an entry point to TDL. Hence, in this paper, we aim to provide a foundational overview over the principles of TDL, serving not only as a unifying framework for the many exciting ideas that have emerged in the literature over recent years, but also as a conceptual starting point to promote the exploration of new ideas. Ultimately, we hope that this work will contribute to the accelerated growth in TDL, which we believe would be a key enabler of transferring deep learning successes to an enlarged range of application scenarios.



By drawing inspiration from traditional topological notions in algebraic topology~\cite{ghrist2014elementary,hatcher2005algebraic} and recent advancements in higher-order networks~\cite{battiston2020networks,torres2021and,bick2021higher,battiston2021physics}, we first introduce \emph{combinatorial complexes (CCs)} as the major building blocks of our TDL framework. CCs constitute a novel topological domain that unifies graphs, simplicial and cell complexes, and hypergraphs as special cases, as illustrated in Figure~\ref{fig:main}\footnote{All figures in this paper should be displayed in color, as different colors communicate different pieces of information.}. 
Similar to hypergraphs, CCs can encode arbitrary set-like relations between collections of abstract entities. Moreover, CCs permit the construction of hierarchical higher-order relations, analogous to those found in simplicial and cell complexes. Hence, CCs generalize and combine the desirable traits of hypergraphs and cell complexes.

In addition, we introduce the necessary operators to construct deep neural networks for learning features and abstract summaries of input anchored on CCs. Such operators provide convolutions, attention mechanisms, message-passing schemes as well as the means for incorporating invariances, equivariances, or other geometric regularities. Specifically, our novel \emph{push-forward operation} allows pushing data across different dimensions, thus forming an elementary building block upon which \emph{higher-order message-passing protocols} and \emph{(un)pooling operations} are defined on CCs. The resulting learning machines, which we call \emph{combinatorial complex neural networks (CCNNs)}, are capable of learning abstract higher-order data structures, as clearly demonstrated in our experimental evaluation.

We envisage our contributions to be a platform encouraging researchers and practitioners to extend our $\CCN$s, and invite the community to build upon our work to expand TDL on higher-order domains.
Our contributions, which are also visually summarized in Figure~\ref{fig:main}, are the following:


\begin{itemize}[topsep=1pt]
\setlength\itemsep{0.1pt}
\item First, we introduce CCs as domains for TDL. We characterize CCs and their properties, and explain how
they generalize major existing domains, such as graphs, hypergraphs, simplicial and cell complexes. CCs can thus serve as a unifying starting point that enables the learning of expressive representations of topological data. 
\item Second, using CCs as domains, we construct $\CCN$s, an abstract class of higher-order message-passing neural networks that provide a unifying blueprint for TDL models based on hypergraphs and cell complexes. 
\begin{itemize}[topsep=1pt]
\setlength\itemsep{0.1pt}
    \item Based upon a push-forward operator defined on CCs, we introduce convolution, attention, pooling and unpooling operators for $\CCN$s. 
    \item We formalize and investigate \emph{permutation and orientation equivariance} of $\CCN$s, paving the way to future work on geometrization of $\CCN$s.
    \item We show how CCNNs can be intuitively constructed via graphical notation.
\end{itemize}
\item Third, we evaluate our ideas in practical scenarios.
\begin{itemize}[topsep=1pt,noitemsep]
    \item We release the source code of our framework as three supporting Python libraries: \emph{TopoNetX}, \emph{TopoEmbedX} and \emph{TopoModelX}.
    \item We show that $\CCN$s attain competing predictive performance against state-of-the-art task-specific neural networks in various applications, including shape analysis and graph learning. 
    \item We establish connections between our work and classical constructions in TDA, such as the \emph{mapper}~\cite{singh2007topological}. Particularly, we realize the mapper construction in terms of our TDL framework and demonstrate how it can be utilized in higher-order (un)pooling on CCs.
    \item  We demonstrate the reducibility of any CC to a special graph called the \emph{Hasse graph}. This enables the characterization of certain aspects of $\CCN$s in terms of graph-based models, allowing us to reduce higher-order representation learning to graph representation learning (using an enlarged computational graph).
\end{itemize}
\end{itemize}

\paragraph{Glossary.} Prior to delving into more details, we present the elementary terminologies of already established concepts used throughout the paper. Some of these terms are revisited formally in Section~\ref{prelim}. Appendix~\ref{appendix_glossary} provides notation and terminology glossaries for novel ideas put forward by the paper.




\vspace{-0.5mm}
\begin{tcolorbox}
\textbf{\href{https://app.vectary.com/p/3EBiRiJcYjFNvkbbWszQ0Z}{Cell complex}}\label{SX.term}: a topological space obtained as a disjoint union of topological disks (cells), with each of these cells being homeomorphic to the interior of a Euclidean ball. These cells are attached together via attaching maps in a locally reasonable manner. \\[1mm]
\textbf{Domain}\label{domain.term}: the underlying space where data is typically supported.\\[1mm]
\textbf{Entity} or \textbf{vertex}\label{entity.term}: an abstract point; it can be thought as an element of a set.\\[1mm]
\textbf{Graph} or \textbf{network}\label{G.term}: a set of entities (vertices) together with a set of edges representing binary relations between the vertices.\\[1mm]
\textbf{Hierarchical structure} on a topological domain: an integer-valued function that assigns a positive integer (rank) to every relation in the domain such that higher-order relations are assigned higher rank values. For instance, a simplicial complex admits a hierarchical structure induced by the cardinality of its simplices.\\[1mm]
\textbf{Higher-order network/topological domain}: generalization of a graph that captures (binary and) higher-order relations between entities. Simplicial/cell complexes and hypergraphs
are examples of higher-order networks.\\[1mm]
\textbf{Hypergraph}: a set of entities (vertices) and a set of hyperedges representing binary or higher-order relations between the vertices.\\[1mm]
\textbf{Message passing}: a computational framework that involves passing data, `messages', among neighbor entities in a domain to update the representation of each entity based on the messages received from its neighbors.\\[1mm]
\textbf{Relation} or \textbf{cell}\label{cell.term}: a subset of a set of entities (vertices).
A relation is called \textbf{binary} if its cardinality is equal to two.
A relation is called \textbf{higher-order} if its cardinality is greater than two.\\[1mm]
\textbf{Set-type relation}: a higher-order network is said to have set-type relations if the existence of a relation is not implied by another relation in the network; e.g., hypergraphs admit set-type relations.\\[1mm]  
\textbf{Simplex}: a generalization of a triangle or tetrahedron to arbitrary dimensions; e.g.,
a simplex of dimension zero, one, two, or three is a point, line segment, triangle, or tetrahedron, respectively.\\[1mm]
\textbf{\href{https://app.vectary.com/p/4HZRioKH7lZ2jWESIBrjhf}{Simplicial complex}}\label{SC.term}: a collection of simplices such that every face of each simplex in the collection is also in the collection, and the intersection of any two simplices in the collection is either empty or a face of both simplices.\\[1mm]
\textbf{Topological data}\label{TD.term}: feature vectors supported on relations in a topological domain.\\[1mm] 
\textbf{Topological deep learning}\label{TDL.term}: the study of topological domains using deep learning techniques, and the use of topological domains to represent data in deep learning models. 
\end{tcolorbox}




\begin{figure}[!t]
\begin{center}
\includegraphics[scale = 0.14, keepaspectratio = 1]{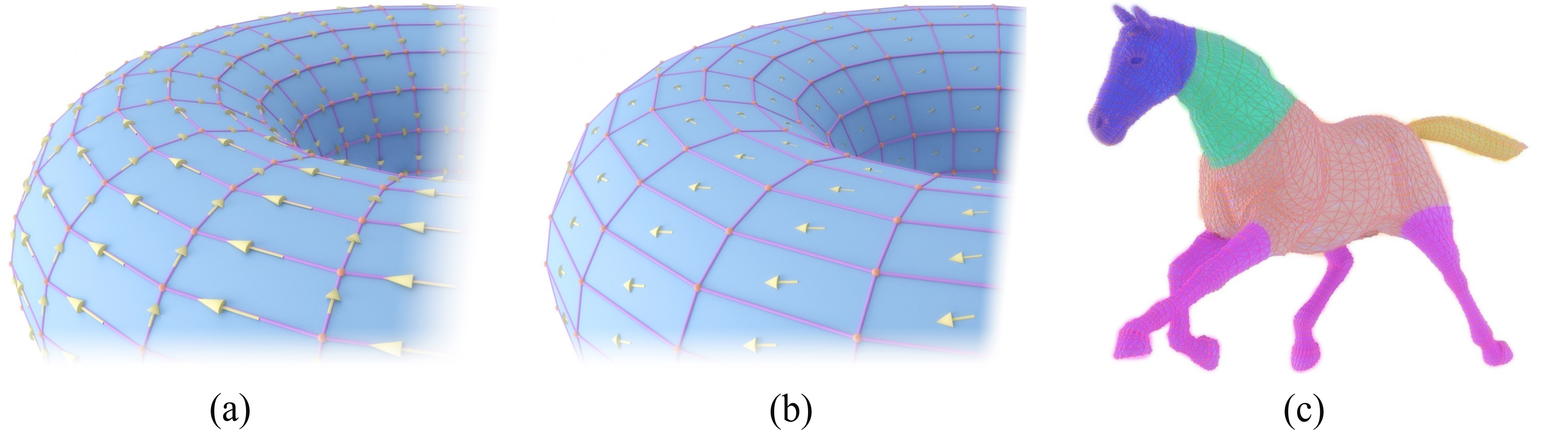}
\end{center}
\caption{Data might be supported naturally on higher-order relations.
(a): An edge-based vector field. (b): A face-based vector field.
Both vector fields in (a) and (b) are defined on a cell complex torus. An interactive visualization of (a--b) is provided \href{https://app.vectary.com/p/5uLAflZj6U2kvACv2kk2tN}{here}.
(c): Class-labeled topological data might naturally be supported on higher-order relations. For instance, mesh segmentation labels for 2-faces are depicted by different colors (blue, green, turquoise, pink, brown) to
represent different parts (head, neck, body, legs, tail) of the horse.\vspace{-3mm}}
\label{fig:support}
\end{figure}

\section{Motivation: what does topology offer to machine learning?}
\label{sec:motivation}

Combinatorial complex neural networks (CCNNs), as presented in this work, generalize graph neural networks (GNNs) and their higher-order analogues via topological constructions.
In this section, we provide the motivation for the use of topological constructions in machine learning, and specifically in deep learning from three angles. 
First, \textit{data modeling}: how topological abstraction can help us to reason and compute with various data types supported on topological spaces.
Second, \textit{utility}: how accounting for topology improves performance. Third, \textit{unification}: how topology can be used to synthesize and abstract disparate concepts.

\subsection{Modeling and learning from data on topological spaces}

In the context of machine learning, the domain on which the data is supported is typically a (linear) vector space.
This underlying vector space facilitates many computations and is often implicitly assumed.
However, as has been recognized within geometric deep learning, it is often essential to consider data supported on different domains.
Indeed, explicitly considering and modeling the domain on which the data is supported can be crucial for various reasons.

First, different domains can have distinct characteristics and properties that require different types of deep learning models to effectively process and analyze the data. 
For example, a graph domain may require a graph convolutional network~\cite{kipf2016semi} to account for the non-Euclidean structure, while a point cloud domain may require a PointNet-like architecture~\cite{qi2017pointnet,zaheer2017deep} to handle unordered sets of points.

Second, the specific data within each domain can vary widely in terms of size, complexity and noise. By understanding specific data properties within a given domain, researchers can develop models that are tailored to those particular properties. For example, a model designed for point clouds with high levels of noise may incorporate techniques such as outlier removal or local feature aggregation.

Third, accurately distinguishing between domain and data is important for developing models that generalize well to new, unseen data. By identifying the underlying structure of the domain, researchers can develop models that are robust to variations in the data while still being able to capture relevant geometric features.
To summarize, by considering both the domain and data together, researchers can develop models that are better suited to handle the specific challenges and complexities of different geometric learning tasks.

\paragraph{Different perspectives on data: domains and relations.}

When referring to topological data or topological data analysis in the literature, there are different views on what the actual data consists of and what it is supposed to model. Here, we follow~\cite{bick2021higher} and distinguish between different types of data and goals of our learning procedures, even though the distinction between these different types of data can sometimes be more blurry than what is presented here.

\begin{figure}[!t]
\begin{center}
\includegraphics[scale = 0.25, keepaspectratio = 1]{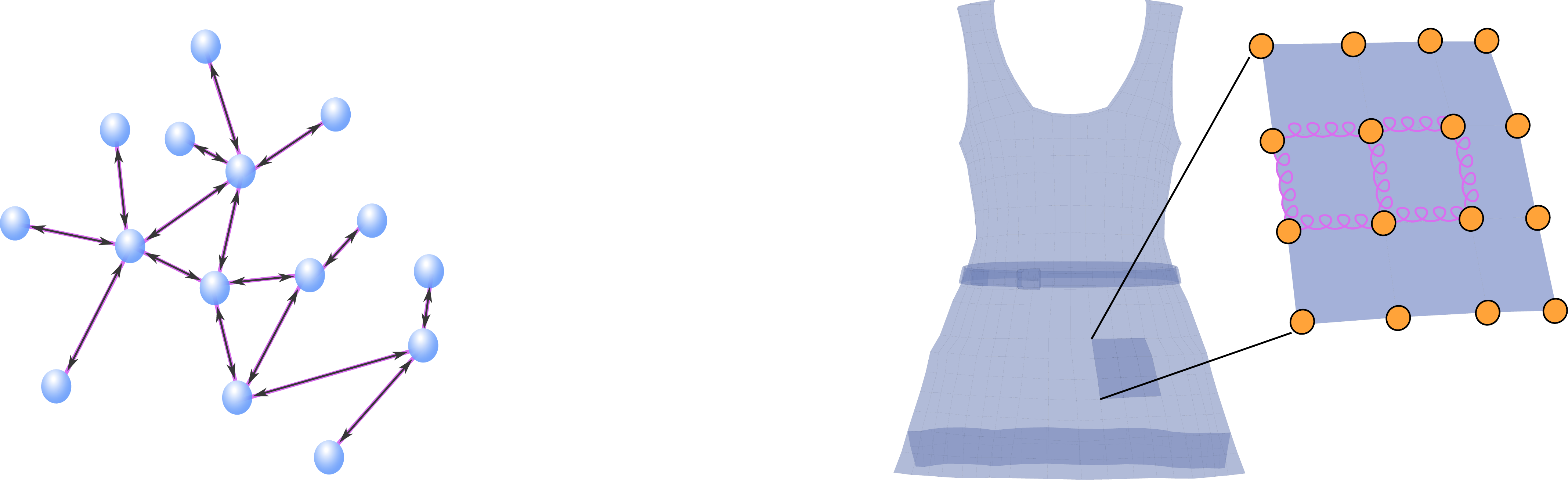}
\put(-280,-27){(a)}
\put(-100,-27){(b)}
\end{center}
\caption{Examples of processing data supported on graphs or on higher-order networks.
(a): Graphs can be used to model particle interactions in fluid dynamics.
Vertices represent particles,
whereas particle-to-particle interactions are modeled
via message passing among vertices~\cite{sanchez2020learning,shlomi2020graph}.
(b): When modeling springs and self-collision, it is natural to work with edges rather than vertices. This is because the behavior of the cloth is determined by the tension and compression forces acting along the edges, and not only by the position of individual particles. To model the interactions among multiple edges, polygonal faces can be used to represent the local geometry of the cloth. The polygonal faces provide a way to compute higher-order message passing among the edges.\vspace{-5mm}}
\label{fig:cloth}
\end{figure}
\emph{Relational data.}
Relational data describes the relations between different entities or objects. 
This fundamental concept can manifest in various ways, such as connections between users in social networks, where friend relationships or follower-ship serve as examples of such relations. 
Traditionally, these relations are understood through graphs, with vertices and edges representing entities and their \emph{pairwise connections}, respectively. Stated differently, the edges within the graph are considered to be units of measurement; i.e., every piece of relational data involves two vertices.

However, many real-world phenomena naturally involve complex, multi-party interactions, where more than two entities interact with each other in intricate ways. For instance, groups of individuals within a society, sets of co-authors, and genes or proteins that interact with each other are all examples of such higher-order relations. Such dependencies, which go beyond pairwise interactions, can be modeled by \emph{higher-order relations} and aptly abstracted as hypergraphs, cell complexes, or CCs. Clearly, topological ideas can play an important role for understanding this data, and crucially enable us to move from graphs modeling pairwise relations to richer representations.

\emph{Data as an instance of a domain.}
A related conceptualization of what our observed data is supposed to represent is adopted in TDA.
Namely, all of observed data is supposed to correspond to a noisy instance of a topological object itself; i.e., we aim to learn the `topological shape' of the data.
Note that in this context, we typically do not directly observe relational data, but instead we \emph{construct relations from observed data}, which we then use to classify the observed dataset.
Persistent homology performed on point cloud data is a mainstream example for this viewpoint.

\emph{Data supported on a topological domain.}
Topological ideas also play an important role when considering data that is defined on top of a domain such as a graph.
This could be particular dynamics, such as epidemic spreading, or it could be any type of other data or signals supported on the vertices (cases of edge-signals on a graph are not considered often, though exceptions exist~\cite{schaub2018denoising,schaub2021signal}).
In the language of \emph{graph signal processing}, a variety of functions or signals can be defined on a graph \emph{domain}, and these are said to be \emph{supported by the domain}. Crucially, the graph itself can be arbitrarily complex but is typically considered to be fixed; the observed data is not relational, but supported on the vertices of a graph.

Moving beyond graphs, more general topological constructions, such as CCs, can support data not just on vertices and edges, but also on other `higher-order' entities, as demonstrated in Figure~\ref{fig:support}(a--b). For example, vector fields defined on meshes in computer graphics are often data supported on edges and faces, and can be conveniently modeled as data supported on a higher-order domain~\cite{de2016vector}. 
Similarly, class-labeled data can be provided on the edges and faces of a given mesh; see Figure~\ref{fig:support}(c) for an example.
To process such data it is again relevant to take into account the structure of the underlying topological domain.

\paragraph{Modeling and processing data beyond graphs: illustrative examples.} 
Graphs are well-suited for modeling systems that exhibit pairwise interactions. 
For instance, particle-based fluid simulations can be effectively represented using graphs as shown in Figure~\ref{fig:cloth}(a), with message passing used to update the physical state of fluid molecules. In this approach, each molecule is represented as a vertex containing its physical state, and the edges connecting them are utilized to calculate their interactions~\cite{sanchez2020learning,shlomi2020graph}.

Graphs may not be adequate for modeling more complex systems such as cloth simulation, where state variables are associated with relations like edges or triangular faces rather than vertices. 
In such cases, higher-order message passing is required to compute and update physical states, as depicted in Figure~\ref{fig:cloth}(b). 
A similar challenge arises in natural language processing, where language exhibits multiple layers of syntax, semantics and context. 
Although GNNs may capture basic syntactic and semantic relations between words, more complex relations such as negation, irony or sarcasm may be difficult to represent~\cite{girault2017towards}. 
Including higher-order and hierarchical relations can provide a better model for these relations, and can enable a more nuanced and accurate understanding of language.


\begin{figure}[!t]
\begin{center}
\includegraphics[scale = 0.22, keepaspectratio = 0.25]{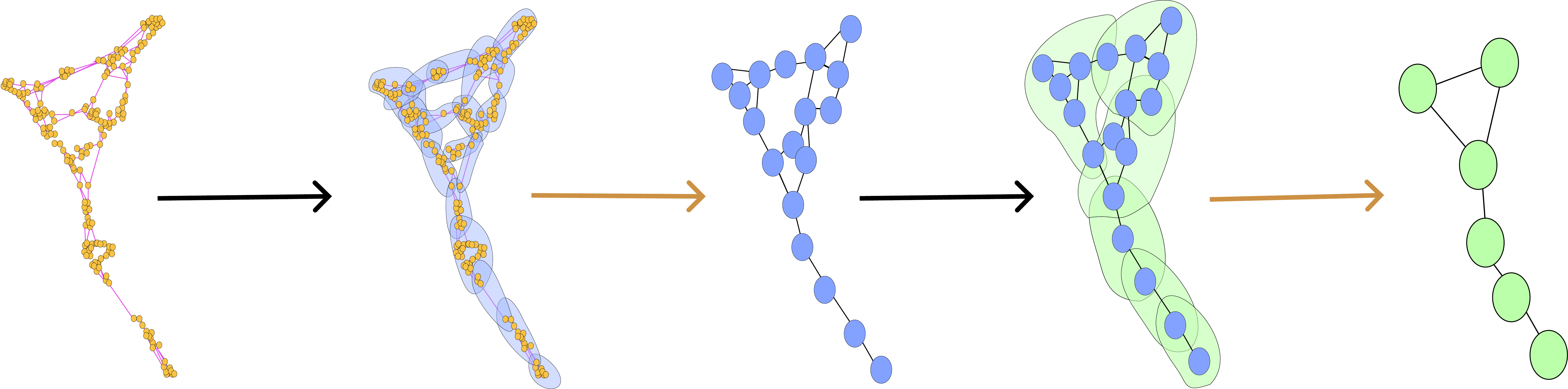}
\put(-399,-16){(a)}
\put(-290,-16){(b)}
\put(-210,-16){(c)}
\put(-110,-16){(d)}
\put(-20,-16){(e)}
\end{center}
\caption{An illustration of representing hierarchical data via higher-order networks.
Black arrows indicate graph augmentation by higher-order relations,
whereas orange arrows indicate coarsened graph extraction.
(a): A graph encodes binary relations (pink edges) among abstract entities (yellow vertices).
(b): Higher-order relations, represented by blue cells,
can be thought as relations among vertices or edges of the original graph.
(c): Extraction of a coarsened version of the original graph.
In the coarsened graph,
vertices represent the higher-order relations (blue cells) of the original graph,
and edges represent the intersections of these blue cells.
(d--e): The same process repeats to obtain a more coarsened version of the original graph.
The entire process corresponds to hierarchical higher-order relations,
that is relations among relations,
which extract meaning and content (including the `shape of data'),
a common task in topological data analysis~\cite{carlsson2009topology,DW22}.}
\label{fig:hs}
\end{figure}
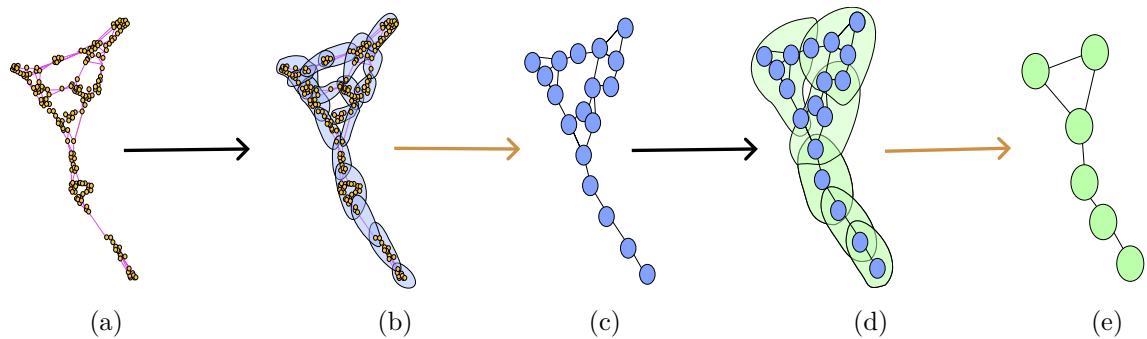    

\subsection{The utility of topology}
\label{subsec:utility}

In addition to providing a versatile framework for modeling complex systems, TDL models have broad utility: they can induce meaningful inductive biases for learning, facilitate efficient message propagation over longer ranges in the underlying domain,  enhance the expressive power of existing graph-based models, and have the potential to unveil crucial characteristics of deep networks themselves, as described next. 

\paragraph{Building hierarchical representations of data.} 
While leveraging higher-order information is important for learning representations, it is also crucial to preserve the complex higher-order relations between entities for achieving powerful and versatile representations. For instance, humans can form abstractions and analogies by building relations among relations hierarchically.
However, graph-based models, which are commonly used for building relational reasoning among various entities~\cite{santoro2017simple,zhang2020deep,chen2019graph,schlichtkrull2018modeling}, are limited in their ability to build higher-order relations among these entities.

The need for hierarchical relational reasoning~\cite{xu2022groupnet} demands methods that can capture deeper abstract relations among relations, beyond modeling relations between raw entities. 
Higher-order network models provide a promising solution for addressing the challenge of higher-order relational reasoning, as depicted in Figure~\ref{fig:hs}.

\begin{figure}[!t]
\begin{center}
\includegraphics[scale = 0.08, keepaspectratio = 1]{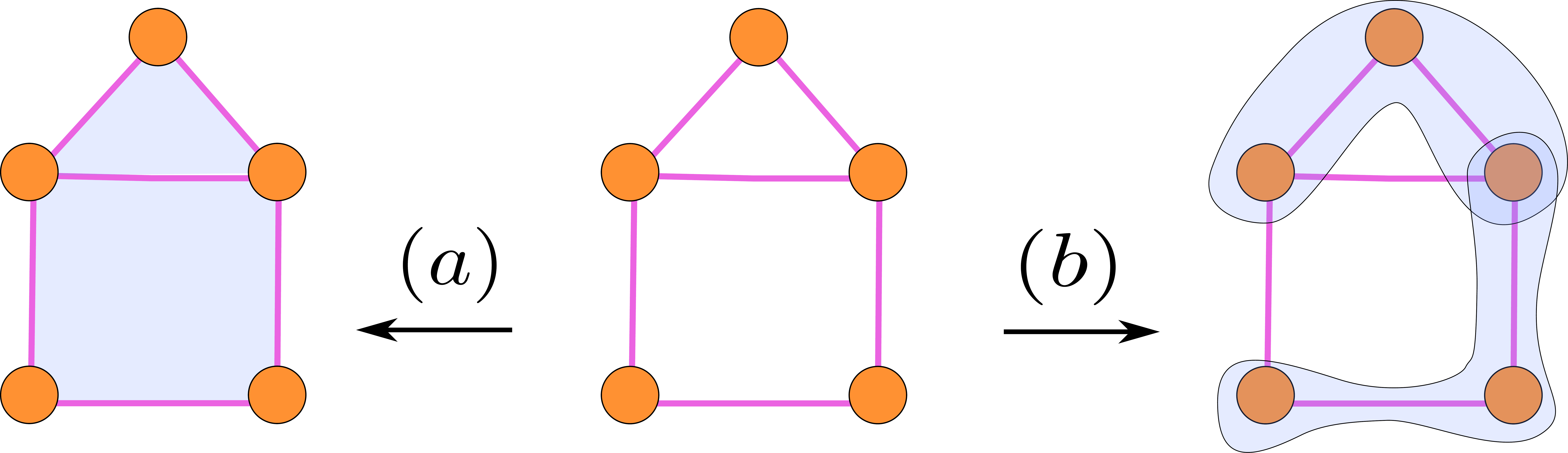}
\end{center}
\caption{The graph in the middle can be augmented with higher-order relations to improve a learning task. In (a), cells have been added to the missing faces; a similar inductive bias has been considered in~\cite{bodnar2021weisfeiler} to improve classification on molecular data. In (b), one-hop neighborhoods have been added to some of the vertices in the graph;~\cite{feng2019hypergraph} use such an inductive bias, based on lifting a graph to its corresponding one-hop neighborhood hypergraph, to improve performance on vertex-based tasks.\vspace{-2mm}}
\label{fig:inductive bias}
\end{figure}

\paragraph{Improving performance through inductive bias.}
A topological deep learning model can be leveraged to enhance the predictive performance on graph-based learning tasks by providing a well-defined procedure to lift a graph to a given higher-order network. The incorporation of multi-vertex relations via this approach can be viewed as an inductive bias, and is utilized to improve predictive performance. Inductive bias allows a learning algorithm to prioritize one solution over another, based on factors beyond the observed data~\cite{mitchell1980need}. 
Figure~\ref{fig:inductive bias} illustrates two forms of augmentation on graphs using higher-order cells.
\vspace{-2mm}

\begin{figure}[!t]
\begin{center}
\includegraphics[scale = 0.04, keepaspectratio = 0.25]{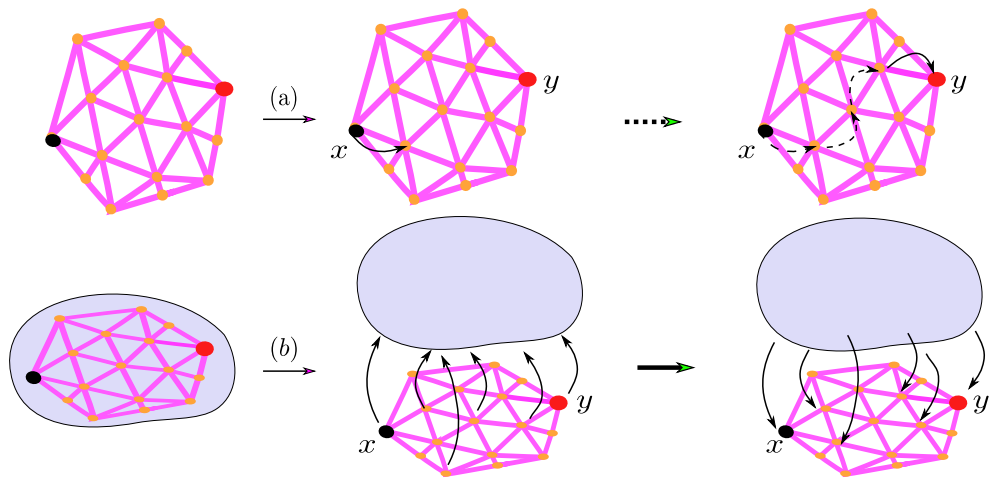}
\end{center}
\caption{Illustration of message-passing among binary or higher-order cells.
(a): Using a graph-based message-passing scheme,
some information that starts at vertex $x$ needs to travel a long distance,
that is a long edge path,
before reaching vertex $y$.
(b): Using a higher-order cell structure, indicated by the the blue cell,
the signal can be lifted from the vertices to a higher-order cell
and thus propagates back and forth
between vertices $x$ and $y$ in fewer steps. This shortcut allows
to share information efficiently, in fewer computational steps,
among all vertices that belong to the blue cell~\cite{hajij2022high}.\vspace{-4mm}}
\label{fig:ft}
\end{figure}

\paragraph{Construction of efficient graph-based message passing over long-range.}
The hierarchical structure of TDL networks facilitates the construction of long-range interactions in an efficient manner. 
By leveraging such structures, signals defined on the vertices of a domain can propagate efficiently distant dependencies among vertices connected by long edge paths. Figure~\ref{fig:ft} illustrates the process of lifting a signal defined on graph vertices by augmenting the graph with additional topological relations and unpooling the signal back to the vertices.~\cite{hajij2022high} employ such pooling and unpooling operations to construct models capable of capturing distant dependencies, leading to more effective and adaptive learning of both the local and global structure of the underlying domain. 
For related work on graph-based pooling, we refer the reader to~\cite{itoh2022multi,su2021hierarchical}.

\paragraph{Improved expressive power.}
TDL can capture intricate and subtle dependencies that elude traditional graph-based models. By allowing richer higher-order interactions, higher-order networks can have improved expressive power and lead to superior performance in numerous tasks~\cite{bodnar2021weisfeiler,morris2019weisfeiler}.

\subsection{A unifying perspective on deep learning and structured computations}

\begin{figure}[!t]
\begin{center}
\includegraphics[width=1.0\linewidth]{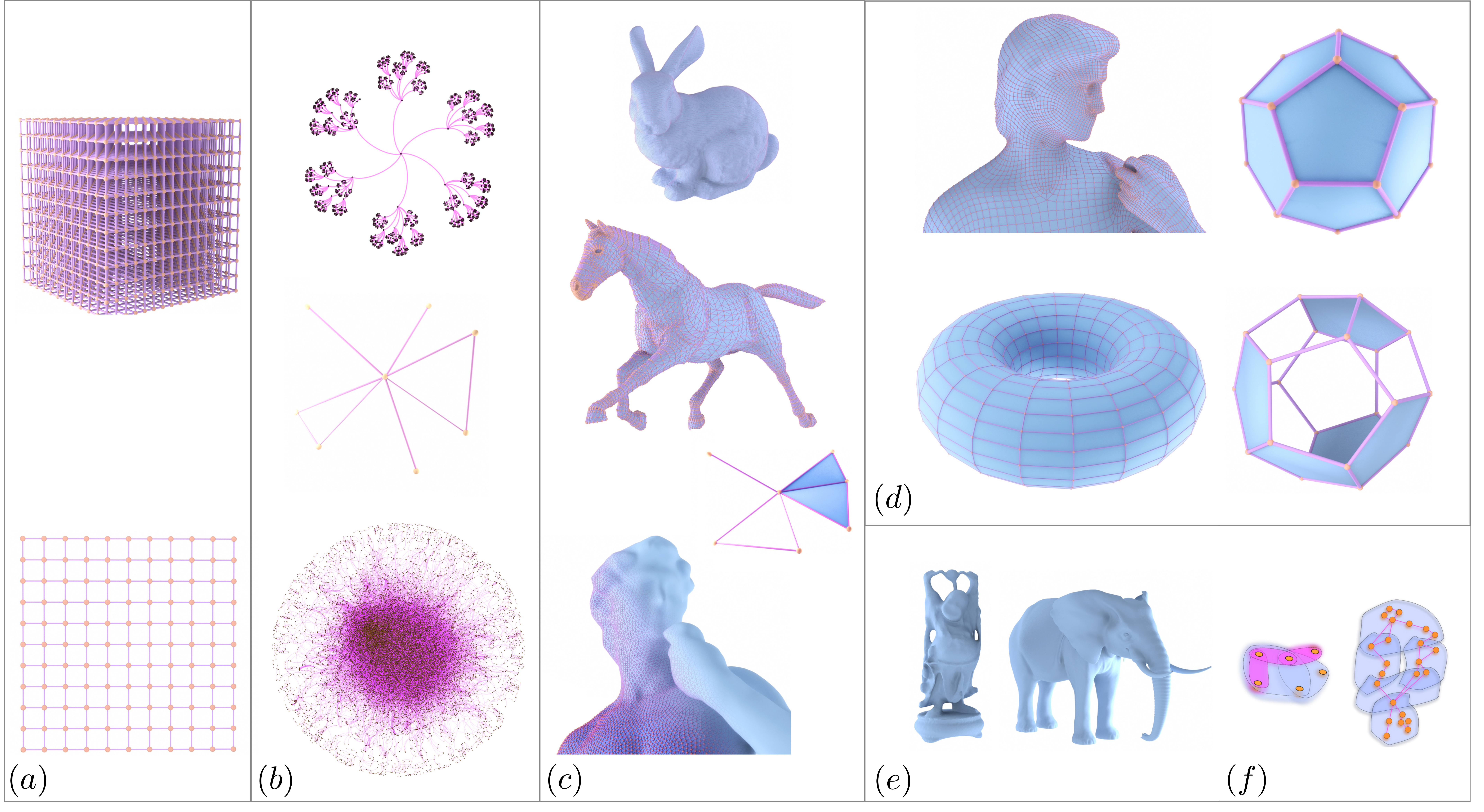}
\end{center}
\caption{This work introduces combinatorial complexes,
which are higher-order networks that
generalize most discrete domains typically encountered in scientific computing,
including
(a) sequences and images,
(b) graphs,
(c) 3D shapes and simplicial complexes,
(d) cubical and cellular complexes,
(e) discrete manifolds,
and (f) hypergraphs.\vspace{-3mm}}
\label{fig:unifying}
\end{figure}

Topology provides a framework to generalize numerous discrete domains that are typically encountered in scientific computations. 
Examples of these domains include graphs, simplicial complexes, and cell complexes (see Figure~\ref{fig:unifying}). 
Computational models supported on such domains can be seen as a generalization of various deep learning architectures; i.e., can provide a unifying framework to compare and design TDL architectures.
Further, understanding the connections between different types of TDL via a unifying theory can provide valuable insights into the underlying structure and behavior of complex systems, where higher-order relations naturally occur~\cite{hajijcell,hansen2019toward,bick2021higher,majhi2022dynamics}.

\paragraph{Understanding why deep networks work.}
A remarkable aspect of contemporary deep neural networks is their capacity to generalize very well to unseen data, even though they have an enormous number of parameters. This contradicts our previous understanding of statistical learning theory, prompting researchers to seek novel approaches in comprehending the underlying workings of deep neural networks~\cite{neyshabur2019role}.
Recent studies have revealed that the topology of graphs induced by deep neural networks or their corresponding training trajectories, modeled as Markov chains, exhibit strong correlations with generalization performance~\cite{birdal2021intrinsic,dupuis2023generalization}. An open question is how far such correlations between topological structure, computational function and generalization properties can be found in learning architectures supported on higher-order domains. We believe that by studying these questions using our novel topological constructs, such as CCs and $\CCN$s, we can gain deeper insights not only into TDL but also into deep learning more generally.

\section{Preliminaries: from graphs to higher-order networks}
\label{prelim}





The notion of proximity among entities in a set $S$ holds significant relevance in various machine learning applications as it facilitates the comprehension of inter-entity relationships in $S$. For example, clustering algorithms aim to group points that are proximal to each other. In recommendation systems, the objective is to suggest items that exhibit similarity to those for which a user has already expressed interest. However, the question is how do we precisely quantify the notion of proximity? 

\begin{figure}[!t]
\begin{center}
\includegraphics[scale = 0.073, keepaspectratio = 0.25]{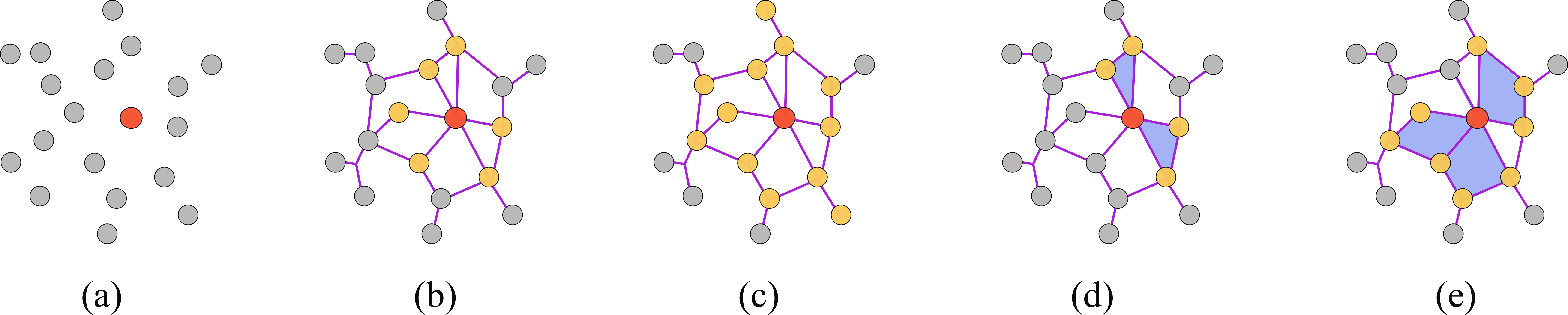}
\end{center}
\caption{Demonstration of the notion of proximity among the entities of a set $S$.
(a): A finite set $S$ of abstract entities.
(b): A neighborhood (in yellow) of an entity $x$ (in red) in $S$.
The neighborhood is defined to be
the set of entities in $S$ that are adjacent to $x$ via an edge.
(c): A neighborhood of $x$ consisting of all yellow entities in $S$
that have distance at most two from red entity $x$.
(d): A neighborhood of $x$ consisting of all yellow entities in $S$
that form triangles (in blue) incident to red entity $x$.
(d): A neighborhood of $x$ consisting of all yellow entities in $S$
that form trapezoids (in blue) incident to red entity $x$.\vspace{-3mm}}
\label{proximity}
\end{figure}

Consider a set $S$ consisting of a collection of abstract entities, as depicted in Figure~\ref{proximity}(a). Consider the red entity (vertex) $x$ in the same figure. We wish to identify the entities (vertices) in $S$ that are `closely related' or `in close proximity' to $x$. However, a set has no inherent notion of proximity or relations between its entities.

Defining binary relations (edges) among the entities of set $S$ provides one way of introducing the concept of proximity, which results in a graph whose vertex set is $S$, as shown in Figure~\ref{proximity}(b). Using this `auxiliary structure' of edges defined on top of set $S$, one may declare the `local neighborhood' of $x$, denoted by $\mathcal{N}(x)$, to be the subset of $S$ consisting of all entities that are adjacent to $x$ via an edge. In Figure~\ref{proximity}(b), the neighborhood of vertex $x$ (in red) in $S$ consists of all vertices colored in yellow.

The choice of neighborhood $\mathcal{N}(x)$ in Figure~\ref{proximity}(b) is arbitrary.
For example, an alternative valid notion of neighborhood is given by defining $\mathcal{N}(x)$
to contain all vertices whose distance from the red vertex $x$ is at most two,
represented by all vertices colored in yellow in Figure~\ref{proximity}(c).
In Figure~\ref{proximity}(d),
the neighborhood $\mathcal{N}(x)$ is chosen to consist of all yellow vertices that
form triangles incident to red vertex $x$.
In Figure~\ref{proximity}(e),
the neighborhood $\mathcal{N}(x)$ is chosen to consist of all yellow vertices that
form trapezoids incident to the red vertex $x$.
It becomes clear from Figures~\ref{proximity}(d) and (e)
that additional auxiliary structures,
such as triangles and trapezoids,
can be used to define the notion of neighborhood.
In practice, the choice of neighborhood typically depends on the application. Recent research has explored using graph geometry to obtain richer neighborhood notions, as seen in~\cite{morris2019weisfeiler,zhao2021stars,hajijcell}. However, the fundamental concept remains the same: one starts by introducing an auxiliary structure defined on a vertex set, and then utilizes the auxiliary structure to induce a well-defined notion of proximity.


Upon generalizing graphs to higher-order networks,
it is natural to also generalize
the notion of neighborhood for graphs to higher-order networks. The precise notion of neighborhood or proximity among entities of a set $S$ has been studied in topology~\cite{munkres1974}. A topology defined on $S$ allows us to meaningfully describe the proximity of the elements in $S$ to each other. This section introduces topological notions and definitions with the aim of generalizing graphs to higher-order networks.


\subsection{Neighborhood functions and topological spaces}

\label{sec:topology}



There are several equivalent ways to define a topological space. For example, topological spaces are commonly defined in terms of `open' or `closed' sets (e.g.,~\cite{munkres1974}).
In this work, we opt to define topological spaces in terms of `neighborhoods'. This definition is more compatible with the message-passing paradigm that is usually defined on graphs~\cite{gilmer2017neural}, and it can be generalized to higher-order networks. We refer the reader to \cite[Section 2.2]{brown2006} for an explanation of why the definition in terms of neighborhoods is equivalent to the definition in terms of open sets.





\begin{tcolorbox}
[width=\linewidth, sharp corners=all, colback=white!95!black]
\begin{definition}[Neighborhood function]
\label{NS}
Let $S$ be a nonempty set.
A \textbf{neighborhood function} on $S$
is a function $\mathcal{N}\colon S\to\mathcal{P}(\mathcal{P}(S))$
that assigns to each point $x$ in $S$
a nonempty collection $\mathcal{N}(x)$ of subsets of $S$.
The elements of $\mathcal{N}(x)$ are called \textbf{neighborhoods} of $x$ with respect to $\mathcal{N}$.
\end{definition}
\end{tcolorbox}





 \begin{tcolorbox}
 [width=\linewidth, sharp corners=all, colback=white!95!black]
\begin{definition}[Neighborhood topology]
\label{NT}
Let $\mathcal{N}$ be a neighborhood function on a set $S$.
$\mathcal{N}$ is called a \textbf{neighborhood topology} on $S$ if it satisfies the following axioms:
\begin{enumerate}
\item If $N$ is a neighborhood of $x$, then $x\in N$.
\item If $N$ is a subset of $S$ containing a neighborhood of $x$,
then $N$ is a neighborhood of $x$.
\item The intersection of two neighborhoods of a point $x$ in $S$ is a neighborhood of $x$.
\item Any neighborhood $N$ of a point $x$ in $S$ contains a neighborhood $M$ of $x$ such that $N$ is a neighborhood of each point of $M$.
\end{enumerate}
\end{definition}
\end{tcolorbox}

 \begin{tcolorbox}
 [width=\linewidth, sharp corners=all, colback=white!95!black]
\begin{definition}[Topological space]
\label{def:topospace}
A pair $(S,\mathcal{N})$ consisting of a nonempty set $S$ 
and a neighborhood topology $\mathcal{N}$ on $S$ is called a \textbf{topological space}.
\end{definition}
\end{tcolorbox}

Hence, a topological space is a set $S$
equipped with a neighborhood function $\mathcal{N}$, which satisfies the properties specified in Definition~\ref{NT}. In Section~\ref{sec:neighborhood_structure},
we introduce a similar notion of proximity in the context of higher-order networks. Further, the choice of neighborhood function $\mathcal{N}$ is the first and most fundamental step in the construction of a deep learning model supported on a higher-order domain (see Section~\ref{CCNNs}).

\subsection{Bridging the gap among higher-order networks}
\label{sec:examples.hod}


Given a finite set $S$ of abstract entities,
a neighborhood function $\mathcal{N}$ on $S$
can be induced by equipping $S$ with an auxiliary structure, such as edges,
as demonstrated in Figure~\ref{proximity}(b).
Edges provide one way of defining relations
among the entities of $S$\footnote{Recall that a relation on $S$ is a nonempty subset of $S$.}. Specifically, each edge defines a binary relation (i.e., a relation between two entities) in $S$. In many applications, it is desirable to permit relations that incorporate more than two entities. The idea of using relations that involve more than two entities is central to higher-order networks. Such higher-order relations allow for a broader range of neighborhood functions to be defined on $S$ to capture multi-way interactions among entities of $S$.

To describe more intricate multi-way interactions, it is necessary to employ more intricate neighborhood functions and topologies. With an eye towards defining a general higher-order network in Section~\ref{sec:CC}
(as motivated in Section~\ref{sec:motivation}), this section reviews the definitions, advantages, and disadvantages of some commonly studied higher-order networks, including (abstract) simplicial complexes, regular cell complexes, and hypergraphs. In Section~\ref{sec:CC}, we introduce combinatorial complexes,
which generalize and bridge the gaps between
all of these commonly studied topological domains.


Simplicial complexes are one of the simplest higher-order domains with many desirable properties, extending the corresponding properties of graphs. For instance, Hodge theory is naturally defined on simplicial complexes,
extending similar notions on 
graphs~\cite{schaub2020random,barbarossa2020topological,schaub2021signal}. 

\begin{tcolorbox}
[width=\linewidth, sharp corners=all, colback=white!95!black]
\begin{definition}[\href{https://app.vectary.com/p/4HZRioKH7lZ2jWESIBrjhf}{Simplicial complex}]
\label{SCs:main}

An \textbf{abstract simplicial complex}
on a nonempty set $S$ is a pair $(S,\mathcal{X})$, where $\mathcal{X}$ is a subset of $\mathcal{P}(S) \setminus \{\emptyset\}$ such that $ x \in \CCX$ and $y \subseteq x $ imply $y \in \CCX$. Elements of $\mathcal{X}$ are called \textbf{simplices}.
\end{definition}
\end{tcolorbox}

Figure~\ref{fig:unifying}(c) displays examples of triangular meshes, which are special cases of simplicial complexes with many applications in computer graphics.
We refer the reader to~\cite{schaub2021signal,crane2013digital}
for a relevant introduction to simplicial complexes.
As it is evident from Definition~\ref{SCs:main},
each relation $x$ on $S$ must contain all relations $y$ with $y\subseteq x$. 
Hence, a simplicial complex may encode a relatively large amount of data, taking up a lot of memory~\cite{roddenberry2021signal}. Further, real-world higher-order data (e.g., traffic-flows on a rectangular street network) may not admit a meaningful simplicial complex structure due to the inherent lack of available simplices on the underlying data space. To address this limitation, cell complexes~\cite{hatcher2005algebraic,hansen2019toward} can generalize simplicial complexes, and overcome many of their drawbacks.


\begin{tcolorbox}
[width=\linewidth, sharp corners=all, colback=white!95!black]
\begin{definition}[\href{https://app.vectary.com/p/3EBiRiJcYjFNvkbbWszQ0Z}{Regular cell complex}]
\label{RCC:main}
A \textbf{regular cell complex} is a topological space $S$ with a partition into subspaces (\textbf{cells}) $\{x_\alpha\}_{\alpha \in P_{S} }$, where $P_{S}$ is an index set, satisfying the following conditions:
\begin{enumerate}
\item $S= \cup_{\alpha \in P_{S}} \Int(x_{\alpha})$,
where $\Int(x)$ denotes the interior of cell $x$.
\item For each $\alpha \in P_S$,
there exists a homeomorphism $\psi_{\alpha}$, called an \textbf{attaching map}, 
from $x_\alpha$ to $\mathbb{R}^{n_\alpha}$ for some $n_\alpha\in \N$,
called the \textbf{dimension} $n_\alpha$ of cell $x_\alpha$.
\item For each cell $x_{\alpha}$,
the boundary $\partial x_{\alpha}$ is a union of finitely many cells,
each having dimension less than the dimension of $x_{\alpha}$. 
\end{enumerate}
\end{definition}
\end{tcolorbox}

For the sake of brevity, we henceforth refer to a `regular cell complex' as a `cell complex'. Cell complexes encompass a wide range of higher-order networks. Several types of higher-order networks can be viewed as instances of cell complexes. For instance, cell complexes are a natural generalization of graphs, simplicial complexes, and cubical complexes~\cite{hajijcell}. Figure~\ref{fig:unifying}(d) shows examples of cell complexes. Intuitively, a cell complex is a disjoint union of cells, with each of these cells being homeomorphic to the interior of a $k$-dimensional Euclidean ball for some $k$. These cells are attached together via attaching maps in a locally suitable manner. The information of attaching maps of a regular cell-complex can be stored combinatorially in a sequence of matrices, called the \textit{incidence matrices}~~\cite{hatcher2005algebraic}. We describe these matrices in detail in Section~\ref{inc-nbhd}.


Condition 3 in Definition~\ref{RCC:main}
is known as the \textit{regularity condition}
for regular cell complexes. The regularity condition implies that the topological information of a cell complex can be realized combinatorially by equipping the index set $P_{S}$ with a poset structure given by $\alpha\leq\beta$ if and only if $x_{\alpha} \subseteq \overline{x_{\beta}}$, where $\overline{x}$ denotes the closure of a cell $x$. This poset structure is typically called the \textit{face poset}~\cite{hansen2019toward}. It can be shown that the topological information encoded in a cell complex is completely determined by the face poset structure~\cite{hansen2019toward}, which allows cell complexes to be represented combinatorially
in practice via a poset
~\cite{aschbacher1996combinatorial,basak2010combinatorial,savoy2022combinatorial,klette2000cell}.


Definition~\ref{RCC:main} implies that the boundary cells of each cell in a cell complex are also cells in the cell complex.
Hence, one may think about a cell complex as a collection of cells of varying dimensions, which are related via their boundaries. In terms of relations,
this implies that the boundary of a cell in a cell complex must also be a cell in the cell complex. While cell complexes form a general class of higher-order networks,
this property sets a constraint on the relations of a cell complex.
Such a constraint may not be desirable in certain applications
if the data do not satisfy it.
To remove all constraints on the relations among entities of a set,
hypergraphs are typically considered.

\begin{tcolorbox}
[width=\linewidth, sharp corners=all, colback=white!95!black]
\begin{definition}[Hypergraph]
\label{hyperG:main}
A \textbf{hypergraph} on a nonempty set $S$ is a pair $ (S,\mathcal{X})$, where $\mathcal{X}$ is a subset of $\mathcal{P}(S)\setminus\{\emptyset\}$. Elements of $\mathcal{X}$ are called \textbf{hyperedges}.
\end{definition}
\end{tcolorbox}

A hyperedge of cardinality two is called an \textit{edge}.
Hypergraphs can be considered as a generalization of simplicial and cell complexes.
However, hypergraphs do not directly entail the notion
of the dimension of a cell (or of a relation),
which is explicitly encoded in the definition of cell complex
and is also indicated by the cardinality of a relation in a simplicial complex.
As we demonstrate in Section~\ref{sec:motiv},
the dimensionality of cells and relations in simplicial and cell complexes
can be used to endow these complexes with 
hierarchical structures,
which can be utilized for (un)pooling type computations on these structures. 

\subsection{Hierarchical structure and set-type relations}
\label{gap}

The properties of simplicial complexes, cell complexes and hypergraphs,
as outlined in Section~\ref{sec:examples.hod},
give rise to two main features of relations on higher-order domains,
namely hierarchies of relations and set-type relations. In the present subsection, we formalize these two features.


\begin{tcolorbox}
[width=\linewidth, sharp corners=all, colback=white!95!black]
\begin{definition}[Rank function]\label{def:rank}
A \textbf{rank function} on a higher-order domain $\mathcal{X}$ is an order-preserving function
$\rk\colon \CCX\to \Znon$; i.e., $x\subseteq  y$ implies $\rk(x) \leq \rk(y)$ for all $x,y\in\CCX$. 
\end{definition}
\end{tcolorbox}

Intuitively, a rank function $\rk$ on a higher-order domain $\mathcal{X}$
attaches a ranking, represented by a non-negative integer value,
to every relation in $\mathcal{X}$ such that
set inclusion in $\mathcal{X}$ is preserved via $\rk$.
Effectively, a rank function induces
a \textit{hierarchical structure} on  $\mathcal{X}$.
Cell and simplicial complexes are
common examples of higher-order domains
equipped with rank functions and therefore with hierarchies of relations. 


\begin{tcolorbox}
[width=\linewidth, sharp corners=all, colback=white!95!black]
\begin{definition}[Set-type relations]
Relations in a higher-order domain are called \textbf{set-type relations}
if the existence of a relation
is not implied by another relation in the domain.
\end{definition}
\end{tcolorbox}

Hypergraphs constitute examples of higher-order domains equipped with set-type relations. Given the modeling limitations of simplicial complexes, cell complexes, and hypergraphs, we develop the combinatorial complex in Section~\ref{sec:CC}, a higher-order domain
that features both hierarchies of relations and set-type relations.

\section{Combinatorial complexes (CCs)}
\label{sec:CC}
\begin{figure}[!t]
\begin{center}
\includegraphics[scale = 0.12, keepaspectratio = 0.25]{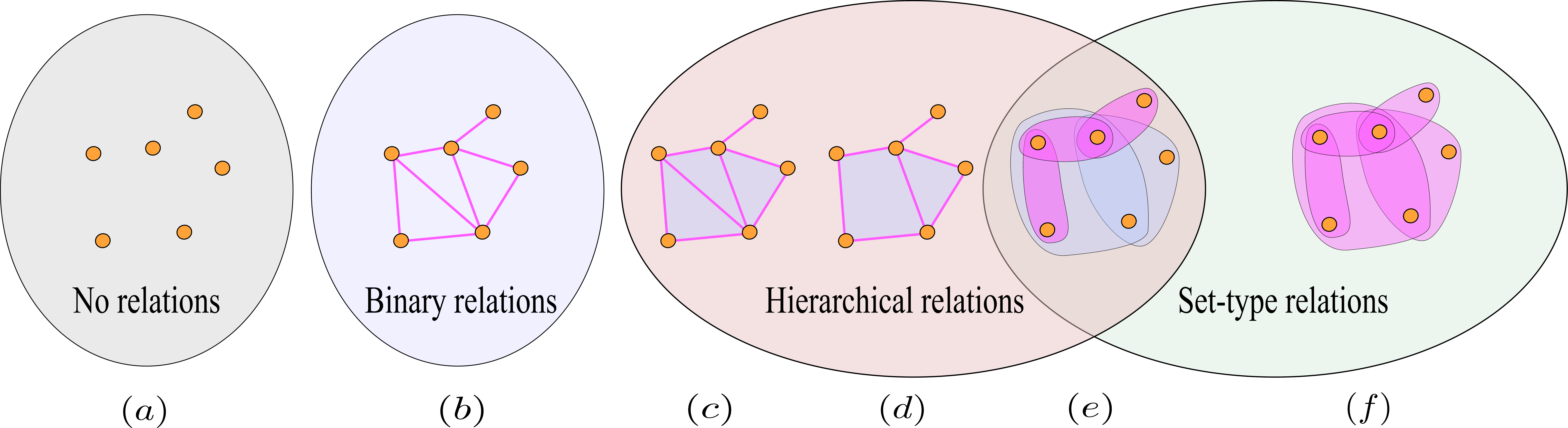}
\end{center}
\caption{Illustration of how CCs generalize various domains.
(a): A set $S$ consists of abstract entities (vertices) with no relations.
(b): A graph models binary relations between its vertices (i.e., elements of $S$). (c): A simplicial complex models a hierarchy of higher-order relations (i.e., relations between the relations) but with rigid constraints on the `shape' of its relations. (d): Similar to simplicial complexes, a cell complex models a hierarchy of higher-order relations, but with more flexibility in the shape of the relations (i.e., `cells'). (f): A hypergraph models arbitrary set-type relations between elements of $S$, but these relations do not have a hierarchy among them. (e): A CC combines features of cell complexes (hierarchy among its relations) and of hypergraphs (arbitrary set-type relations), generalizing both domains.\vspace{-3mm}}
\label{fig:HON}
\end{figure}

This section introduces combinatorial complexes (CCs), a novel class of higher-order domains that generalizes
graphs, simplicial complexes, cell complexes, and hypergraphs. Figure~\ref{fig:HON} shows a first illustration of the generalization that CCs provide over such domains. Moreover, Table~\ref{complex_summary} enlists the relation-related features of higher-order domains and graphs, thus summarizing the relation-related generalization
attained by CCs.

\begin{table}[t!]
	\begin{center}
	\begin{small}
		\begin{tabular}{l|c|c|c|c|c}
			\hline
			\multicolumn{1}{c|}{\multirow{2}{*}{Relation-related features}} &
			\multicolumn{4}{c}{Topological domains} &
                \multicolumn{1}{|c}{\multirow{2}{*}{Graph}} \\ \cline{2-5}
			& CC & Hypergraph & Cell complex     & Simplicial complex   &   \\ \hline
			Hierarchy of relations &  \checkmark     &   &  \checkmark    &  \checkmark   &     \\ \hline
			Set-type relations     &  \checkmark  &  \checkmark  &   &  &     \\ \hline
			Multi-relation coupling     &   \checkmark   &  &   \checkmark    &  \checkmark  &   \\ \hline
			Rank $\nRightarrow$ cardinality
   &   \checkmark   &   \checkmark   &   \checkmark    &   &      \\ \hline
		\end{tabular}
	\end{small}
	\end{center}
 \caption{Tabular summary of relation-related features of topological domains and of graphs.
 Recall that a relation is an element of a domain.
 A domain is specified via its relations and via the way these relations are related to each other. Desirable relation-related features are indicated in the first column. A \textit{hierarchy of relations} implies that relations of the higher-order domain can have different rankings. \textit{Set-type relations} are free from constraints among relations or their lengths. 
 \textit{Multi-relation coupling} implies that each relation
 can have other neighboring relations via multiple neighborhood functions
 defined on the higher-order domain.
 `\textit{Rank $\nRightarrow$ cardinality}' indicates that relations with the same ranking
 in a given hierarchy on the higher-order domain do not need to have the same cardinality.}
\label{complex_summary}
\end{table}

In Section~\ref{sec:CC.def}, we introduce the definition of CC and provide examples of CCs. In Section~\ref{sec:cc_homomorphisms},
we define the notion of CC-homomorphism and present relevant examples.
In Section~\ref{sec:motiv}, we present the motivation behind the CC structure from a practical point of view.
In Section~\ref{sec:neighborhood_structure},
we demonstrate the computational version
of neighborhood functions on neighborhood matrices. Finally, in Section~\ref{sec:data}, we introduce the notion of CC-cochain.



\subsection{CC definition}
\label{sec:CC.def}

We seek to define a structure that bridges the gap
between simplicial/cell complexes and hypergraphs,
as discussed in Section~\ref{gap}.
To this end, we introduce the combinatorial complex (CC),
a higher-order domain that can be viewed from three perspectives:
as a simplicial complex whose cells and simplices are allowed to be missing;
as a generalized cell complex with relaxed structure;
or as a hypergraph enriched through the inclusion of a rank function. 


\begin{tcolorbox}
[width=\linewidth, sharp corners=all, colback=white!95!black]
\begin{definition}[CC]
\label{def:cc} A \textbf{combinatorial complex (CC)} is a triple $(S, \CCX, \rk)$ consisting of a set $S$, a subset $\CCX$ of $\mathcal{P}(S)\setminus\{\emptyset\}$, and a function
$\rk \colon \CCX\to \Znon$ with the following properties:
	 \begin{enumerate}
	 \item for all $s\in S$, $\{s\}\in\CCX$, and
	 \item the function $\rk$ is order-preserving,
  which means that if $x,y\in \CCX$ satisfy $x\subseteq  y$, then $\rk(x) \leq \rk(y)$.
	 \end{enumerate}
  Elements of $S$ are called \textbf{entities} or \textbf{vertices}, elements of $\CCX$ are called \textbf{relations} or \textbf{cells}, and $\rk$ is called the \textbf{rank function} of the CC.
\end{definition}
\end{tcolorbox}

For brevity,
$\mathcal{X}$ is used as shorthand notation
for a CC $(S,\mathcal{X},\rk)$. 
Definition~\ref{def:cc} sets up
a framework to construct higher-order networks upon which we can define general purpose higher-order deep learning architectures. Observe that CCs exhibit both hierarchical and set-type relations. Particularly, the rank function $\rk$ of a CC induces a hierarchy of relations in the CC. Further, CCs encompass set-type relations as there are no relation constraints in Definition~\ref{def:cc}. Thus, CCs subsume cell complexes and hypergraphs in the sense that they combine the relation-related features of both. Table~\ref{complex_summary} provides a comparative summary of relation-related features
among CCs and common higher-order networks and graphs.


\begin{remark}
We typically require that $\rk(\{s\})=0$
for each singleton cell $\{s\}$ in a CC.
Such a convention aligns CCs naturally with simplicial and cellular complexes. 
\end{remark}



The rank of a cell $x\in\mathcal{X}$ is
the value $\rk(x)$ of the rank function $\rk$ at $x$.
The \emph{dimension} $\dim(\CCX)$ of a CC $\CCX$
is the maximal rank among its cells. A cell of rank $k$ is called a \emph{$k$-cell} and is denoted by $x^k$. The \textit{$k$-skeleton} of a CC $\CCX$, denoted $\CCX^{(k)}$, is the set of cells of rank at most $k$ in $\CCX$. The set of cells of rank exactly $k$ is denoted by $\mathcal{X}^k$. Note that this set corresponds to $\CCX^k=\rk^{-1}(\{k\})$. The $1$-cells are called the \emph{edges} of $\CCX$.
In general, an edge of a CC may contain more than two nodes. CCs whose edges have exactly two nodes are called \emph{graph-based} CCs. In this paper, we primarily work with graph-based CCs.

\begin{figure}[t!]
\begin{center}
\includegraphics[scale = 0.12, keepaspectratio = 0.20]{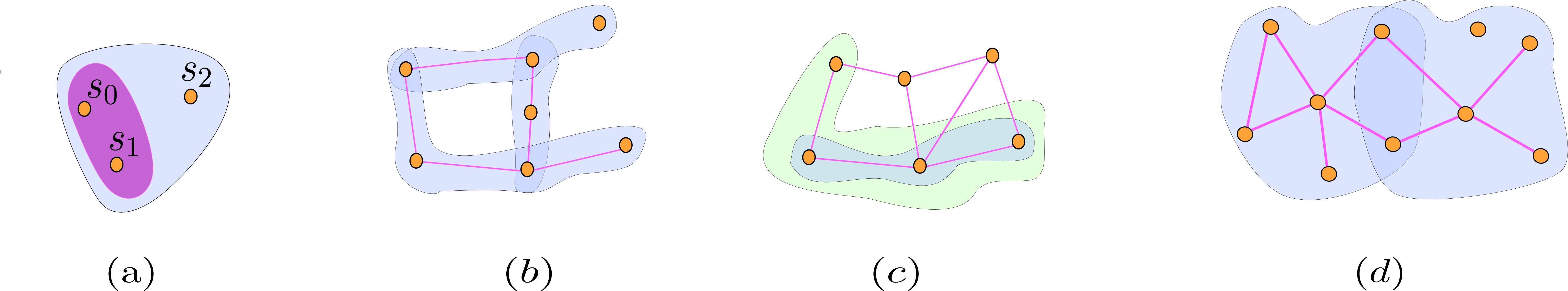}
\end{center}
\caption{Examples of CCs. Orange circles represent vertices. Pink, blue, and green colors represent cells of rank one, two, and three, respectively.
Each of the CCs in (a), (b), and (d) has dimension equal to two, whereas the CC in (c) has dimension equal to three.}
\label{other_examples}
\end{figure}

\begin{example}[CCs of dimension two and three]
\label{ex1}
Figure~\ref{other_examples} shows four examples of CCs. For instance, Figure~\ref{other_examples}(a) shows a 2-dimensional CC on a vertex set $S=\{s_0,s_1,s_2\}$,
consisting of
$0$-cells $\{s_0\}$, $\{s_1\}$, and $\{s_2\} $ (shown in orange),
$1$-cell $\{s_0, s_1\}$ (purple),
and $2$-cell $\{s_0, s_1, s_2\} = S$ (blue).


\end{example}

\subsection{CC-homomorphisms and sub-CCs}
\label{sec:cc_homomorphisms}

CC-homomorphisms are maps relating CCs to one another. CC-homomorphisms play an important role
in delineating the process
of lifting graphs or other higher-order domains to CCs.
Intuitively, a lifting map is a well-defined procedure
that converts a domain of certain type, such as a graph,
to a domain of another type, such as a CC.
The usefulness of lifting maps lie in their ability
to enable the application of deep learning models defined on CCs to more common domains, such as graphs, cell complexes or simplicial complexes.
We study lifting maps in detail and provide examples of them in Appendix~\ref{appdx:liftings}.
Definition~\ref{maps} formalizes the notion of CC-homomorphism.

\begin{tcolorbox}
[width=\linewidth, sharp corners=all, colback=white!95!black]
\begin{definition}[CC-homomorphism]
\label{maps}
 A homomorphism from a CC
 $ (S_1, \mathcal{X}_1, \rk_1)$ to a CC
 $(S_2, \mathcal{X}_2, \rk_2)$,
also called a \textbf{CC-homomorphism},
 is a function $f \colon \mathcal{X}_1 \to \mathcal{X}_2 $ that satisfies the following conditions:
\begin{enumerate}
\item If $x,y\in\mathcal{X}_1$ satisfy $x\subseteq y$, then $f(x) \subseteq f(y)$. 

\item If $x\in\mathcal{X}_1$, then $\rk_1(x)\geq \rk_2(f(x)).$
\end{enumerate}
\end{definition}
\end{tcolorbox}

The second condition in Definition~\ref{maps} assures that a CC-homomorphism may only map a $k$-cell in $ \mathcal{X}_1$ to a cell in $ \mathcal{X}_2 $ with rank no greater than $k$. When $\rk_1(x) = \rk_2(f(x))$ for all $x \in \mathcal{X}_1$ and $f$ is injective, then we call the homomorphism $f$ a \emph{CC-embedding}. CC-embeddings are useful in practice as they can be used to `lift' a domain, such as a graph, to a CC by augmenting that domain with higher-order cells. Example~\ref{ex:augmentation} displays three CC-embeddings,
while Example~\ref{ex:homomorphism} presents a CC-homomorphism that is not a CC-embedding.

\begin{figure}[!t]
\begin{center}
\includegraphics[scale = 0.108, keepaspectratio = 0.20]{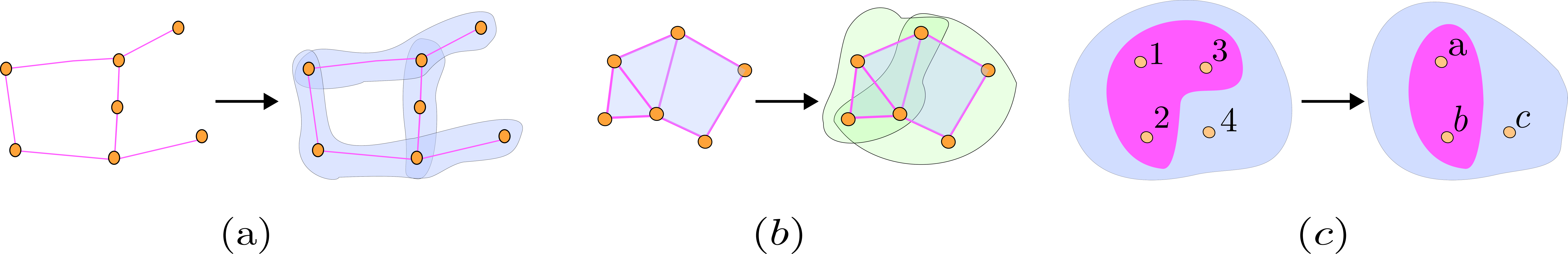}
\end{center}
\caption{Examples of CC-homomorphisms.
Pink, blue, and green colors represent cells of rank one, two, and three, respectively.
(a): An embedding of a $1$-dimensional CC into a $2$-dimensional CC.
(b): An embedding of a $2$-dimensional CC into a $3$-dimensional CC. (c) An example of a CC-homomorphism, as defined in Example~\ref{ex:homomorphism}. The CC $\CCX_1$ is on the left, $\CCX_2$ is on the right, and the homomorphism $f$ is represented by the black arrow. 
 Intuitively, the CC-homomorphism $f$ can be viewed as a combinatorial analogue to a continuous function between $S_1 = \{1, 2, 3,4\}$ and $S_2 = \{a, b, c\}$ that `collapses' the cell $\{1,2,3\}$ to the cell $\{a,b\}$, and the cell $\{1,2,3,4\}$ to the cell $\{a,b,c\}$. Observe that CC-homomorphisms generalize simplicial maps~\cite{munkres2018elements} from this perspective.}

\label{A non-embedding CC-homomorphism}
\end{figure}

\begin{example}[CC-embeddings]
\label{ex:augmentation}
Figures~\ref{A non-embedding CC-homomorphism}(a)
and~(b)
show two CC-embeddings. 
In Figure~\ref{A non-embedding CC-homomorphism}(a), each cell in the graph on the left-hand side is sent to its corresponding cell in the CC on the right-hand side.
Similarly, in Figure~\ref{A non-embedding CC-homomorphism}(b), each cell in the cell complex on the left is sent to its corresponding cell in the CC on the right.
It is easy to verify that each of these two maps is a CC-embedding.
\end{example}

\begin{example}[A CC-homomorphism]
\label{ex:homomorphism}
We present an example of a CC-homomorphism that is not a CC-embedding.
Consider the sets $S_1 = \{1, 2, 3,4\}$ and $S_2 = \{a, b, c\}$.
Let $\CCX_1$ denote the CC on $S_1$ consisting of
one 3-cell $\{1,2,3,4\}$, one 2-cell $\{1,2,3\}$, and four 0-cells corresponding to the elements of $S_1$.
Likewise, let $\CCX_2$ denote the CC on $S_2$ consisting of
one 3-cell $\{a,b,c\}$, one 2-cell $\{a,b\}$, and three 0-cells corresponding to the elements of $S_2$.
CCs $\CCX_1$ and $\CCX_2$ are visualized in Figure~\ref{A non-embedding CC-homomorphism}(c).
Consider the function $f \colon S_1 \to S_2$ defined as
$f(1) = f(2) = a,~f(3) = b$ and $f(4) = c$.
It is easy to verify that $f$ induces a CC-homomorphism from $\CCX_1$ to $\CCX_2$.
\end{example}


\begin{tcolorbox}
[width=\linewidth, sharp corners=all, colback=white!95!black]
\begin{definition}[Sub-CC]
\label{sub-cc}
Let $(S,\CCX, \rk)$ be a CC.
A \textbf{sub-combinatorial complex (sub-CC)} of a CC $(S,\CCX, \rk)$ is a CC $(A,\mathcal{Y},\rk^{\prime})$ such that $A\subseteq S$, $\mathcal{Y}\subseteq\mathcal{X}$ and
$\rk^{\prime} = \rk|_{\mathcal{Y}}$ is the restriction of $\rk$ on $\mathcal{Y}$.
\end{definition}
\end{tcolorbox}

For brevity, we refer to the sub-CC $(A,\mathcal{Y},\rk^{\prime})$ as $\mathcal{Y}$.
Any subset of $A \subseteq S$ can be used to induce a sub-CC as follows. Consider the set $\CCX_A = \{x \in \CCX  \mid x \subseteq A\}$ together with the restriction $\rk|_{\CCX_A}$. It is easy to see that the triple $(A,\CCX_A,\rk|_{\CCX_A})$ constitutes a CC,
which we call the \emph{sub-CC of $\CCX$ induced by $A$}.
Note that any cell in the set $\CCX$ induces a sub-CC obtained by considering all the cells that are contained in it. Finally, for any $k$,
it is easy to see that the skeleton $\mathcal{X}^{k}$ of a CC $\mathcal{X}$ is a sub-CC.

\begin{example}[Sub-CC]
Recall the CC
$\CCX= \{\{s_0\}, \{s_1\}, \{s_2\}, \{s_0, s_1\}, \{s_0, s_1, s_2\}\}$
displayed in Figure~\ref{other_examples}(a).
The set $A = \{s_0, s_1\}$ induces the sub-CC $\CCX_A = \{\{s_0\}, \{s_1\}, \{s_0, s_1\}\}$
of $\CCX$.    
\end{example}

\subsection{Motivation for CCs}
\label{sec:motiv}
Definition~\ref{def:cc} for CCs
aims to fulfill all motivational aspects
of higher-order modeling, as outlined in Section~\ref{sec:motivation}.
To further motivate Definition~\ref{def:cc},
we consider the pooling operations on CCs along with several structural advantages of CCs.

\subsubsection{Pooling operations on CCs}
\label{sec:pooling.motivation}

\begin{figure}[!t]
\begin{center}
\includegraphics[scale = 0.04, keepaspectratio = 1]{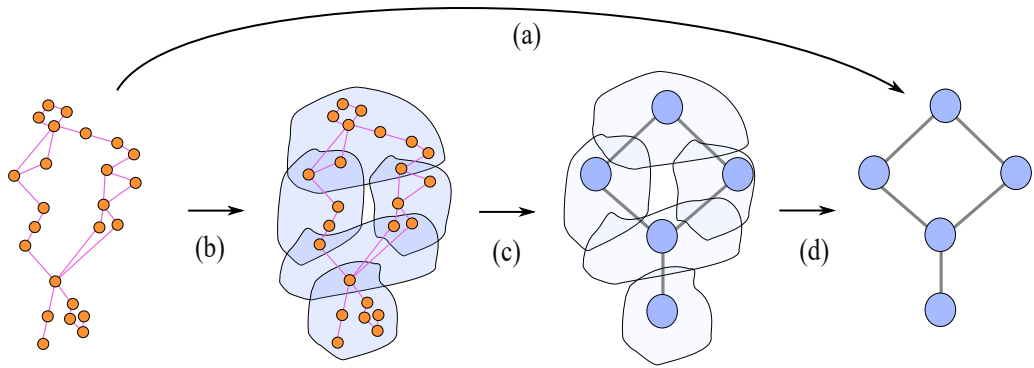}
\end{center}
\caption{Graph-based pooling formulated via CCs.
(a): Pooling on graphs
is expressed here in terms of a pooling function that maps a graph
along with the data defined on it to a coarser version of the graph.
The super-vertices in the pooled graph on the right correspond
to a clustering of vertices in the original graph on the left,
whereas edges in the graph on the right indicate the presence or absence
of a connection among these clusters.
(b-d): The super-vertices in the pooled graph on the right can be realized
as augmented higher-order cells (blue cells) in a CC obtained from the original graph.
The edges in the pooled graph on the right can be realized
in terms of an incidence matrix of the CC
between the vertices and the higher-order augmented cells.}
\label{fig:pooling_motivation}
\end{figure}

We first consider the general characteristics of pooling on graphs, and then demonstrate how graph-based pooling can be realized in a unified manner via CCs.
A general pooling function on a graph $\mathcal{G}$ is a function $ \mathcal{POOL} \colon \mathcal{G} \to \mathcal{G}^{\prime }$, where $\mathcal{G}^{\prime}$ is a pooled graph that represents a coarsened version of $\mathcal{G}$.
A vertex of $\mathcal{G}^{\prime}$ corresponds to a cluster of vertices (a super-vertex) in the original graph $\mathcal{G}$, while edges in $\mathcal{G}^{\prime}$ indicate the presence or absence of a connection among such clusters. See Figure~\ref{fig:pooling_motivation} for an example.

The formalism proposed via CCs encompasses graph-based pooling as a special case. Specifically, the super-vertices (cluster of vertices in $\mathcal{G}$) defined by the vertices in $\mathcal{G}^{\prime}$ can be realized as higher-order ranked cells in a CC obtained by augmenting $\mathcal{G}$ with the cells representing these super-vertices. The hierarchical structure consisting of the original graph $\mathcal{G}$ as well as augmented higher-order cells induces a CC. This realization of graph-based pooling in terms of CC-based higher-ranked cells is hierarchical because the new cells can be grouped together in a recursive manner to obtain a coarser version of the underlying space. Given such a notion of higher-order pooling operations on CCs, we demonstrate how to express graph/mesh pooling and image-based pooling in Sections \ref{graph.pooling} and \ref{image.pooling}, respectively. From this perspective, CCs provide a general framework for defining pooling operations on higher-order networks, which include graphs and images as special cases.
 
In addition, two main features of CCs are exploited via higher-order pooling operations. First, the ability of CCs to model set-type cells provides flexibility in the shape of the clusters that define the pooling operation. This flexibility is needed to realize novel user-defined pooling operations, in which the shape of the clusters might be task-specific. Second, higher-order ranked cells in a CC correspond to a coarser version of the underlying space. The type of higher-order pooling proposed here enables the construction of coarser representations. See Figure~\ref{fig:pooling_motivation} for an example. Note that less general structures, such as hypergraphs and cell complexes, do not accommodate simultaneous flexibility in cluster shaping and generation of coarser representations of the underlying space.

\subsubsection{Structural advantages of CCs}

In addition to the realization of graph-based pooling, CCs offer several structural advantages. Specifically, CCs unify numerous commonly used higher-order networks, enable fine-grained analysis of topological properties, facilitate message passing of topological features in deep learning, and accommodate flexible modeling of relations among relations.

\paragraph{Flexible higher-order structure and fine-grained message passing.}
Message-passing graph models pass messages between vertices in a graph to learn the representations of the graph, with messages computed based on the vertex and edge features. These messages update the vertex and edge features, and gather information from the graph's local neighborhood. The rank functions of CCs
render CCs more versatile than other higher-order networks and than graphs in two fronts.
First, rank functions make CCs
more flexible in terms of higher-order structure representation. Second, in the context of deep learning,
rank functions equip CCs with more fine-grained message-passing capabilities.
For instance, each hyperedge in a hypergraph is treated as a set without any notion of a rank,
and consequently all hyperedges are treated uniformly without any distinction. Further details are provided in Section~\ref{CCNNs}.

\paragraph{Flexible modeling of relations among relations.}
In the process of populating a topological domain with topological data, constructing meaningful relations can be challenging due to the inherent lack of data that can be naturally supported on all cells in the domain. This is particularly true when working with simplicial or cell complexes. For instance, any relation between $k$ entities of a simplicial complex must be built from the relations of all corresponding subsets of $k-1$ entities. Real-world data may contain a subset of these relations, and not all of them. While cell complexes offer more flexibility in modeling relations, the boundary conditions that cell complexes must satisfy constraints the types of permissible relations. In order to remove all restrictions among relations, hypergraphs come to aid as they allow arbitrary set-type relations. However, hypergraphs do not offer hierarchical features, which can be disadvantageous in applications that require considering local and global features simultaneously.

\subsection{Neighborhood functions on CCs}
\label{sec:neighborhood_structure}

We introduce the notion of a CC-neighborhood function on a CC as a mechanism of exploiting the topological information stored in the CC.
In practice, crafting a neighborhood function is usually a part of the learning task. For our purposes, we restrict the discussion to two types of generalized neighborhood functions, namely those specifying adjacency and incidence.
From a deep learning perspective,
CC-neighborhood functions set the foundations
to extend the general message-passing schemes of deep learning models,
thus subsuming several state-of-the-art GNNs~\cite{fey2019fast,loukas2019graph,battaglia2018relational,morris2019weisfeiler,battaglia2016interaction,kipf2016semi}.

Given a CC, we aim to describe cells
in the local proximity of a sub-CC of the CC.
To this end, we define the CC-neighborhood function, an analogue of Definition~\ref{NS} in the context of CCs.

\begin{tcolorbox}
[width=\linewidth, sharp corners=all, colback=white!95!black]
\begin{definition}[CC-neighborhood function]
\label{neighborhood_structure}
A \textbf{CC-neighborhood function} on a CC $(S,\CCX, \rk)$
is a function $\mathcal{N}$ that assigns to every sub-CC
$(A,\mathcal{Y},\rk^{\prime})$ of the CC
a nonempty collection $\mathcal{N}(\mathcal{Y})$ of subsets of $S$.
\end{definition}
\end{tcolorbox}

Without loss of generality, we assume that the elements of the neighborhood $\mathcal{N}(\mathcal{Y})$ are cells or sub-CCs of $\mathcal{X}$. Intuitively, the neighborhood $\mathcal{N}(\mathcal{Y})$ of the sub-CC $\mathcal{Y}$ is a set of subsets of $S$ that are in the `local vicinity' of $\mathcal{Y}$. The term `local vicinity' is generally stated here, since it is typically context-specific. 



\begin{figure}[!t]
\begin{center}
\includegraphics[scale = 0.15, keepaspectratio = 0.20]{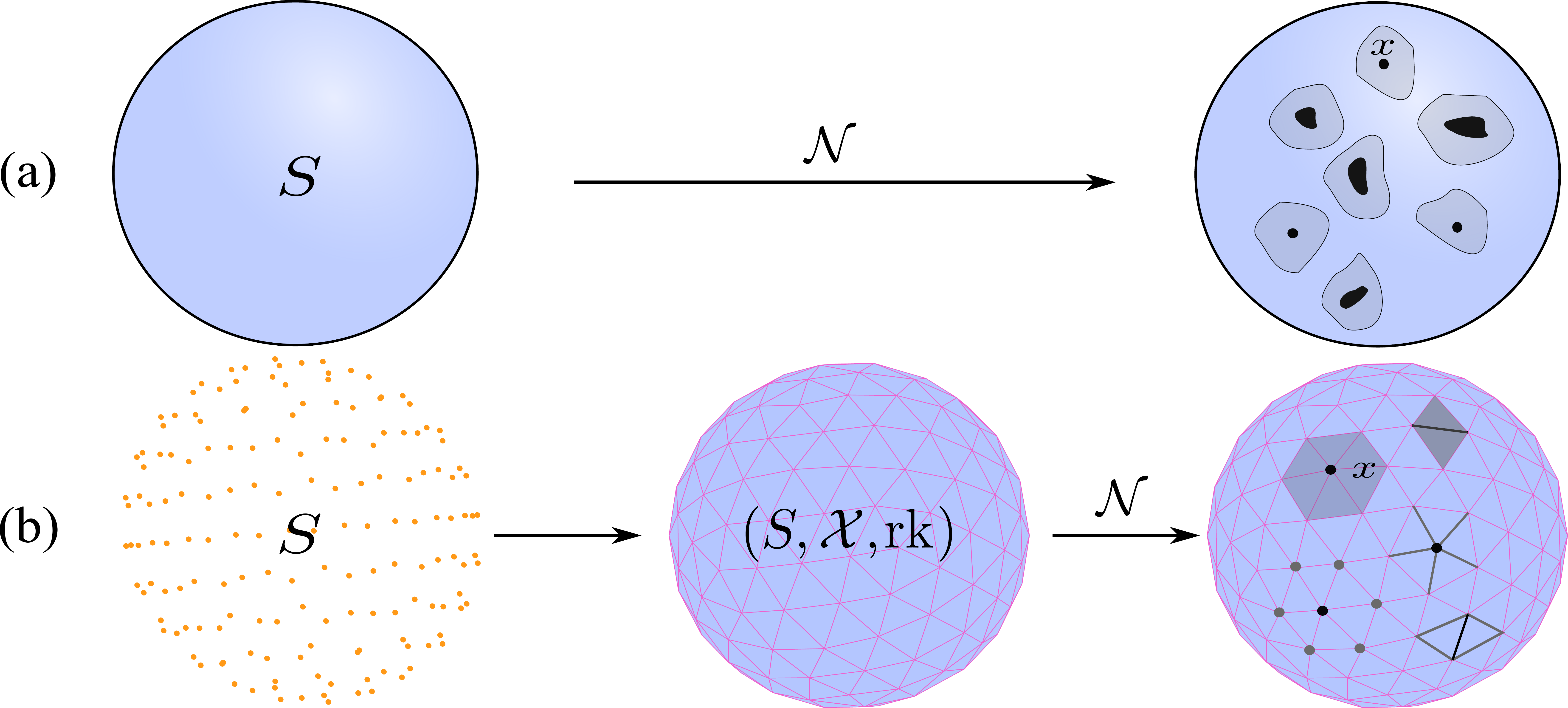}
\end{center}
\caption{A visual comparison between 
neighborhood functions with continuous domains
and CC-neighborhood functions, adhering to
Definitions~\ref{NS} and~\ref{neighborhood_structure}, respectively.
(a): A neighborhood function with a continuous domain $S$
assigns to $x\in S$
a set $\mathcal{N}(x)$ of subsets of $S$
that are in the local vicinity of $x$.
(b): Similarly,
a CC-neighborhood function on a CC $(S,\CCX, \rk)$
assigns to $x\in S$ a set $\mathcal{N}(x)$ of subsets of $\mathcal{X}$
that are in the local vicinity of $x$.\vspace{-3mm}}
\label{CC-nbhd}
\end{figure}

Definition~\ref{neighborhood_structure} is a discrete analogue
of the well-known Definition~\ref{NS}.
The correspondence between these two definitions
is demonstrated in Figure~\ref{CC-nbhd}.
For the rest of the paper,
CC-neighborhood functions are succinctly called neighborhood functions. In practice, the information encoded in CC-neighborhood functions
is represented in terms of matrices as described next. 

\paragraph{Neighborhood matrix induced by a neighborhood function.}
For the purposes of computation,
it is convenient to represent neighborhood functions as matrices. Incidence, adjacency, and coadjacency matrices are well-known examples of matrices that encode respective neighborhood functions.
In Definition~\ref{neighborhood_matrix}, we introduce a generalization of these matrices called the `neighborhood matrix'.
In this definition and henceforth, we denote the cardinality of a set $S$ by $|S|$.

\begin{tcolorbox}
[width=\linewidth, sharp corners=all, colback=white!95!black]
\begin{definition}[Neighborhood matrix]
\label{neighborhood_matrix}
Let $\mathcal{N}$ be a neighborhood function defined on a CC $\CCX$.
Moreover, let $\mathcal{Y}=\{y_1,\ldots,y_n\}$ and $\mathcal{Z}=\{z_1,\ldots,z_m\}$
be two collections of cells in $\CCX$ such that $\mathcal{N}(y_{i}) \subseteq \CCZ$
for all $1\leq i \leq n $.
The \textbf{neighborhood matrix
of $\mathcal{N}$ with respect to $\mathcal{Y}$ and $\mathcal{Z}$} is the
$|\CCZ| \times |\CCY|$ binary matrix $G$ whose 
$(i,j)$-th entry $[G]_{ij}$ has value $1$ if $z_i \in \mathcal{N}(y_j)$
and $0$ otherwise.
\end{definition}
\end{tcolorbox}

\begin{remark}
\label{neighborhood_matrix_remark}
In Definition~\ref{neighborhood_matrix},
$\mathcal{N}(y_j)$ is stored in the $j$-th column
of the associated neighborhood matrix $G$. For this reason,
we denote by $\mathcal{N}_{G}(j)$ the neighborhood function of the
cell $y_j$ when we work with the neighborhood matrix $G$.
\end{remark}


There are many ways to define useful neighborhood functions on a CC.  In this work, we constrain ourselves to the most immediate neighborhood functions: the incidence and adjacency neighborhood functions. 


\subsubsection{Incidence in a CC} 
\label{inc-nbhd}

We define three notions of incidence
to capture different facets
of incidence structures of cells in a CC.
First, in Definition~\ref{def:nstruct},
we introduce the down-incidence and up-incidence neighborhood functions
to describe the incidence structure of a cell
via cells of arbitrary rank.
Second, in Definition~\ref{filtered_incidence},
we introduce the $k$-down and $k$-up incidence neighborhood functions
to describe the incidence structure of a cell
via cells of a particular rank $k$.
Third, in Definition~\ref{def:inc_mat},
we introduce $(r, k)$-incidence matrices 
to describe the incidence structures of cells
of particular ranks $r$ and $k$.
In what follows, we assume that the cells
in the set $\CCX$ of a CC $(S,\CCX, \rk)$
are given a fixed order.

\begin{tcolorbox}
[width=\linewidth, sharp corners=all, colback=white!95!black]
\begin{definition}[Down/up-incidence neighborhood functions] 
\label{def:nstruct}
\hfill Let $(S,\CCX, \rk)$ be a CC.
Two cells $x, y\in\mathcal{X}$ of the CC
are called \textbf{incident} if either $x \subsetneq y$ or $y \subsetneq x$.
In particular, the \textbf{down-incidence neighborhood function} $\mathcal{N}_{\searrow}(x)$
of a cell $x\in\mathcal{X}$ is defined to be the set
$\{ y\in \CCX \mid y \subsetneq x\}$,
while the \textbf{up-incidence neighborhood function} $\mathcal{N}_{\nearrow}(x)$ of $x$
is defined to be the set
$\{ y\in \CCX \mid x \subsetneq y\}$.
\end{definition}
\end{tcolorbox}

Definition~\ref{filtered_incidence}
provides a more granular specification of incidence
than Definition~\ref{def:nstruct}.
In particular,
Definitions~\ref{filtered_incidence} and~\ref{def:nstruct}
describe the incidence structure of a cell
with respect to cells of a particular rank
or of arbitrary rank, respectively.

\begin{tcolorbox}
[width=\linewidth, sharp corners=all, colback=white!95!black]
\begin{definition}[$k$-down/up incidence neighborhood functions]
\label{filtered_incidence}
Let $(S,\CCX, \rk)$ be a CC.
For any $k\in\N$,
the \textbf{$k$-down incidence neighborhood function}
$\mathcal{N}_{\searrow,k}(x)$ of a cell $x\in\mathcal{X}$
is defined to be the set
$\{ y\in \CCX \mid y \subsetneq x, \rk(y)=\rk(x)-k \}$.
The \textbf{$k$-up incidence neighborhood function}
$\mathcal{N}_{\nearrow,k}(x)$ of $x$
is defined to be the set
$\{ y\in \CCX \mid y \subsetneq x, \rk(y)=\rk(x)+k \}$.
\end{definition}
\end{tcolorbox}

\begin{figure}[!t]
\begin{center}
\includegraphics[scale = 0.11, keepaspectratio = 0.20]{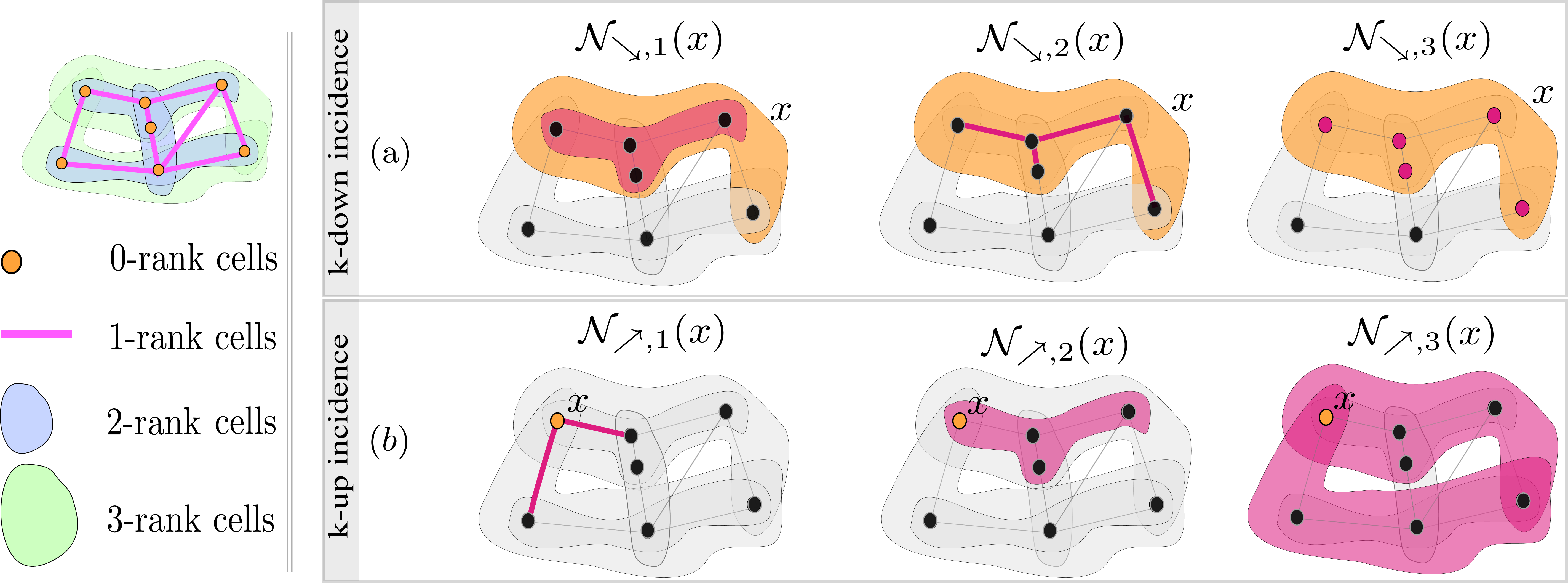}
\end{center}
\caption{Illustration of
$k$-down and $k$-up incidence neighborhood functions on a CC of dimension three.
(a): Illustration of $k$-down incidence neighborhood functions.
The target orange cell $x$ has rank three.
From left to right, the red cells represent
$\mathcal{N}_{\searrow,1}(x)$, $\mathcal{N}_{\searrow,2}(x)$
and $\mathcal{N}_{\searrow,3}(x)$. 
(b): Illustration of $k$-up incidence neighborhood functions.
The target orange cell $x$ has rank zero.
From left to right, the red cells represent
$\mathcal{N}_{\nearrow,1}(x)$, $\mathcal{N}_{\nearrow,2}(x)$
and $\mathcal{N}_{\nearrow,3}(x)$.}
\label{structure_inc}
\end{figure}

Clearly,
$\mathcal{N}_{\searrow}(x)= \bigcup_{k\in \N} \mathcal{N}_{\searrow,k}(x)$
and
$\mathcal{N}_{\nearrow}(x)= \bigcup_{k\in\N} \mathcal{N}_{\nearrow,k}(x)$. 
Immediate incidence is particularly important.
To this end,
the set of \emph{faces} of a cell $x \in \CCX $ is defined to be
$\mathcal{N}_{\searrow,1} (x)$,
and the set of \emph{cofaces} of a cell $x$ is defined to be
$\mathcal{N}_{\nearrow,1} (x)$.
An illustration of
$k$-down and $k$-up incidence neighborhood functions
is given in Figure~\ref{structure_inc}.

\begin{tcolorbox}
[width=\linewidth, sharp corners=all, colback=white!95!black]
\begin{definition}[Incidence matrix]
\label{def:inc_mat}
Let $(S,\CCX, \rk)$ be a CC.
For any $r,k \in \Znon$ with $0\leq r<k \leq \dim(\CCX) $, the
\textbf{$(r,k)$-incidence matrix} 
$B_{r,k}$ between $\CCX^{r}$ and $\CCX^{k}$
is defined to be the $ |\CCX^r| \times |\CCX^k|$ binary matrix
whose $(i, j)$-th entry $[B_{r,k}]_{ij}$
equals one if $x^r_i$ is incident to $x^k_j$
and zero otherwise.
\end{definition}
\end{tcolorbox}

The incidence matrices $B_{r,k}$
of Definition~\ref{def:inc_mat} determine a neighborhood function
on the underlying CC.
Neighborhood functions
induced by incidence matrices 
are utilized to construct
a higher-order message passing scheme
on CCs, as described in Section~\ref{CCNNs}.



\subsubsection{Adjacency in a CC}
\label{adj-nbhd}

CCs admit incidence as well as other neighborhood functions. For instance, for a CC that is reduced to a graph, a more natural neighborhood function is based on the notion of adjacency relation.
Incidence defines relations among cells of different ranks, while adjacency defines relations among cells of similar ranks.
Definitions~\ref{def:coa_nf},~\ref{def:k_coa_nf} and~\ref{def:adj}/\ref{def:coadj}
introduce the (co)adjacency analogues
of the respective incidence-related
Definitions~\ref{def:nstruct},~\ref{filtered_incidence}, and~\ref{def:inc_mat}.

\begin{tcolorbox}
[width=\linewidth, sharp corners=all, colback=white!95!black]
\begin{definition}[(Co)adjacency neighborhood functions]
\label{def:coa_nf}
Let $(S,\CCX, \rk)$ be a CC.
The \textbf{adjacency neighborhood function} $\mathcal{N}_{a}(x)$ of a cell $x\in \CCX$
is defined to be the set 
\begin{equation*}
\{  y \in \CCX \mid \rk(y)=\rk(x), \exists z \in \CCX \text{ with } \rk(z)>\rk(x) \text{ such that } x,y\subsetneq z\}.
\end{equation*}
The \textbf{coadjacency neighborhood function} $\mathcal{N}_{co}(x)$ of $x$ is defined to be the set
\begin{equation*}
\{ y \in \CCX \mid \rk(y)=\rk(x), \exists z \in \CCX
\text{ with } \rk(z)<\rk(x) \text{ such that } z\subsetneq y\text{ and }z\subsetneq x \}.
\end{equation*}
A cell $z$ satisfying the conditions of
either $\mathcal{N}_{a}(x)$
or $\mathcal{N}_{co}(x)$
is called a \textbf{bridge cell}.
\end{definition}
\end{tcolorbox}

\begin{tcolorbox}
[width=\linewidth, sharp corners=all, colback=white!95!black]
\begin{definition}[$k$-(co)adjacency neighborhood functions]
\label{def:k_coa_nf}
Let $(S, \CCX, \rk)$ be a CC.
For any $k\in\N$,
The \textbf{$k$-adjacency neighborhood function} $\mathcal{N}_{a,k}(x)$
of a cell $x \in \CCX$ is defined to be the set
\begin{equation*}
\{ y \in \CCX \mid \rk(y)=\rk(x),
\exists z \in \CCX
\text{ with } \rk(z)=\rk(x)+k \text{ such that } x,y\subsetneq z \}.
\end{equation*}
The \textbf{$k$-coadjacency neighborhood function}
$\mathcal{N}_{co,k}(x)$ of $x$ is defined to be the set
\begin{equation*}
 \{ y \in \CCX \mid \rk(y)=\rk(x),
\exists z \in \CCX
\text{ with } \rk(z)=\rk(x)-k 
\text{ such that } z\subsetneq y\text{ and }z\subsetneq x \}.
\end{equation*}
\end{definition}
\end{tcolorbox}

\begin{tcolorbox}
[width=\linewidth, sharp corners=all, colback=white!95!black]
\begin{definition}[Adjacency matrix]
\label{def:adj}
For any $r\in\Znon$ and $k\in \mathbb{Z}_{>0} $ with $0\leq r<r+k \leq \dim(\CCX) $,
the \textbf{$(r,k)$-adjacency matrix} $A_{r,k}$
among the cells of $\CCX^{r}$
with respect to the cells of $\CCX^{k}$ 
is defined to be the $ |\CCX^r| \times |\CCX^r|$ binary matrix
whose $(i, j)$-th entry $[A_{r,k}]_{ij}$ equals one
if $x^r_i$ is $k$-adjacent to $x^r_j$ and zero otherwise.
\end{definition}
\end{tcolorbox}

\begin{tcolorbox}
[width=\linewidth, sharp corners=all, colback=white!95!black]
\begin{definition}[Coadjacency matrix]
\label{def:coadj}
For any $r\in \Znon$ and $k\in\N$ with $0\leq r-k<r \leq \dim(\CCX) $, the \textbf{$(r,k)$-coadjacency matrix} $coA_{r,k}$
among the cells of $\CCX^{r}$
with respect to the cells of $\CCX^{k}$ 
is defined to be the $ |\CCX^r| \times |\CCX^r|$ binary matrix
whose $(i, j)$-th entry $[coA_{r,k}]_{ij}$ has value $1$
if $x^r_i$ is $k$-coadjacent to $x^r_j$ and $0$ otherwise.
\end{definition}
\end{tcolorbox}

\begin{figure}[!t]
	\begin{center}
		\includegraphics[scale = 0.11, keepaspectratio = 0.20]{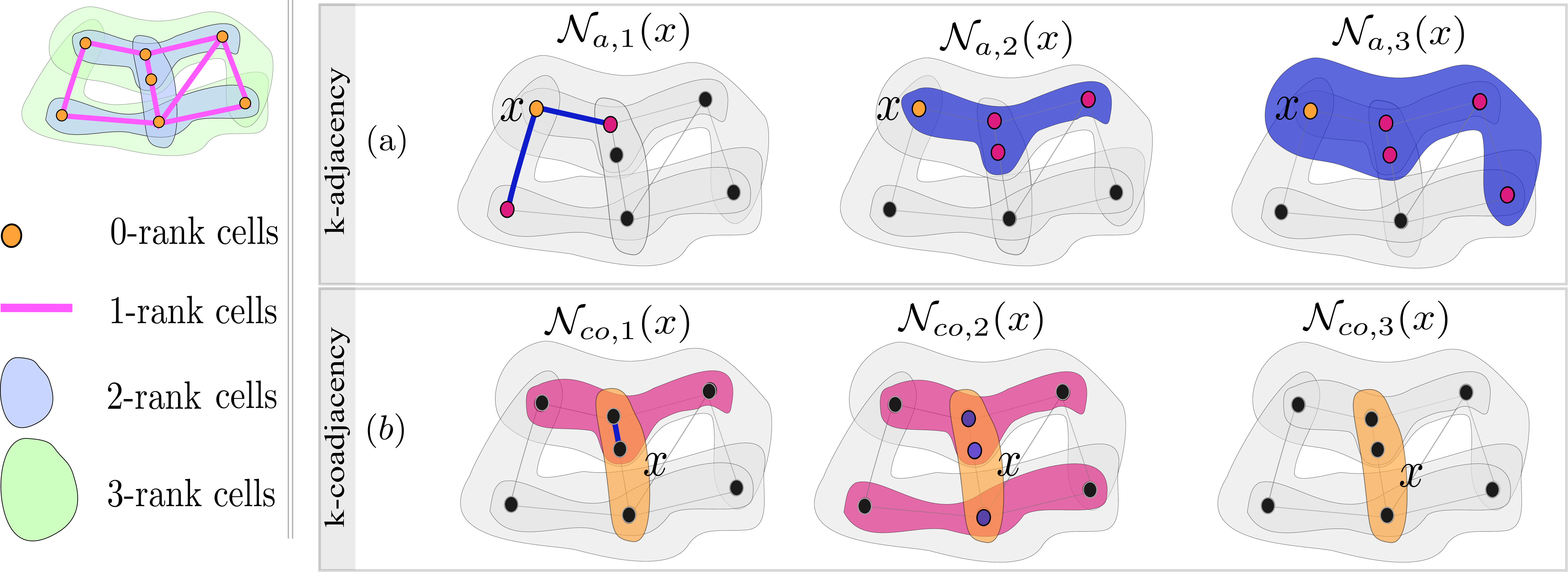}
	\end{center}
	\caption{Illustration of $k$-(co)adjacency neighborhood functions on a CC of dimension three.
 (a): Illustration of $k$-adjacency neighborhood functions.
		The target orange cell $x$ has rank zero. From left to right, the red cells represent $\mathcal{N}_{a,1}(x)$, 
		$\mathcal{N}_{a,2}(x)$ and $\mathcal{N}_{a,3}(x)$.
		(b): Illustration of $k$-coadjacency neighborhood functions.
		The target orange cell $x$ has rank two. From left to right, the red cells represent $\mathcal{N}_{co,1}(x)$,
		$\mathcal{N}_{co,2}(x)$ and $\mathcal{N}_{co,3}(x)$. }
	\label{structure_adj}
\end{figure}

Clearly,
$\mathcal{N}_{a}(x)= \cup_{k\in\N} \mathcal{N}_{a,k}(x)$
and
$\mathcal{N}_{co}(x)= \cup_{k\in\N} \mathcal{N}_{co,k}(x)$.
An illustration of
$k$-adjacency and $k$-coadjacency neighborhood functions
is given in Figure~\ref{structure_adj}.

\subsection{Data on CCs: cochain spaces and maps}
\label{sec:data}

As we are interested in processing the data defined over a CC $(S,\CCX,\rk)$, we introduce $k$-cochain spaces, $k$-cochains and cochain maps.

\begin{tcolorbox}
[width=\linewidth, sharp corners=all, colback=white!95!black]
\begin{definition}[k-cochain spaces]
Let $\mathcal{C}^k(\CCX,\mathbb{R}^d )$
be the $\mathbb{R}$-vector space of functions
$\mathbf{H}_k\colon\CCX^k\to \mathbb{R}^d$ for a rank $k \in \Znon$ and dimension $d$.
$d$ is called the \textbf{data dimension}.
$\mathcal{C}^k(\CCX,\mathbb{R}^d)$ is called the \textbf{$k$-cochain space}.
Elements $\mathbf{H}_k$ in $\mathcal{C}^k(\CCX,\mathbb{R}^d)$
are called \textbf{$k$-cochains} or \textbf{$k$-signals}. 
\end{definition}
\end{tcolorbox}

\begin{figure}[!t]
\begin{center}
\includegraphics[scale = 0.22, keepaspectratio = 0.20]{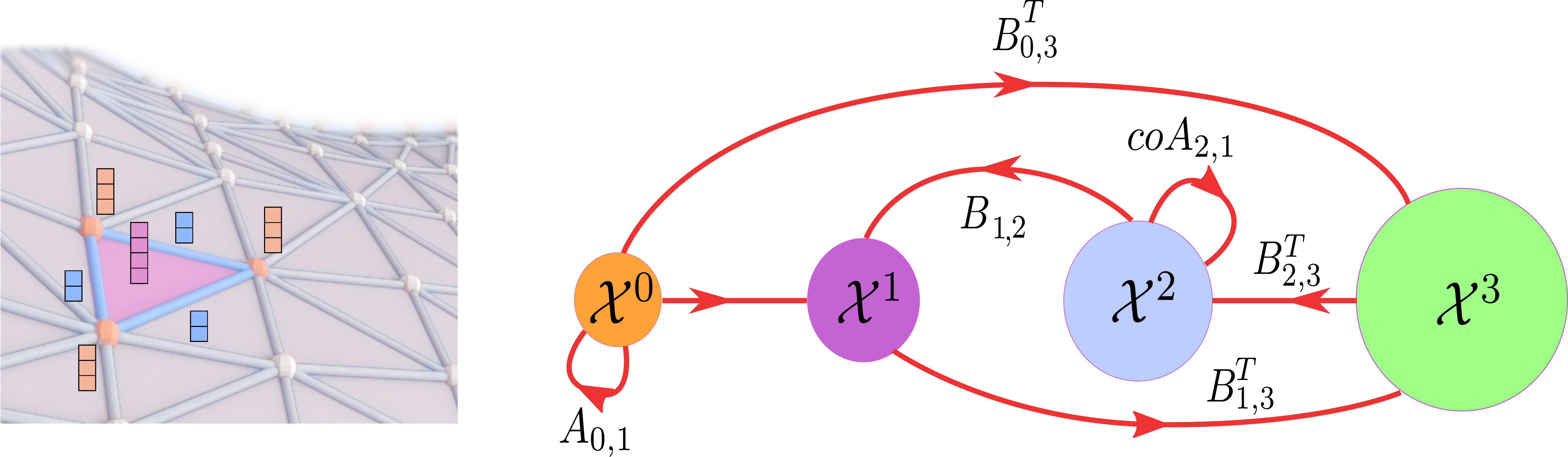}
\end{center}
\caption{Examples of $k$-cochains (left) and cochain maps (right) supported on a CC of dimension four.
Left: a $k$-cochain can be interpreted as a signal or a feature vector defined on the $k$-cells. In the figure, a 3-dimensional cochain is attached to the vertices, 2-dimensional cochains are attached to the 1-cells, and 4-dimensional cochains to the 2-cells.
Right: each of $coA_{r,k}$ and $A_{r,k}$ defines a cochain map between cochain spaces of equal dimension,
whereas each $B_{r,k}$ defines a cochain map between cochain spaces of different dimensions.\vspace{-2mm}}
\label{CC-cochain}
\end{figure}

We use the notation $\mathcal{C}^k(\CCX)$ or $\mathcal{C}^k$
when the underlying CC is clear.
Moreover, we say that a $k$-cochain space $\mathcal{C}^k(\CCX)$
is defined on $\mathcal{X}$.
Intuitively, a $k$-cochain can be interpreted as a signal
defined on the $k$-cells of $\CCX$~\cite{grady2010discrete}.
Figure~\ref{CC-cochain}(a) shows an example of a cochain
supported on 0, 1, and 2-rank cells of a simplicial complex.

When $\CCX$ is a graph, $0$-cochains correspond to graph signals~\cite{ortega2018graph}.  Ordering the cells in $ \CCX^k$, we canonically identify $\mathcal{C}^k(\CCX,\mathbb{R}^d ) $ with the Euclidean vector space $\mathbb{R}^{ |\CCX^k| \times d}$ and explicitly write $\mathbf{H}_k  $ as the vector $ [ \mathbf{h}_{x^k_1},\ldots,\mathbf{h}_{x^k_{|\CCX^k|} }]$, where $\mathbf{h}_{x^k_j} \in \mathbb{R}^d$ is a feature vector associated with the cell $x^k_j$.
The notation $\mathbf{H}_{k,j}$ refers to the feature vector $\mathbf{h}_{x^k_j}$ to avoid explicit reference to cell $x^k_j$. We also work with maps between cochain spaces, which we call cochain maps.

\begin{tcolorbox}
[width=\linewidth, sharp corners=all, colback=white!95!black]
\begin{definition}[Cochain maps]
For $r< k$, an incidence matrix $B_{r,k}$ induces a map
\begin{align*}
    B_{r,k}\colon \mathcal{C}^k(\CCX) &\to   \mathcal{C}^r(\CCX),\\
    \mathbf{H}_k &\to  B_{r,k}(\mathbf{H}_k),
\end{align*}
where $B_{r,k}(\mathbf{H}_k)$ denotes the usual product $B_{r,k} \mathbf{H}_k$
of matrix $B_{r,k}$ with vector $\mathbf{H}_k$.
Similarly, an $(r,k)$-adjacency matrix $A_{r,k}$ induces a map
\begin{align*}
    A_{r,k}\colon \mathcal{C}^r(\CCX) &\to   \mathcal{C}^r(\CCX),\\
    \mathbf{H}_r &\to  A_{r,k}(\mathbf{H}_r).
\end{align*}
\end{definition}
These two types of maps between cochain spaces are called \textbf{cochain maps}.
\end{tcolorbox}

Cochain maps serve as operators that `shuffle' and `redistribute' the data on the 
underlying CC. In the context of TDL, cochain maps are the main tools for defining
higher-order message passing (Section~\ref{sec:homp}) as well 
as (un)pooling operations (Section~\ref{subsec:cc_pool}).
Each adjacency matrix $A_{r,k}$
or coadjacency matrix $coA_{r,k}$
defines a cochain map between cochain spaces of equal dimension,
while each incidence matrix $B_{r,k}$
defines a cochain map between different dimensions.
Figure~\ref{CC-cochain}(b)
shows examples of cochain maps on a CC of dimension 4.

\section{Combinatorial complex neural networks (CCNNs)}
\label{CCNNs}

The modelling flexibility of CCs enables the exploration and analysis of a wide spectrum of CC-based neural network architectures. A CC-based neural network can exploit all neighborhood matrices or a subset of them, thus accounting for multi-way interactions among various cells in the CC to solve a learning task. In this section, we introduce the blueprint for TDL by developing the general principles of CC-based TDL models. We utilize our TDL blueprint framework for examining current approaches and offer directives for designing novel models.

The learning tasks in TDL can be broadly classified into three categories:
cell classification, complex classification, and cell prediction. Our numerical experiments in Section~\ref{exp}
provide examples on cell and complex classification. In more detail, the learning tasks of the three categories are the following:
\begin{itemize}[itemsep=1pt,topsep=1pt]
 \item \textit{Cell classification}: the goal is to predict targets for each cell in a complex. To accomplish this, we can utilize a TDL classifier that takes into account the topological neighbors of the target cell and their associated features. An example of cell classification is triangular mesh segmentation, in which the task is to predict the class of each face or edge in a given mesh.
\item \textit{Complex classification}: the aim is to predict targets for an entire complex. To achieve this, we can reduce the topology of the complex into a common representation using higher-order cells, such as pooling, and then learn a TDL classifier over the resulting flat vector. An example of complex classification is class prediction for each input mesh.
\item \textit{Cell prediction}: the objective is to predict properties of cell-cell interactions in a complex, and, in some cases, to predict whether a cell exists in the complex. This can be achieved by utilizing the topology and associated features of the cells. A relevant example is the prediction of linkages among entities in hyperedges of a hypergraph.
\end{itemize}

\begin{figure}[!t]
\begin{center}
\includegraphics[scale = 0.037, keepaspectratio = 0.25]{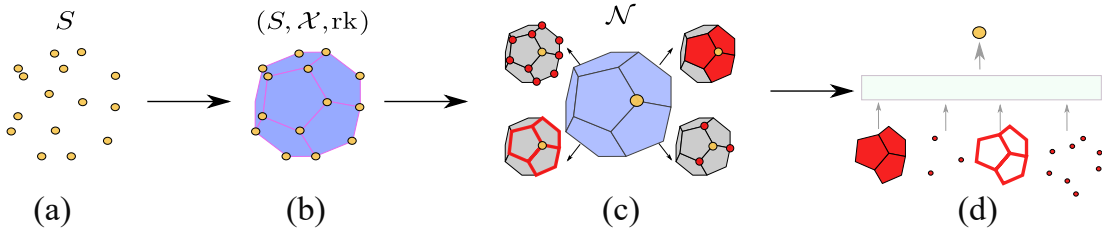}
\end{center}
\caption{A TDL blueprint.
(a): A set of abstract entities.
(b): A CC $(S, \mathcal{X}, \rk)$ is defined on $S$.
(c): For an element $x \in \mathcal{X}$,
we select a collection of neighborhood functions defined on the CC.
(d): We build a neural network on the CC using the neighborhood functions selected in (c).
The neural network exploits the neighborhood functions selected in (c) to update the data supported on $x$.}
\label{fig:TDL}
\end{figure}

Figure~\ref{fig:TDL} outlines our general setup for TDL. Initially, a higher-order domain, represented by a CC, is constructed on a set $S$.
A set of neighborhood functions defined on the domain is then selected. The neighborhood functions are usually selected based on the learning problem at hand and they are used to build a topological neural network.
To develop our general TDL framework, we introduce
\textit{combinatorial complex neural networks (CCNNs)},
an abstract class of neural networks supported on CCs
that effectively captures the pipeline of Figure~\ref{fig:TDL}.
CCNNs can be thought of as a \textit{template} that generalizes
many popular architectures,
such as convolutional and attention-based neural networks.
The abstraction of CCNNs offers many advantages.
First, any result that holds for CCNNs
is immediately applicable to any particular instance of CCNN architecture.
Indeed, the theoretical analysis and results in this paper are applicable to any CC-based neural network as long as it satisfies the CCNN definition. Second, working with a particular parametrization might be cumbersome
if the neural network has a complicated architecture. In Section~\ref{subsec:building_ccnns},
we elaborate on the intricate architectures of
parameterized TDL models.
The more abstract high-level representation of CCNNs simplifies the notation and the general purpose of the learning process, thereby making TDL modelling more intuitive to handle.

\begin{tcolorbox}
[width=\linewidth, sharp corners=all, colback=white!95!black]
\begin{definition}[CCNN]
\label{hoans_definition}
Let $\CCX$ be a CC.
Let $\mathcal{C}^{i_1}\times\mathcal{C}^{i_2}\times \ldots \times  \mathcal{C}^{i_m}$ and $\mathcal{C}^{j_1}\times\mathcal{C}^{j_2}\times \ldots \times  \mathcal{C}^{j_n}$ be a Cartesian product of $m$ and $n$ cochain spaces defined on $\CCX$. A \textbf{combinatorial complex neural network (CCNN)} 
is a function of the form
\begin{equation*} 
\CCN: \mathcal{C}^{i_1}\times\mathcal{C}^{i_2}\times \ldots \times  \mathcal{C}^{i_m} \longrightarrow \mathcal{C}^{j_1}\times\mathcal{C}^{j_2}\times \ldots \times \mathcal{C}^{j_n}.
\end{equation*}
\end{definition}
\end{tcolorbox}

Intuitively, a CCNN
takes a vector of cochains $(\mathbf{H}_{i_1},\ldots, \mathbf{H}_{i_m})$ as input
and returns a vector of cochains $(\mathbf{K}_{j_1},\ldots, \mathbf{K}_{j_n})$ as output.
In Section~\ref{subsec:building_ccnns}, we show how neighborhood functions
play a central role in the construction of a general CCNN.
Definition~\ref{hoans_definition} does not show how a CCNN can be computed in general.
Sections
~\ref{homp-merge} and~\ref{pooling-hoans}
formalize the computational workflow in CCNNs.





%

\subsection{Building CCNNs: tensor diagrams}
\label{subsec:building_ccnns}

Unlike graphs that involve vertex or edge signals, higher-order networks entail a higher number of signals (see Figure~\ref{CC-cochain}). Thus, constructing a CCNN requires building a non-trivial amount of interacting sub-networks. Due to the relatively large number of cochains in a CCNN, we introduce \textit{tensor diagrams}, a diagrammatic notation for higher-order networks.

\begin{remark}
 Diagrammatic notation is common in the geometric topology literature~\cite{hatcher2005algebraic, turaev2016quantum}, and it is typically used to construct functions built from simpler building blocks. See Appendix~\ref{tqft} for further discussion. See also~\cite{roddenberry2021principled} for related constructions on simplicial neural networks. 
\end{remark}

\begin{tcolorbox}
[width=\linewidth, sharp corners=all, colback=white!95!black]
\begin{definition}[Tensor diagram]
\label{TD}
A \textbf{tensor diagram} represents a CCNN via a directed graph. The signal on a tensor diagram flows from the \textbf{source nodes} to the \textbf{target nodes}. The source and target nodes correspond to the domain and codomain of the CCNN.
 \end{definition}
\end{tcolorbox}

\begin{figure}[!t]
\begin{center}
\includegraphics[scale = 0.90, keepaspectratio = 1]{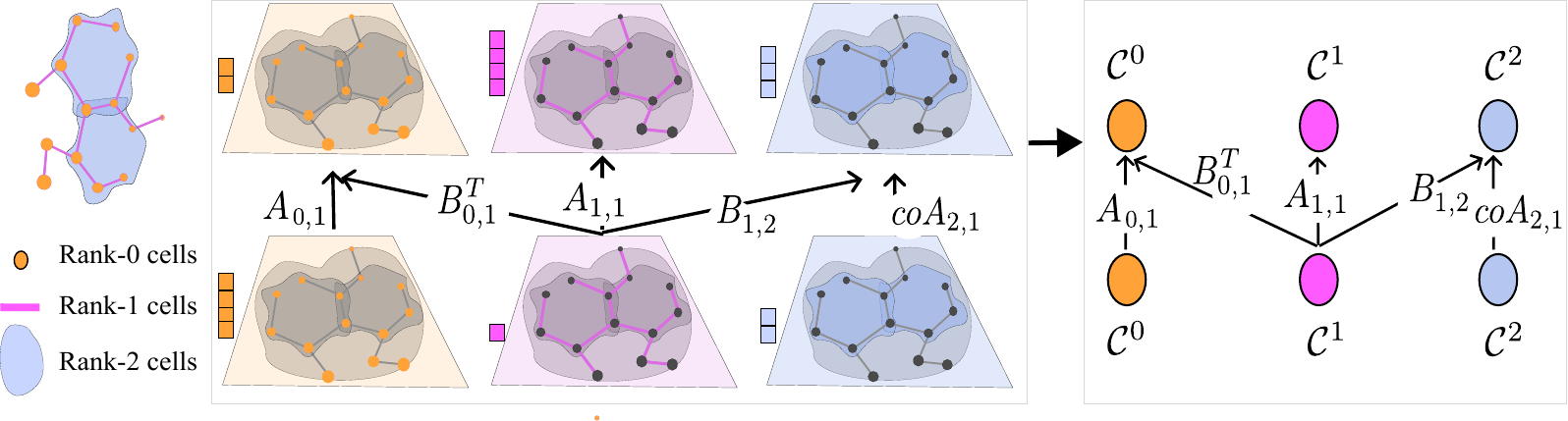}
\end{center}
\caption{A tensor diagram is a diagrammatic representation of a CCNN
that captures the flow of signals on the CCNN.}
\label{fig:TD}
\end{figure}

Figure~\ref{fig:TD} depicts an example of a tensor diagram.
On the left, a CC of dimension three is shown.
Consider a 0-cochain $\mathcal{C}^0$,
a 1-cochain $\mathcal{C}^1$ and
a 2-cochain $\mathcal{C}^2$.
The middle figure displays a CCNN
that maps a cochain vector in
$\mathcal{C}^0 \times \mathcal{C}^1\times \mathcal{C}^2$
to a cochain vector in $\mathcal{C}^0\times\mathcal{C}^1 \times \mathcal{C}^2$.
On the right, a tensor diagram representation of the CCNN is shown.
We label each edge on the tensor diagram by a cochain map or by its matrix representation.
The edge labels on the tensor diagram of Figure~\ref{fig:TD}
are $A_{0,1}, B_{0,1}^{T}, A_{1,1}, B_{1,2}$ and $coA_{2,1}$. 
Thus, the tensor diagram specifies the flow of cochains on the CC.

The  labels on the arrows of a tensor diagram
form a sequence $\mathbf{G}= (G_i)_{i=1}^l$ 
of cochain maps defined on the underlying CC.
In Figure~\ref{fig:TD} for example,
$\mathbf{G}=(G_i)_{i=1}^5 = (A_{0,1}, B_{0,1}^{T}, A_{1,1}, B_{1,2}, coA_{2,1})$.
When a tensor diagram is used to represent a CCNN, we use the notation $\CCN_{\mathbf{G}}$ for the tensor diagram and for its corresponding CCNN.
The cochain maps $(G_i)_{i=1}^l$ reflect the structure of the CC
and are used to determine the flow of signals on the CC.
Any of the neighborhood matrices mentioned in Section~\ref{sec:neighborhood_structure}
can be used as cochain maps.
The choice of cochain maps depends on the learning task.

\begin{figure}[!t]
\begin{center}
\includegraphics[scale = 0.20, keepaspectratio = 1]{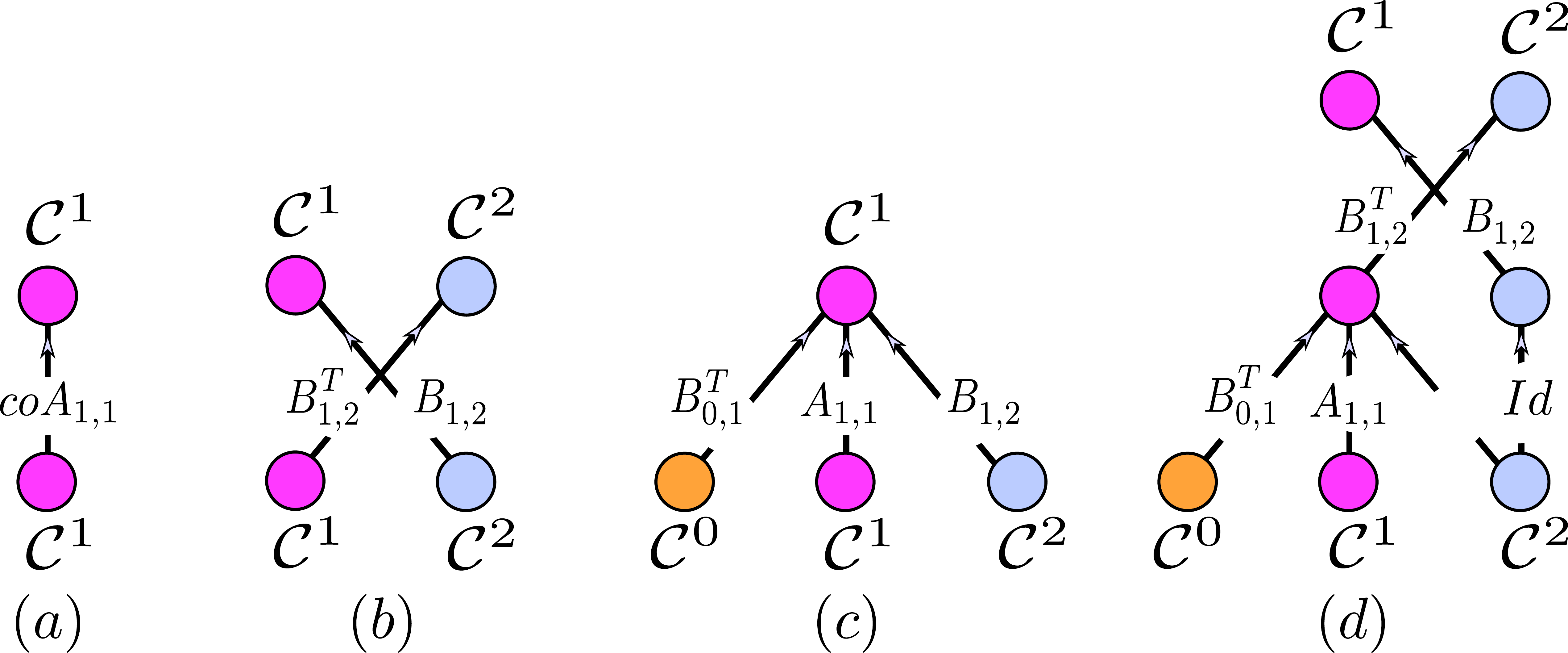}
\end{center}
\caption{Examples of tensor diagrams.
(a): Tensor diagram of a $\CCN_{coA_{1,1}}\colon \mathcal{C}^1 \to \mathcal{C}^1$.
(b): Tensor diagram of a
$\CCN_{ \{B_{1,2}, B_{1,2}^T\}} \colon \mathcal{C}^1 \times \mathcal{C}^2
\to \mathcal{C}^1 \times \mathcal{C}^2$.
(c): A merge node that merges three cochains.
(d): A tensor diagram generated by vertical concatenation
of the tensor diagrams in (c) and (b).
The edge label $Id$ denotes the identity matrix.}
\label{fig:tensor}
\end{figure}

Figure~\ref{fig:tensor} visualizes additional examples of tensor diagrams.
The \textit{height} of a tensor diagram is the number of edges on a longest path from a source node to a target node. For instance, the heights of the tensor diagrams
in Figures~\ref{fig:tensor}(a) and~\ref{fig:tensor}(d) are one and two, respectively.
The vertical concatenation of two tensor diagrams
represents the composition of their corresponding CCNNs.
For example, the tensor diagram in Figure~\ref{fig:tensor}(d)
is the vertical concatenation of the tensor diagrams
in Figures~\ref{fig:tensor}(c) and~(b).
 
If a node in a tensor diagram receives one or more signals, we call it a \textit{merge node}. Mathematically, a merge node is a function $\mathcal{M}_{G_1,\ldots ,G_m}\colon \mathcal{C}^{i_1}\times\mathcal{C}^{i_2}\times \ldots \times \mathcal{C}^{i_m} \to \mathcal{C}^{j}$
given by
\begin{equation}
\label{sum}
(\mathbf{H}_{i_1},\ldots,\mathbf{H}_{i_m}) \xrightarrow[]{\mathcal{M}} \mathbf{K}_{j}=
\mathcal{M}_{G_1,\ldots,G_m}(\mathbf{H}_{i_1},\ldots,\mathbf{H}_{i_m}),
\end{equation}
where $G_k \colon C^{i_k}(\CCX)\to C^{j}(\CCX), k=1,\ldots,m$,
are cochain maps. We think of $\mathcal{M}$ as a message-passing function that takes into account the messages outputted by maps $G_1,\ldots,G_m$, which collectively act on a cochain vector $(\mathbf{H}_{i_1},\ldots,\mathbf{H}_{i_m})$, to obtain an updated cochain $\mathbf{K}_{j}$.
See Sections~\ref{closer_look} and \ref{relation} for more details.
Figure~\ref{fig:tensor}(c) shows a merge node example. 





\subsection{Push-forward operator and merge node}
\label{closer_look}

We introduce the push-forward operation,
a computational scheme that enables sending a cochain
supported on $i$-cells to $j$-cells.
The push-forward operation
is a computational building block used to formalize
the definition of the merge nodes given in Equation~\ref{sum},
the higher-order message passing introduced in Section~\ref{homp-merge},
and the (un)pooling operations introduced in Section~\ref{pooling-hoans}.

\begin{tcolorbox}
[width=\linewidth, sharp corners=all, colback=white!95!black]
\begin{definition}[Cochain push-forward]
\label{pushing_exact_definition}
Consider a CC $\CCX$,
a cochain map $G\colon\mathcal{C}^i(\CCX)\to \mathcal{C}^j(\CCX)$,
and a cochain $\mathbf{H}_i$ in $\mathcal{C}^i(\CCX)$.
A \textbf{(cochain) push-forward} induced by $G$ is an operator
$\mathcal{F}_G \colon \mathcal{C}^i(\CCX)\to \mathcal{C}^j(\CCX)$ 
defined via
    \begin{equation}
    \mathbf{H}_i \to \mathbf{K}_j=[ \mathbf{k}_{y^j_1},\ldots,\mathbf{k}_{y^j_{|\CCX^j|} }] = \mathcal{F}_G(\mathbf{H}_i), 
\end{equation}
such that for $k=1,\ldots,|\CCX^j|$,
    \begin{equation}
    \label{functional}
    \mathbf{k}_{y_k^j}= \bigoplus_{x_l^i \in \mathcal{N}_{G^T(y_k^j)}} \alpha_{G} ( \mathbf{ \mathbf{h}_{x_l^i}}  ),
    \end{equation}
where $\bigoplus$ is a permutation-invariant aggregation function and $\alpha_G$ is a differentiable function. 
\end{definition}
\end{tcolorbox}

The operator $\mathcal{F}_{G}$ pushes forward an $i$-cochain $\mathbf{H}_i$ supported on $\CCX^i$ to a $j$-cochain $\mathcal{F}_{G}(\mathbf{H}_i)$ supported on $\CCX^j $.
For every cell $ y \in \CCX^j$,
Equation~\ref{functional} constructs the vector $\mathbf{k}_y$
by aggregating all vectors $\mathbf{h}_x$ attached to the neighbors $x \in \CCX^i$ of $y$
with respect to the neighborhood function $\mathcal{N}_{G^T}$, and
by then applying a differentiable function $\alpha_G$
on the set of aggregated vectors $\{ \mathbf{h}_x| x\in \mathcal{N}_{G^T}(y)\}$.

\begin{figure}[!t]
\begin{center}
\includegraphics[scale = 0.12, keepaspectratio = 1]{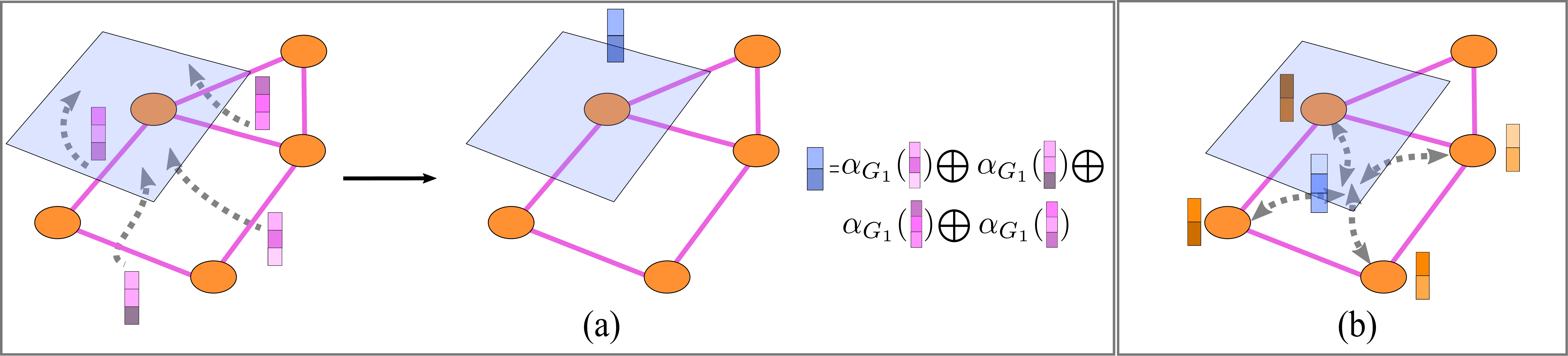}
\end{center}
\caption{Examples of push-forward operators.
 (a): Let $G_1\colon \mathcal{C}^1\to \mathcal{C}^2$ be a cochain map. A push-forward $\mathcal{F}_{G_1}$ induced by $G_1$ takes as input a 1-cochain $\textbf{H}_{1}$ defined on the edges of the underlying CC $\mathcal{X}$ and `pushes-forward' this cochain to a 2-cochain $\mathbf{K}_2$ defined on $\mathcal{X}^2$. The cochain $\mathbf{K}_2$ is formed by aggregating the information in $\mathbf{H}_1$ using the neighborhood function $\mathcal{N}_{G_1^T}$.
 In this case, the neighbors of the 2-rank (blue) cell with respect to $G_1$ are the four (pink) 
 edges on the boundary of this cell.
 (b): Similarly, $G_2\colon \mathcal{C}^0\to \mathcal{C}^2$ induces a push-forward map  $\mathcal{F}_{G_2}\colon \mathcal{C}^0\to \mathcal{C}^2$ that sends a 0-cochain $\mathbf{H}_0$ to a 2-cochain $\mathbf{K}_2$.
 The cochain $\mathbf{K}_2$ is defined
 by aggregating the information in $\mathbf{H}_0$
 using the neighborhood function $\mathcal{N}_{G_2^T}$.}
\label{fig:push_forward}
\end{figure}

Figure~\ref{fig:push_forward} visualizes two examples of push-forward operators.
Example~\ref{non-trainable-pushforward} provides a push-forward function
induced by an indicence matrix.
The push-forward function in Example~\ref{non-trainable-pushforward}
does not contain any parameters, therefore it is not trainable.
In Section~\ref{CCCN},
we give examples of parameterized push-forward operations,
whose parameters can be learnt.

\begin{example}
\label{non-trainable-pushforward}
Consider a CC $\mathcal{X}$ of dimension 2.
Let $B_{0,2}\colon \mathcal{C}^2 (\mathcal{X})\to \mathcal{C}^0 (\mathcal{X})$
be an incidence matrix.
The function $\mathcal{F}^{m}_{B_{0,2}}\colon\mathcal{C}^2 (\mathcal{X})\to \mathcal{C}^0 (\mathcal{X})$
defined by $\mathcal{F}^{m}_{B_{0,2}}(\mathbf{H_{2}})= B_{0,2} (\mathbf{H}_{2})$ is a push-forward induced by $B_{0,2}$.
$\mathcal{F}^{m}_{B_{0,2}}$ pushes forward the cochain $ \mathbf{H}_{2}\in \mathcal{C}^2$
to cochain $ B_{0,2} (\mathbf{H}_{2}) \in \mathcal{C}^0$.   
\end{example}

\begin{figure}[!t]
	\begin{center}
	\includegraphics[scale = 0.20, keepaspectratio = 1]{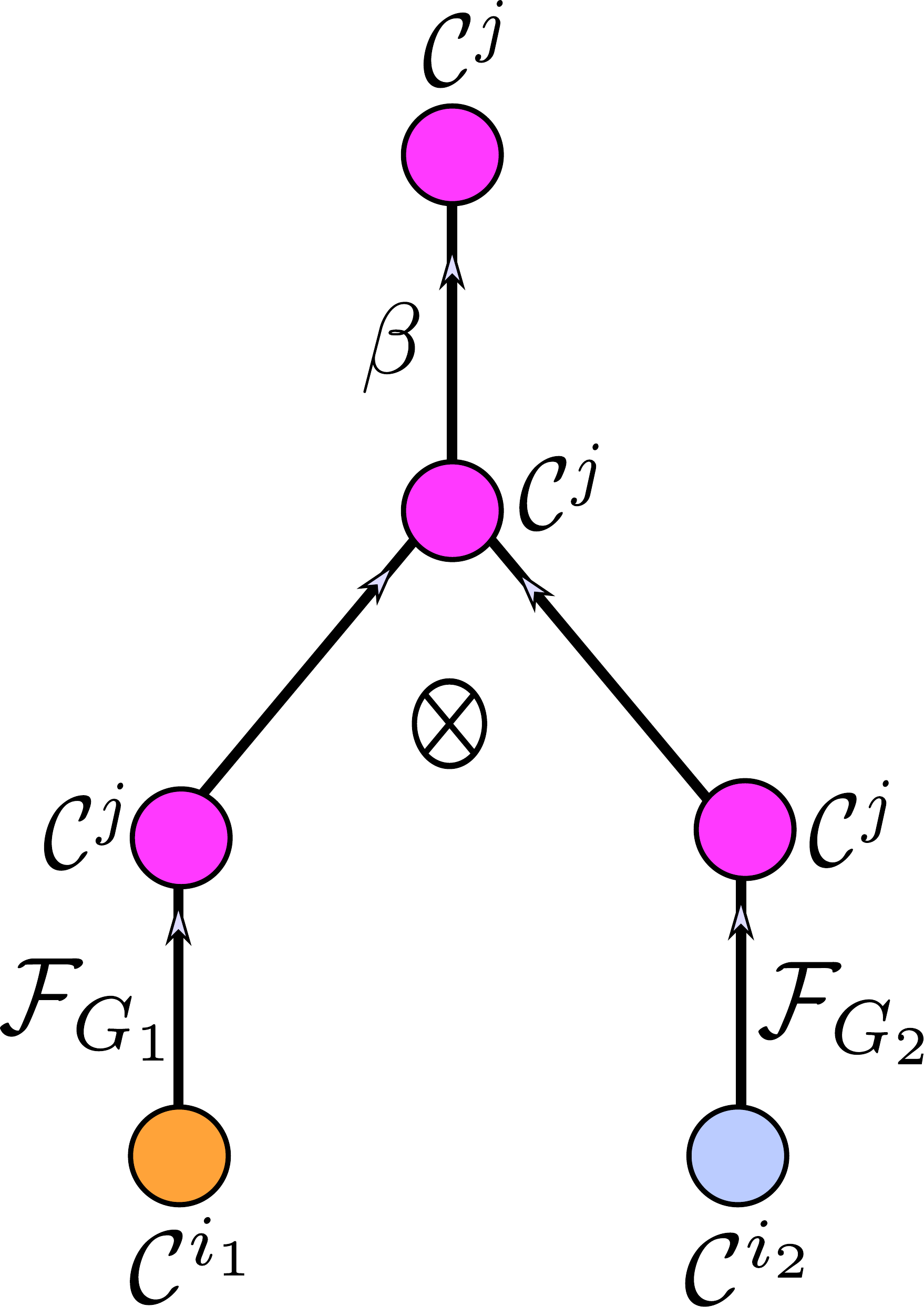}
	\end{center}
\caption{A tensor diagram
depicting
Definition~\ref{exact_definition_merge_node}
of merge node.}
	\label{fig:merge_node}
\end{figure}

In Definition~\ref{exact_definition_merge_node},
we formulate the notion of merge node
using push-forward operators.
Figure~\ref{fig:merge_node} visualizes
Definition~\ref{exact_definition_merge_node}
of merge node via a tensor diagram.

\begin{tcolorbox}
[width=\linewidth, sharp corners=all, colback=white!95!black]
\begin{definition}[Merge node]
\label{exact_definition_merge_node}
Let $\CCX$ be a CC.
Moreover, let $G_1\colon\mathcal{C}^{i_1}(\mathcal{X})\to\mathcal{C}^j(\mathcal{X})$ and
$G_2\colon\mathcal{C}^{i_2}(\mathcal{X})\to\mathcal{C}^j(\mathcal{X})$ be two cochain maps.
Given a cochain vector $(\mathbf{H}_{i_1},\mathbf{H}_{i_2}) \in \mathcal{C}^{i_1}\times \mathcal{C}^{i_2}$, a \textbf{merge node} $\mathcal{M}_{G_1,G_2}\colon\mathcal{C}^{i_1} \times \mathcal{C}^{i_2} \to \mathcal{C}^j$ is defined as
\begin{equation}
    \mathcal{M}_{G_1,G_2}(\mathbf{H}_{i_1},\mathbf{H}_{i_2})= \beta\left( \mathcal{F}_{G_1}(\mathbf{H}_{i_1})  \bigotimes \mathcal{F}_{G_2}(\mathbf{H}_{i_2}) \right),
\end{equation}
where $\bigotimes \colon \mathcal{C}^j \times \mathcal{C}^j \to \mathcal{C}^j $ is an aggregation function, $\mathcal{F}_{G_1}$ and $\mathcal{F}_{G_2}$ are push-forward operators induced by $G_1$ and $G_2$, and $\beta$ is an activation function.
\end{definition}
\end{tcolorbox}

\subsection{Three tensor operations for the construction of CCNNs}
\label{sec:three}

Any tensor diagram representation of a CCNN can be built from two elementary operations:
the push-forward operator and the merge node.
In practice, it is convenient to introduce other operations
that facilitate building involved neural network architectures more effectively.
For example, one useful operation
is the dual operation of the merge node, which we call the split node.

\begin{tcolorbox}
[width=\linewidth, sharp corners=all, colback=white!95!black]
\begin{definition} [Split node]
 \label{exact_definition_split_node}
Let $\CCX$ be a CC.
Moreover, let $G_1\colon\mathcal{C}^{j}(\mathcal{X})\to\mathcal{C}^{i_1}(\mathcal{X})$
and $G_2\colon\mathcal{C}^{j}(\mathcal{X})\to\mathcal{C}^{i_2}(\mathcal{X})$ be two cochain maps. Given a cochain $\mathbf{H}_{j} \in \mathcal{C}^{j}$, a \textbf{split node} $\mathcal{S}_{G_1,G_2}\colon\mathcal{C}^j \to \mathcal{C}^{i_1} \times \mathcal{C}^{i_2} $ is defined as
\begin{equation}
    \mathcal{S}_{G_1,G_2}(\mathbf{H}_{j})= \left(  \beta_1(\mathcal{F}_{G_1}(\mathbf{H}_{j})) , \beta_2(\mathcal{F}_{G_2}(\mathbf{H}_{j})) \right),
\end{equation}
where  $\mathcal{F}_{G_i}$ is a push-forward operator induced by $G_i$,
and $\beta_i$ is an activation function for $i=1, 2$.
\end{definition}
\end{tcolorbox}

While it is clear from Definition~\ref{exact_definition_split_node}
that split nodes are simply tuples of push-forward operations,
using split nodes allows us to build neural networks more effectively and intuitively.
Definition~\ref{def:elem_opers}
puts forward a set of elementary tensor operations,
including split nodes,
to facilitate the formulation of CCNNs in terms of tensor diagrams.

\begin{tcolorbox}
[width=\linewidth, sharp corners=all, colback=white!95!black]
\begin{definition}[Elementary tensor operations]
\label{def:elem_opers}
We refer collectively to push-forward operations, merge nodes and split nodes
as \textbf{elementary tensor operations}.
\end{definition}
\end{tcolorbox}

\begin{figure}[t!]
\begin{center}
\includegraphics[scale = 0.165, keepaspectratio = 0.20]{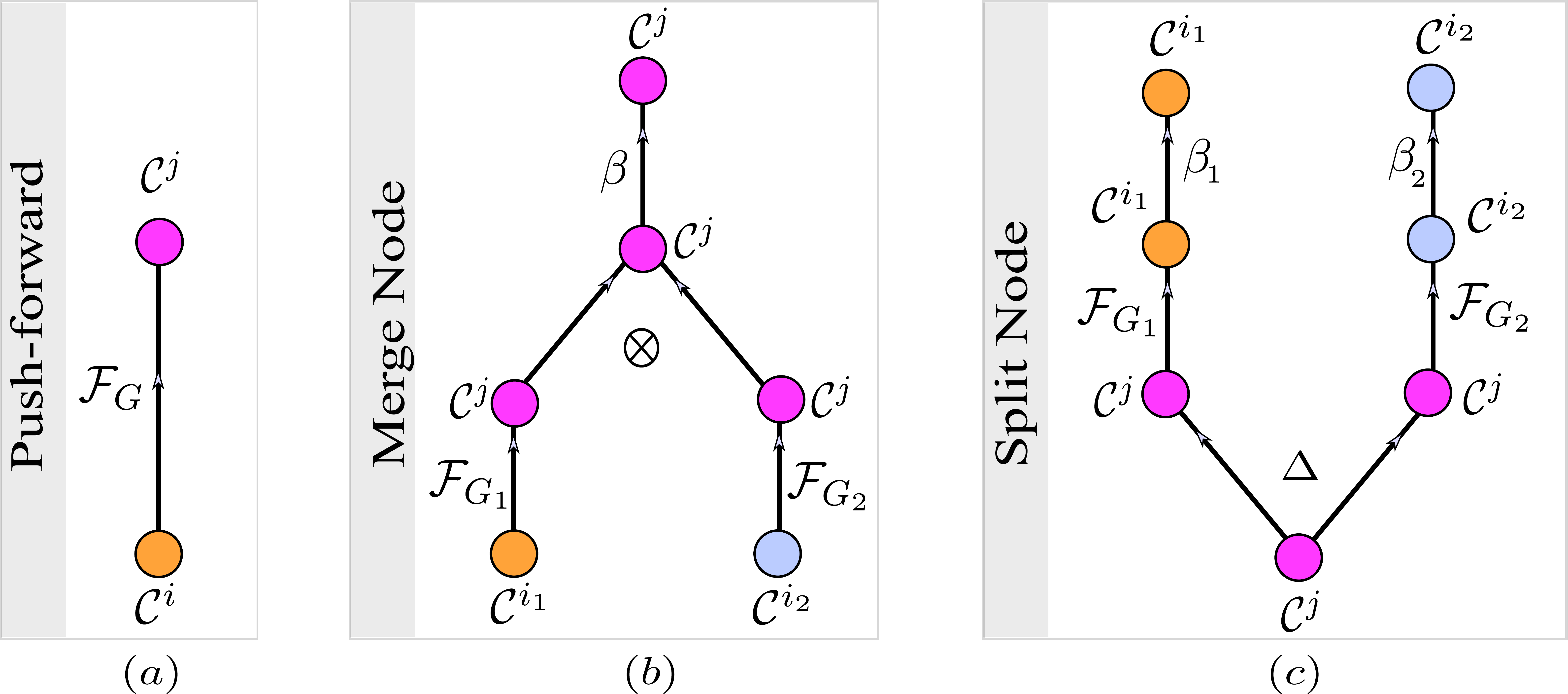}
\end{center}
\caption{Tensor diagrams of the elementary tensor operations,
namely of push-forward operations, merge nodes and split nodes.
These three elementary tensor operations are building blocks
for constructing tensor diagrams of general CCNNs. 
A general tensor diagram can be formed using compositions
and horizontal concatenations of the three elementary tensor operations.
(a): A tensor diagram of a push-forward operation induced by a cochain map $G\colon\mathcal{C}^i \to \mathcal{C}^j$.
(b): A merge node induced by two cochain maps
$G_1\colon\mathcal{C}^{i_1} \to \mathcal{C}^j$ and
$G_2\colon\mathcal{C}^{i_2} \to \mathcal{C}^j$.
(c): A split node induced by two cochain maps $G_1\colon\mathcal{C}^{j}\to\mathcal{C}^{i_1}$ and $G_2\colon\mathcal{C}^{j}\to\mathcal{C}^{i_2}$.
In this illustration, the function
$\Delta \colon \mathcal{C}^{j}\to \mathcal{C}^{j}\times \mathcal{C}^{j}$
is defined as $\Delta(\mathbf{H}_j)= (\mathbf{H}_j,\mathbf{H}_j)$.}
\label{split_merge_pushforward}
\end{figure}

\begin{figure}[t!]
\begin{center}
\includegraphics[scale = 0.15, keepaspectratio = 0.20]{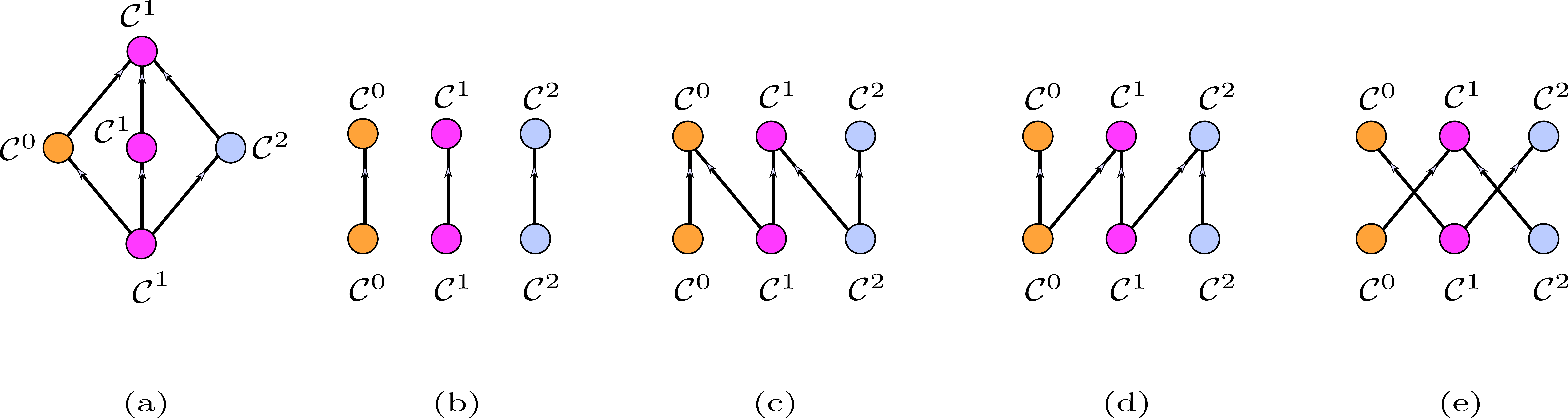}
\end{center}
\caption{Examples of existing neural networks
that can be realized in terms of the elementary operations given in Figure~\ref{split_merge_pushforward}. Edge labels are dropped to simplify exposition.
(a): The simplicial complex net (SCoNe), proposed by~\cite{roddenberry2021principled},
can be realized as a composition of a split node that splits an input 1-cochain to three cochains of dimensions zero, one, and two,
followed by a merge node that merges these cochains into a 1-cochain.
(b): The simplicial neural network (SCN), proposed by~\cite{ebli2020simplicial}, can be realized in terms of push-forward operations.
(c)--(e): Examples of cell complex neural networks (CXNs); see~\cite{hajijcell}. Note that (e) can be realized in terms of a single merge node that merges the 0 and 2-cochains to a 1-cochain as well as a single split node that splits the 1-cochain to 0- and 2-cochains.}
\label{prior_work}
\end{figure}

Figure~\ref{split_merge_pushforward}
displays tensor diagrams of elementary tensor operations,
and Figure~\ref{prior_work}
exemplifies how existing topological neural networks are expressed
via tensor diagrams based on elementary tensor operations.
For example, the simplicial complex net (SCoNe),
a Hodge decomposition-based neural network
proposed by~\cite{roddenberry2021principled},
can be effectively realized in terms of split and merge nodes,
as shown in Figure~\ref{prior_work}(a).

\begin{remark}
The elementary tensor operations constitute the only framework needed to define any parameterized topological neural network. Indeed, since tensor diagrams can be built via the three elementary tensor operations, then it suffices to define the push-forward and the merge operators in order to fully define a parameterized class of CCNNs (recall that the split node is completely determined by the push-forward operator). In Sections~\ref{CCCN} and~\ref{sec:hoans}, we build two parameterized classes of CCNNs: the convolutional and attention classes. In both cases, we only define their corresponding parameterized elementary tensor operations.
Beyond convolutional and attention versions of CCNNs,
the three elementary tensor operations allow us to build arbitrary parameterized tensor diagrams,
therefore providing scope to discover novel topological neural network architectures
on which our theory remains applicable.
\end{remark}

An alternative way of constructing CCNNs draws ideas
from topological quantum field theory (TQFT). In Appendix~\ref{tqft},
we briefly discuss this relationship in more depth.



\subsection{Combinatorial complex convolutional networks (CCCNNs)}
\label{CCCN}

One of the fundamental computational requirements for deep learning on higher-order 
domains is the ability to define and compute convolutional operations.
Here, we introduce CCNNs equipped with convolutional operators, which we call
\textit{combinatorial complex convolutional neural networks (CCCNNs)}.
In particular, we put forward two convolutional operators for CCCNNs:
CC-convolutional push-forward operators and
CC-convolutional merge nodes.

We demonstrate how CCCNNs can be introduced from the two basic blocks: the
push-forward and the merge operations which have been defined abstractly in Section~\ref{closer_look}.
In its simplest form, a CC-convolutional push-forward,
as conceived in Definition~\ref{def:cc_conv_pushforward},
is a generalization of the convolutional graph neural network
introduced in~\cite{kipf2016semi}.

\begin{tcolorbox}
[width=\linewidth, sharp corners=all, colback=white!95!black]
\begin{definition}[CC-convolutional push-forward]
\label{def:cc_conv_pushforward}
Consider a CC $\mathcal{X}$,
a cochain map $G\colon \mathcal{C}^i (\mathcal{X})
\to \mathcal{C}^j(\mathcal{X})$,
and a cochain $\mathbf{H}_i \in C^i(\mathcal{X}, \mathbb{R}^{{s}_{in}})$.
A \textbf{CC-convolutional push-forward} is a cochain map $\mathcal{F}^{conv}_{G;W} \colon C^i(\mathcal{X}, \mathbb{R}^{{s}_{in}}) \to C^j(\mathcal{X}, \mathbb{R}^{{t}_{out}}) $ defined as
\begin{equation}
\label{CC-conv-push-forward}
 \mathbf{H}_i \to  \mathbf{K}_j=  G \mathbf{H}_i W ,
\end{equation}
where $ W \in \mathbb{R}^{d_{s_{in}}\times d_{s_{out}}} $ are trainable parameters.
\end{definition}
\end{tcolorbox}

Having defined the CC-convolutional push-forward,
the CC-convolutional merge node (Definition~\ref{CC-convolutional}) is a
straightforward application of Definition \ref{exact_definition_merge_node}.
Variants of Definition~\ref{CC-convolutional} 
have appeared in recent works on higher-order 
networks~\cite{bunch2020simplicial, schaub2020random, schaub2021signal, roddenberry2021principled,hajijcell,roddenberry2021signal,ebli2020simplicial,hajij2021simplicial,calmon2022higher,yang2023convolutional}. 

\begin{tcolorbox}
[width=\linewidth, sharp corners=all, colback=white!95!black]
\begin{definition}[CC-convolutional merge node]
\label{CC-convolutional}
Let a $\CCX$ be a CC.
Moreover, let $G_1\colon\mathcal{C}^{i_1}(\mathcal{X})
\to\mathcal{C}^j(\mathcal{X})$
and $G_2\colon\mathcal{C}^{i_2}(\mathcal{X})\to\mathcal{C}^j(\mathcal{X})$
be two cochain maps. Given a cochain vector $(\mathbf{H}_{i_1},\mathbf{H}_{i_2}) \in \mathcal{C}^{i_1}\times \mathcal{C}^{i_2}$,
a \textbf{CC-convolutional merge node} $\mathcal{M}^{conv}_{\mathbf{G};\mathbf{W}}
\colon\mathcal{C}^{i_1} \times \mathcal{C}^{i_2} \to \mathcal{C}^j$ is defined as
\begin{equation}
\begin{aligned}
\mathcal{M}^{conv}_{\mathbf{G};\mathbf{W}}
(\mathbf{H}_{i_1},\mathbf{H}_{i_2}) &=
\beta\left( \mathcal{F}^{conv}_{G_1;W_1}(\mathbf{H}_{i_1})
+ \mathcal{F}^{conv}_{G_2;W_2}(\mathbf{H}_{i_2})  \right)\\
&= \beta ( G_1 \mathbf{H}_{i_1} W_1  +  G_2 \mathbf{H}_{i_2} W_2 ),\\
\end{aligned}
\end{equation}
where
$\mathbf{G}=(G_1, G_2)$,
$\mathbf{W}=(W_1, W_2)$ is a tuple of trainable parameters, and
$\beta$ is an activation function.
\end{definition}
\end{tcolorbox}

In practice, the matrix representation of the cochain map $G$
in Definition~\ref{def:cc_conv_pushforward}
might require problem-specific normalization during training.
For various types of normalization
in the context of higher-order convolutional operators,
we refer the reader to~\cite{kipf2016semi,bunch2020simplicial,schaub2020random}.

We have used the notation
$\CCN_{\mathbf{G}}$
for a tensor diagram and its corresponding CCNN.
Our notation
indicates that the CCNN is composed of
elementary tensor operations
based on a sequence $\mathbf{G}= (G_i)_{i=1}^l$ 
of cochain maps defined on the underlying CC.
When the elementary tensor operations 
that make up the CCNN are
parameterized by a sequence
$\mathbf{W}= (W_i)_{i=1}^k$
of trainable parameters,
we denote the CCNN and its tensor diagram representation by
$\CCN_{\mathbf{G};\mathbf{W}}$.





\subsection{Combinatorial complex attention neural  networks (CCANNs)}
\label{sec:hoans}

The majority of higher-order deep learning models focus on layers that use \textit{isotropic aggregation}, which means that neighbors in the vicinity of an element contribute equally to an update of the element's representation. As information is aggregated in a diffusive manner, such isotropic aggregation can limit the expressiveness of these learning models, leading to phenomena such as oversmoothing~\cite{beaini2021directional}. In contrast, \textit{attention-based learning}~\cite{choi2017gram} allows deep learning models to assign a probability distribution to neighbors in the local vicinity of elements in the underlying domain,
thus highlighting components
with the most task-relevant information~\cite{velickovic2017graph}. 
Attention-based models are successful in practice as they ignore noise in the domain, thereby improving the signal-to-noise ratio~\cite{boaz2019,mnih2014recurrent}. 
Accordingly, attention-based models have achieved remarkable success on
traditional machine learning tasks on graphs,
including node classification and link prediction~\cite{li2021learning},
node ranking~\cite{sun2009rankclus},
and attention-based embeddings~\cite{choi2017gram,lee2018graph}.

After introducing the CC-convolutional push-forward in Section~\ref{CCCN},
our second example of a push-forward operation
is the CC-attention push-forward,
which is introduced in the present section.
Thus, it becomes possible to use CCNNs equipped with CC-attention push-forward operators,
which we call \textit{combinatorial complex attention neural networks (CCANNs)}.
We first provide the general notion of attention of a sub-CC $\mathcal{Y}_0$ of a CC
with respect to other sub-CCs in the CC.

\begin{tcolorbox}
[width=\linewidth, sharp corners=all, colback=white!95!black]
\begin{definition}[Higher-order attention]
\label{hoa}
Let $\CCX$ be a CC, $\mathcal{N}$
a neighborhood function defined on $\CCX$, and $\CCY_0$ a sub-CC of $\CCX$.
Let $\mathcal{N}(\mathcal{Y}_0)=
\{ \CCY_1,\ldots, \CCY_{|\mathcal{N}(\CCY_0)|} \}$ be a set
of sub-CCs that are in the vicinity of $\mathcal{Y}_0$ with respect to the neighborhood function $\mathcal{N}$.
A \textbf{higher-order attention}
of $\CCY_0$ with respect to $\mathcal{N}$
is a function
$a\colon {\CCY_0}\times \mathcal{N}(\CCY_0)\to [0,1]$
that assigns a weight $a(\CCY_0, \CCY_i)$
to each element $\CCY_i\in\mathcal{N}(\CCY_0)$
such that $\sum_{i=1}^{| \mathcal{N}(\CCY_0)|} a(\CCY_0,\CCY_i)=1$.     
\end{definition}
\end{tcolorbox}

As seen from Definition~\ref{hoa},
a higher-order attention of a sub-CC $ \mathcal{Y}_0 $ 
with respect to a neighborhood function $\mathcal{N}$
assigns a discrete distribution to the neighbors of $\mathcal{Y}_0$.
Attention-based learning typically aims to learn the function $a$. Observe that the function $a$ relies on the neighborhood function $\mathcal{N}$. In our context, we aim to learn the function $a$ whose neighborhood function is either an incidence or a (co)adjacency function, as introduced in Section~\ref{sec:neighborhood_structure}.

Recall from Definition~\ref{hoa} that a weight
$a(\mathcal{Y}_0,\mathcal{Y}_i)$ requires
both a source sub-CC $\mathcal{Y}_0$
and a target sub-CC $\mathcal{Y}_i$ as inputs.
Thus, a CC-attention push-forward operation
requires two cochain spaces.
Definition~\ref{hoan_sym} introduces
a notion of CC-attention push-forward in which
the two underlying cochain spaces
contain cochains supported on cells of equal rank.

\begin{tcolorbox}
[width=\linewidth, sharp corners=all, colback=white!95!black]
\begin{definition}[CC-attention push-forward for cells of equal rank]
	\label{hoan_sym}
	Let $G\colon C^{s}(\CCX)\to C^{s}(\CCX)$ be a neighborhood matrix. A \textbf{CC-attention push-forward}
	induced by $G$ is a cochain map $\mathcal{F}^{att}_{G}\colon C^{s}(\CCX, \mathbb{R}^{d_{s_{in}}}) \to C^{s}(\CCX,\mathbb{R}^{d_{s_{out}}}) $ defined as
	\begin{equation}
	\label{attention1}
	\mathbf{H}_s \to \mathbf{K}_{s} = 
	(G \odot att)  \mathbf{H}_{s}  W_{s} ,
	\end{equation}
 where $\odot$ is the Hadamard product, $W_{s}  \in \mathbb{R}^{d_{s_{in}}\times d_{s_{out}}} $ are trainable
	parameters, and $att\colon C^{s}(\CCX)\to C^{s}(\CCX) $ is a \textbf{higher-order attention matrix} that has the same dimension as matrix $G$. The $(i,j)$-th entry of matrix $att$ is defined as
	\begin{equation}
	att(i,j) =  \frac{e_{ij}}{ \sum_{k \in \mathcal{N}_{G}(i) e_{ik} } },    
	\end{equation}
	where $e_{ij}= \phi(a^T [W_{s} \mathbf{H}_{s,i}||W_{s} \mathbf{H}_{s,j} ] )$,
	$a \in \mathbb{R}^{2 \times s_{out}} $ is a trainable vector,  $[a ||b ]$ denotes the concatenation of $a$ and $b$, $\phi$ is an activation function, and $\mathcal{N}_{G}(i)$ is the neighborhood of cell $i$ with respect to matrix $G$. 
\end{definition}
\end{tcolorbox}

Definition~\ref{hoan_asym} treats a more general case than Definition~\ref{hoan_sym}. Specifically, Definition~\ref{hoan_asym}
introduces a notion of CC-attention push-forward in which the two underlying cochain spaces
contain cochains supported on cells of different ranks.

\begin{tcolorbox}
[width=\linewidth, sharp corners=all, colback=white!95!black]
\begin{definition}[CC-attention push-forward for cells of unequal ranks]
\label{hoan_asym}
For $s\neq t$,
let $G\colon C^{s}(\CCX)\to C^{t}(\CCX)$ be a neighborhood matrix. A \textbf{CC-attention block} induced by $G$ is a cochain map $\mathcal{F}_{G}^{att}  {\mathcal{A}}\colon C^{s}(\CCX,\mathbb{R}^{d_{s_{in}}}) \times C^{t}(\CCX,\mathbb{R}^{d_{t_{in}}}) \to C^{t}(\CCX,\mathbb{R}^{d_{t_{out}}}) \times C^{s}(\CCX,\mathbb{R}^{d_{s_{out}}}) $ defined as
	\begin{equation}
	(\mathbf{H}_{s},\mathbf{H}_{t}) \to  (\mathbf{K}_{t}, \mathbf{K}_{s} ), 
	\end{equation} 
	with
	\begin{equation}
	\label{attention2}
	\mathbf{K}_{t} =   ( G \odot att_{s\to t})  \mathbf{H}_{s} W_{s} ,\;
        \mathbf{K}_{s} = (G^T \odot att_{t\to s})  \mathbf{H}_{t}  W_{t} , 
	\end{equation}
	where $W_s \in \mathbb{R}^{d_{s_{in}}\times d_{t_{out}}} , W_t \in \mathbb{R}^{d_{t_{in}}\times d_{s_{out}}} $ are
	trainable
	parameters, and $att_{s\to t}^{k}\colon C^{s}(\CCX)\to C^{t}(\CCX) , att_{t\to s}^{k}\colon C^{t}(\CCX)\to C^{s}(\CCX) $ are \textbf{higher-order attention matrices} that have the same dimensions as matrices $G$ and $G^T$, respectively. The $(i,j)$-th entries of matrices  $att_{s\to t}$ and $att_{t\to s}$ are defined as
	\begin{equation}
\label{eq:ast_ats}
	(att_{s\to t})_{ij} =  \frac{e_{ij}}{ \sum_{k \in \mathcal{N}_{G} (i) e_{ik} } },\;
 (att_{t\to s})_{ij} =  \frac{f_{ij}}{ \sum_{k \in \mathcal{N}_{G^T} (i) f_{ik} } },
	\end{equation} 
	with
	\begin{equation}
 \label{eq:ef}
	e_{ij} = \phi((a)^T [W_s \mathbf{H}_{s,i}||W_t \mathbf{H}_{t,j}] ),\;
	f_{ij} = \phi(rev(a)^T [W_t \mathbf{H}_{t,i}||W_s \mathbf{H}_{s,j}]),
	\end{equation}
	where $a \in \mathbb{R}^{t_{out} + s_{out}} $ is a trainable vector, and
	$rev(a)= [ a^l[:t_{out}]||a^l[t_{out}:] ].$
\end{definition}
\end{tcolorbox}

\begin{figure}[ht]
\begin{center}
\includegraphics[scale = 0.20, keepaspectratio = 1]{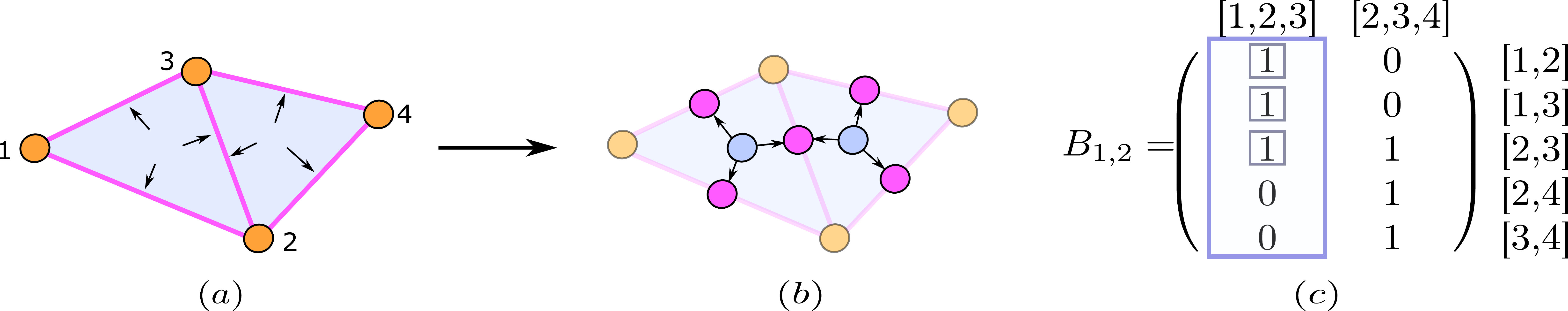}
\end{center}
\caption{Illustration of notion of CC-attention for cells of unequal ranks.
(a): A CC. Each $2$-cell (blue face) of the CC attends to its incident $1$-cells (pink edges).
(b): The attention weights reside on a graph constructed from the cells and their incidence relations; see Section~\ref{sec:Hasse} for details.
(c): Incidence matrix $B_{1,2}$ of the CC given in (a). The non-zero elements in column $[1,2,3]$ correspond to the neighborhood $\mathcal{N}_{B_{1,2}}([1,2,3])$ of $[1,2,3]$ with respect to $B_{1,2}$.}
\label{fig:B12}
\end{figure}

The incidence matrices of Definition~\ref{def:inc_mat}
can be employed as neighborhood matrices in Definition~\ref{hoan_asym}.
In Figure~\ref{fig:B12},
we illustrate the notion of CC-attention for incidence neighborhood matrices.
Figure~\ref{fig:B12}(c) shows the non-squared incidence matrix $B_{1,2}$
associated with the CC displayed in Figure~\ref{fig:B12}(a).
The attention block $\HB_{B_{1,2}}$ learns two incidence matrices $att_{s\to t}$ and $att_{t\to s}$. The matrix $att_{s\to t}$ has the same shape as $B_{1,2}$, and non-zero elements exactly where $B_{1,2}$ has elements equal to one.
Each column $i$ in $att_{s\to t}$ represents a probability distribution that defines the attention of the $i$-th 2-cell to its incident 1-cells. The matrix $att_{t\to s}$ has the same shape as $B_{1,2}^T$, and similarly represents the attention of 1-cells to 2-cells.

\begin{remark}
The computation of a push-forward cochain $\mathbf{K}_t$ requires two cochains,
namely $\mathbf{H}_s$ and $\mathbf{H}_t$.
While $\mathbf{K}_t$ depends on $\mathbf{H}_s$
directly in Equation~\ref{attention2},
it depends on $\mathbf{H}_t$
only indirectly via $att_{s\to t}$ as seen from
Equations~\ref{attention2}--\ref{eq:ef}.
Moreover, the cochain $\mathbf{H}_t$ is only needed for $att_{s\to t}$ 
during training, and not during inference.
In other words, the computation of a CC-attention push-forward block
requires only a single cochain, $\mathbf{H}_s$, during inference,
in agreement with the computation of a general cochain push-forward
as specified in Definition~\ref{pushing_exact_definition}.
\end{remark}

The operators $G \odot att$ and $G \odot att_{s\to t}$
in Equations~\ref{attention1} and~\ref{attention2}
can be viewed as learnt attention versions of $G$.
This perspective allows to employ CCANNs
to learn arbitrary types of discrete exterior calculus operators,
as elaborated in Appendix~\ref{linear}.



\section{Higher-order message passing}
\label{homp-merge}

In this section, we explain the relation between the notion of the merge node introduced in Section~\ref{closer_look} and higher-order message passing. In particular, we prove that higher-order message passing on CCs can be realized in terms of the elementary tensor operations introduced in Section~\ref{sec:three}. Further, we demonstrate the connection between CCANNs (Section~\ref{sec:hoans}) and higher-order message passing, and introduce an attention version of higher-order message passing. We first define higher-order message passing on CCs,
generalizing notions introduced in~\cite{hajijcell}. 

We remark that many of the constructions discussed here are presented in their most basic form, but can be extended further. 
An important aspect in this direction is the construction of message-passing protocols that are invariant or equivariant with respect to the action of a specific group. 

\subsection{Definition of higher-order message passing}
\label{sec:homp}

\begin{figure}[!t]
\begin{center}
\includegraphics[scale = 0.052, keepaspectratio = 1]{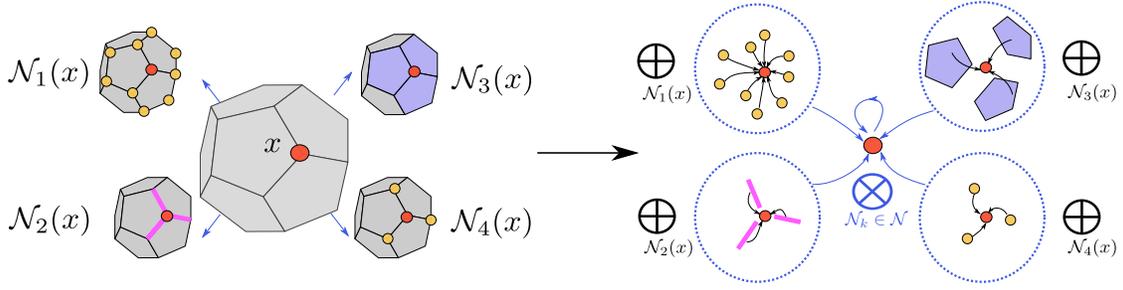}
\end{center}
\caption{An illustration of higher-order message passing (Definition~\ref{homp-definition}).
Left-hand side: a collection of neighborhood functions $\mathcal{N}_1,\ldots,\mathcal{N}_k$ are selected.
The selection typically depends on the learning task.
Right-hand side: for each $\mathcal{N}_k$,
the messages are aggregated using an intra-neighborhood function $\bigoplus$.
The inter-neighborhood function $\bigotimes$
aggregates the final messages obtained from all neighborhoods.}
\label{fig:HOMP}
\end{figure}

Higher-order message passing refers to a computational framework that involves exchanging messages among entities and cells in a higher-order domain using a set of neighborhood functions.
In Definition~\ref{homp-definition},
we formalize the notion of higher-order message passing for CCs.
Figure~\ref{fig:HOMP}
illustrates Definition~\ref{homp-definition}.

\begin{tcolorbox}
[width=\linewidth, sharp corners=all, colback=white!95!black]
\begin{definition}[Higher-order message passing on a CC]
\label{homp-definition}
Let $\CCX$ be a CC.  Let $\mathcal{N}=\{ \mathcal{N}_1,\ldots,\mathcal{N}_n\}$ be a set of neighborhood functions defined on  $\CCX$. Let $x$ be a cell and $y\in \mathcal{N}_k(x)$ for some $\mathcal{N}_k \in \mathcal{N}$. A \textbf{message} $m_{x,y}$ between cells $x$ and $y$
is a computation that depends on these two cells
or on the data supported on them.
Denote by $\mathcal{N}(x)$ the multi-set  $\{\!\!\{ \mathcal{N}_1(x) , \ldots ,  \mathcal{N}_n (x) \}\!\!\}$,
and by $\mathbf{h}_x^{(l)}$ some data supported on the cell $x$ at layer $l$. \textbf{Higher-order message passing} on $\CCX$, induced by $\mathcal{N}$, is defined via the following four update rules:
\begin{align}
\label{eqn:homp0}
m_{x,y} &= \alpha_{\mathcal{N}_k}(\mathbf{h}_x^{(l)},\mathbf{h}_y^{(l)}), \\ \label{eqn:homp1}
m_{x}^k &=  \bigoplus_{y \in \mathcal{N}_k(x)}  m_{x,y}, \; 1\leq k \leq n,   \\ \label{eqn:homp2}
m_{x} &=  \bigotimes_{ \mathcal{N}_k \in \mathcal{N} } m_x^k, \\ \label{eqn:homp3}
\mathbf{h}_x^{(l+1)} &= \beta (\mathbf{h}_x^{(l)}, m_x). 
\end{align}
Here, $\bigoplus$ is a permutation-invariant aggregation function
called the \textbf{intra-neighborhood} of $x$,
$\bigotimes$ is an aggregation function
called the \textbf{inter-neighborhood} of $x$,
and $\alpha_{\mathcal{N}_k},\beta$ are differentiable functions.
\end{definition}
\end{tcolorbox}

Some remarks on Definition~\ref{homp-definition} are as follows.
First, the message $m_{x,y}$ in Equation~\ref{eqn:homp0}
does not depend only on the data $\mathbf{h}_x^{(l)}$, $\mathbf{h}_y^{(l)}$ supported on the cells $x, y$;
it also depends on the cells themselves.
For instance, if $\CCX$ is a cell complex, the \textit{orientation} of both $x$ and $y$ factors into the computation of message $m_{x,y}$. Alternatively, $x\cup y$ or $x\cap y$ might be cells in $\CCX$ and it might be useful to include their data in the computation of message $m_{x,y}$. This unique characteristic only manifests in higher-order domains,
and does not occur in graphs-based message-passing frameworks~\cite{bronstein2021geometric,gilmer2017neural}\footnote{The message `direction' in $m_{x,y}$ is from $y$ to $x$. In general, $m_{x,y}$ and $m_{y,x}$ are not equal.}.
Second, higher-order message passing relies
on the choice of a set $\mathcal{N}$ of neighborhood functions. This is also a unique characteristic that only occurs in a higher-order domain, where a neighborhood function is necessarily described by a set of neighborhood relations rather than graph adjacency as in graph-based message passing.
Third, in Equation~\ref{eqn:homp0}, since $y$ is implicitly defined with respect to a neighborhood relation $\mathcal{N}_k \in \mathcal{N},$ the function $\alpha_{\mathcal{N}_k}$ and the message $m_{x,y}$ depend on $\mathcal{N}_k$.
Fourth, the inter-neighborhood $\bigotimes$ does not necessarily have to be a permutation-invariant aggregation function. For instance,
it is possible to set an order on the multi-set
$\mathcal{N}(x)$
and compute $m_x$ with respect to this order.
Finally, higher-order message passing relies on two aggregation functions,
the intra-neighborhood and inter-neighborhood,
whereas graph-based message passing relies on a single aggregation function.
The choice of set $\mathcal{N}$,
as illustrated in Section~\ref{sec:CC},
enables the use of a variety of neighborhood functions in higher-order message passing.

\begin{remark}
The push-forward operator given in Definition~\ref{pushing_exact_definition}
is related to the update rule of Equation~\ref{eqn:homp0}. On one hand, Equation~\ref{eqn:homp0} requires two cochains $\mathbf{X}_i= [\mathbf{h}_{x^i_1}^{(l)},\ldots,\mathbf{h}_{x^i_{|\CCX^i|}}^{(l)}]$  and $\mathbf{Y}_{j}^{(l)}=
[\mathbf{h}_{y^{j}_1}^{(l)},\ldots,\mathbf{h}_{y^{j}_{|\CCX^{j}|}}^{(l)}]$ to compute 
$\mathbf{X}^{(l+1)}_i = [\mathbf{h}_{x^i_1}^{(l+1)},\ldots,\mathbf{h}_{x^i_{|\CCX^i|}}^{(l+1)}] $, so signals on both $\mathcal{C}^j$ and $\mathcal{C}^i$ must be present in order to execute Equation~\ref{eqn:homp0}. From this perspective, it is natural and customary to think about this operation as an update rule. On the other hand, the push-forward operator
of Definition~\ref{pushing_exact_definition}
computes a cochain $\mathbf{K}_{j} \in \mathcal{C}^j $ given a cochain $\mathbf{H}_i\in \mathcal{C}^i$. As a single cochain $\mathbf{H}_i$ is required to perform this computation, it is natural to think about Equation~\ref{functional} as a function.
See Section~\ref{merge and message passing} for more details.
\end{remark}

The higher-order message-passing framework given in Definition~\ref{homp-definition} can be used to construct novel neural network architectures on a CC,
as we have also alluded in Figure~\ref{fig:TDL}. First, a CC $\mathcal{X}$ and cochains $\mathbf{H}_{i_1}\ldots, \mathbf{H}_{i_m}$ supported on $\mathcal{X}$ are given. Second, a collection of neighborhood functions are chosen, taking into account the desired learning task. Third, the update rules of Definition~\ref{homp-definition} are executed on the input cochains $\mathbf{H}_{i_1}\ldots, \mathbf{H}_{i_m}$ using the chosen neighborhood functions. The second and the third steps are repeated to obtain the final computations.

\begin{tcolorbox}
[width=\linewidth, sharp corners=all, colback=white!95!black]
\begin{definition}
[Higher-order message-passing neural network]
\label{HMPSNN}
We refer to to any neural network constructed using Definition \ref{homp-definition} as a \textbf{higher-order message-passing neural network.}
\end{definition}
\end{tcolorbox}

\subsection{Higher-order message-passing neural networks are CCNNs}
\label{relation}


In this section,
we show that higher-order message-passing computations
can be realized in terms of merge node computations,
and therefore that
higher-order message-passing neural networks are CCNNs.
As a consequence, higher-order message passing
unifies message passing on
simplicial complexes, cell complexes and hypergraphs
through a coherent set of update rules and, alternatively,
through the expressive language of tensor diagrams.

\begin{theorem}
\label{HMP are CCNNs}
The higher-order message-passing computations of Definition~\ref{homp-definition} can be realized in terms of merge node computations.
\end{theorem}

\begin{proof}
Let $\CCX$ be a CC.  Let $\mathcal{N}=\{ \mathcal{N}_1,\ldots,\mathcal{N}_n\}$ be a set of neighborhood functions as specified in Definition~\ref{homp-definition}. Let $G_k$ be the matrix induced by the neighborhood function $\mathcal{N}_k$. We assume that the cell $x$ given in Definition~\ref{homp-definition} is a $j$-cell and the neighbors $y \in \mathcal{N}_k(x)$ are $i_k$-cells. We will show that Equations~\ref{eqn:homp0}--\ref{eqn:homp3} can be realized as applications of merge nodes. In what follows,
we define the neighborhood function to be $\mathcal{N}_{Id}(x)=\{x\}$ for $x\in \mathcal{X}$. Moreover, we denote the associated neighborhood matrix of $\mathcal{N}_{Id}$ by $Id\colon\mathcal{C}^j\to \mathcal{C}^j$,
as it is the identity matrix. 

Computing message $m_{x,y}$ of Equation~\ref{eqn:homp0}
involves two cochains:
\begin{equation*}
\mathbf{X}_j^{(l)}=
[\mathbf{h}_{x^j_1}^{(l)},\ldots,\mathbf{h}_{x^j_{|\CCX^j|}}^{(l)}],~
\mathbf{Y}_{i_k}^{(l)}=
[\mathbf{h}_{y^{i_k}_1}^{(l)},\ldots,\mathbf{h}_{y^{i_k}_{|\CCX^{i_k}|}}^{(l)}].
\end{equation*}
Every message $m_{x^{^j}_t, y^{i_k}_s }$ corresponds to the entry $[G_k]_{st}$ of matrix $G_k$. In other words, there is a one-to-one correspondence between non-zero entries of matrix $G_k$ and messages $m_{x^{^j}_t, y^{i_k}_s }$.

It follows from Section~\ref{closer_look} that computing $\{m_x^k\}_{k=1}^n$ corresponds to a merge node $\mathcal{M}_{Id_j,G_k}\colon \mathcal{C}^j\times \mathcal{C}^{i_k}\to \mathcal{C}^j $ that performs the computations determined via $\alpha_k$ and $\bigoplus$, and yields 
\begin{equation*}
\mathbf{m}_j^k=[m_{x^j_1}^k,\ldots,m_{x^j_{|\CCX^j|}}^k]=
\mathcal{M}_{Id_j,G_k}(\mathbf{X}_j^{(l)},\mathbf{Y}_{i_k}^{(l)}) \in \mathcal{C}^{j}.
\end{equation*}
At this stage, we have $n$ $j$-cochains $\{\mathbf{m}_j^k\}_{k=1}^n$. Equations~\ref{eqn:homp2} and~\ref{eqn:homp3} merge these cochains with the input $j$-cochain $\mathbf{X}_j^{(l)}$. Specifically, computing $m_x$ in Equation~\ref{eqn:homp2} corresponds to $n-1$ applications of merge nodes of the form $\mathcal{M}_{Id_k,Id_k}\colon\mathcal{C}^j \times \mathcal{C}^j \to \mathcal{C}^j$ on the cochains $\{\mathbf{m}_j^k\}_{k=1}^n$.
Explicitly, we first merge $\mathbf{m}_j^1$ and $\mathbf{m}_j^2$ to obtain $\mathbf{n}_j^1=\mathcal{M}_{Id_j,Id_j}(\mathbf{m}_j^1,\mathbf{m}_j^2)$. Next,
we merge the $j$-cochain $\mathbf{n}_j^1$ with the $j$-cochain $\mathbf{m}_j^3$, and so on. The final merge node in this stage performs the merge $\mathbf{n}_j^{n-1}=\mathcal{M}_{Id_j,Id_j}(\mathbf{n}_j^{n-2},\mathbf{m}_j^n)$, which is $\mathbf{m}_j = [ m_{x_1^j},\ldots, m_{x_{|\mathcal{X}^j|}^j }  ] $\footnote{Recall that while we use the same notation $\mathcal{M}_{Id_k,Id_k}$ for all merge nodes, these nodes have in general different parameters.}. Finally, computing
$\mathbf{X}_j^{(l+1)}$
is realized by a merge node $\mathcal{M}_{(Id_j,Id_j)}(\mathbf{m}_j, \mathbf{X}_j^{(l)})$ whose computations are determined by function $\beta$ of Equation~\ref{eqn:homp3}. 
\end{proof}

Theorem~\ref{HMP are CCNNs} shows that higher-order message-passing networks defined on CCs can be constructed from the elementary tensor operations, and hence they are special cases of CCNNs. We state this result formally in Theorem~\ref{unifying}.

\begin{theorem}
\label{unifying}
 A higher-order message-passing neural network is a CCNN.
\end{theorem}

\begin{proof}
The conclusion follows immediately
from Definition~\ref{HMPSNN} and Theorem~\ref{HMP are CCNNs}.
\end{proof}

It follows from Theorem \ref{unifying} that higher-order message-passing neural networks defined on higher-order domains that are less general than CCs (such as 
simplicial complexes, cell complexes and hypergraphs) are also special cases of CCNNs. Thus, tensor diagrams, as introduced in Definition~\ref{TD}, form a general diagrammatic method for expressing neural networks defined on commonly studied higher-order domains.

\begin{theorem}
\label{unifying1}
Message-passing neural networks defined on simplicial complexes, cell complexes or hypergraphs can be expressed in terms of tensor diagrams and their computations can be realized in terms of the three elementary tensor operators.
\end{theorem}

\begin{proof}
The conclusion follows from Theorem~\ref{unifying} and
from the fact that simplicial complexes, cell complexes and hypergraphs
can be realized as special cases of CCs.
\end{proof}

Theorems~\ref{unifying} and~\ref{unifying1}
put forward a unifying TDL framework based on tensor diagrams,
thus providing scope for future developments. For instance,~\cite{mathilde2023}
have already used our framework to express existing TDL architectures for simplicial complexes, cell complexes and hypergraphs in terms of tensor diagrams.

\subsection{Merge nodes and higher-order message passing:
a qualitative comparison}
\label{merge and message passing}

Higher-order message passing, as
given in Definition~\ref{homp-definition},
provides an update rule to obtain the vector $\mathbf{h}_x^{l+1}$
from vector $\mathbf{h}_x^{l}$ using a set of neighborhood vectors $\mathbf{h}_y^{l}$ determined by $\mathcal{N}(x)$. 
Clearly, this computational framework assumes that vectors $\mathbf{h}_x^{(l)}$ and $\mathbf{h}_{y}^{(l)}$ are provided as inputs. In other words, performing higher-order message passing according to Definition~\ref{homp-definition} requires the cochain  $\mathbf{X}_j^{(l)} \in \mathcal{C}^{j} $ in the target domain as well as the cochains $\mathbf{Y}_{i_k}^{(l)} \in \mathcal{C}^{i_k} $ in order to compute the updated $j$-cochain $\mathbf{X}_j^{(l+1)}$. On the other hand,
performing a merge node computation
requires a cochain vector $(\mathbf{H}_{i_1},\mathbf{H}_{i_2})$,
as seen from Equation~\ref{sum} and Definition~\ref{exact_definition_merge_node}.

\begin{figure}[!t]
\begin{center}
\includegraphics[scale = 0.2, keepaspectratio = 0.20]{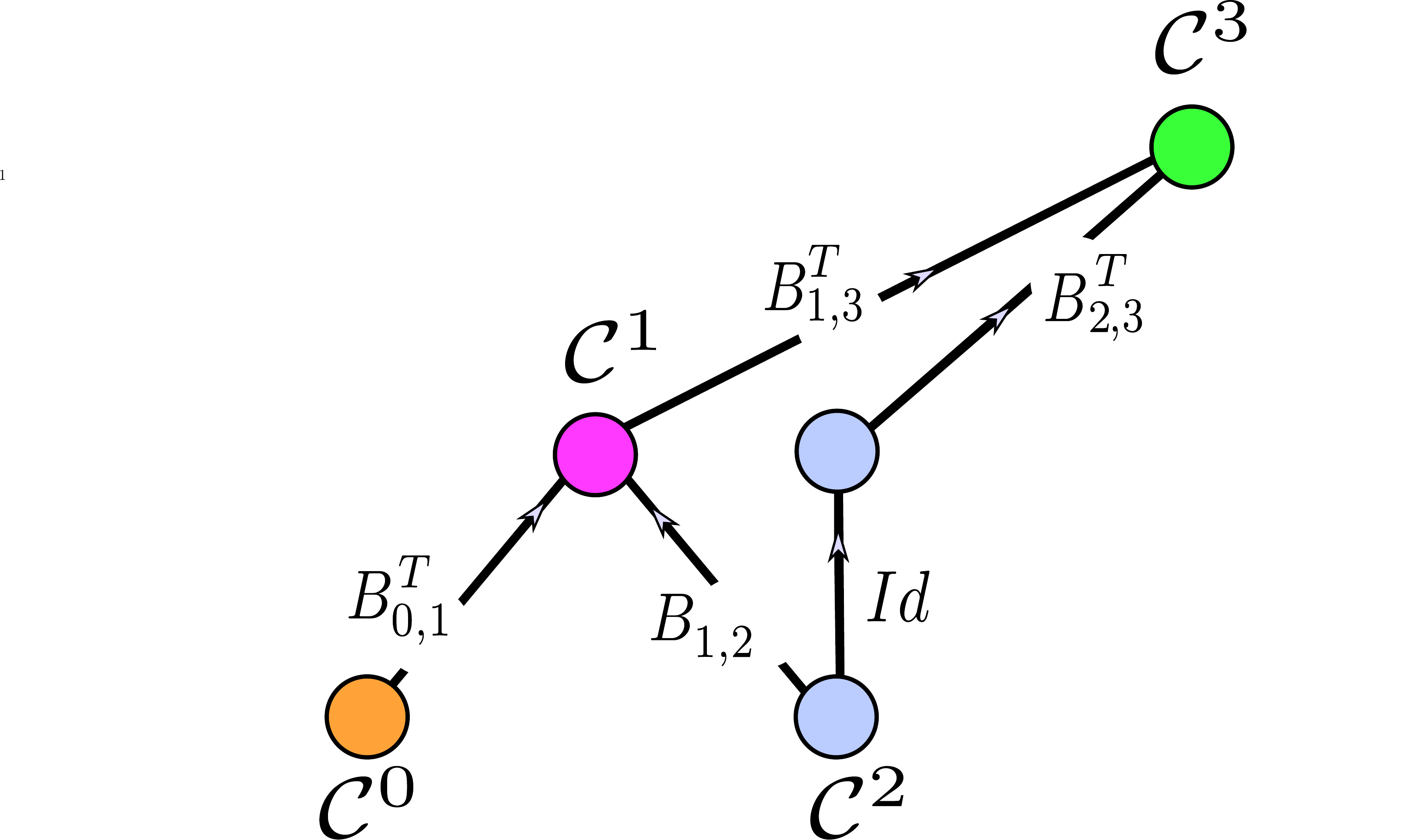}
\end{center}
\caption{The depicted neural network can be realized as a composition of two merge nodes. More specifically, the input is a cochain vector $(\mathbf{H}_0,\mathbf{H}_2)$. The first merge node computes the 1-chain $\mathbf{H}_1 = \mathcal{M}_{B_{0,1}^T,B_{1,2}} (\mathbf{H}_0,\mathbf{H}_2)$. Similarly, the second merge node  $\mathcal{M}_{B_{1,3}^T, B_{2,3}} \colon\mathcal{C}^1 \times \mathcal{C}^2 \to \mathcal{C}^3 $ computes the 3-cochain $\mathbf{H}_3=\mathcal{M}_{B_{1,3}^T, B_{2,3}}(\mathbf{H}_1,\mathbf{H}_2)$. The merge node perspective allows to compute a cochain supported on the 1-cells and 3-cells without having initial cochains of these cells. On the other hand, the higher-order message-passing framework has two major constraints: it assumes as input initial cochains supported on all cells of all dimensions of the domain, and it updates all cochains supported on all cells of all dimensions of the input domain at every iteration.}
\label{merge_example}
\end{figure}

The difference between these two computational frameworks might seem notational and the message passing perspective might seem more intuitive, especially when working with graph-based models. However, we argue that the merge node framework is more natural and flexible computationally in the presence of a custom higher-order network architecture. To illustrate this, we consider the example visualized in Figure~\ref{merge_example}.

In Figure~\ref{merge_example}, the displayed neural network has a cochain input vector $(\mathbf{H}_0,\mathbf{H}_2) \in  \mathcal{C}^0 \times \mathcal{C}^2$.
In the first layer,
the neural network computes the cochain $\mathbf{H}_1 \in \mathcal{C}^1 $, while in the second layer it computes the cochain $\mathbf{H}_3\in \mathcal{C}^3$.
To obtain cochain $\mathbf{H}_1$ in the first layer, we need to consider the neighborhood functions induced by $B_{0,1}^T$ and $B_{1,2}$. However, if we employ Equations~\ref{eqn:homp0} and~\ref{eqn:homp1} to perform the computations determined by the first layer of the tensor diagram in Figure~\ref{merge_example}, then we notice that no cochain is provided on $\mathcal{C}^1$ as
part of the input. Hence, when applying Equations~\ref{eqn:homp0} and~\ref{eqn:homp1}, a special treatment is required since the vectors $\mathbf{h}_{x^1_j}$ have not been computed yet. Note that such an artifact is not present in GNNs, since they often update node features, which are typically provided as part of the input. To be specific, in GNNs, the first two arguments in the update rule of Equation~\ref{eqn:homp0} are cochains that are supported on the 0-cells of the underlying graph.

Similarly, to compute the cochain $\mathbf{H}_3 \in  \mathcal{C}^3$ in the second layer of Figure~\ref{merge_example}, we must consider the neighborhood functions induced by $B_{1,3}^T$ and $B_{2,3}$, and we must use the cochain vector $(\mathbf{H}_1,\mathbf{H}_2)$. This means that the cochains $\mathbf{H}_1$ and $\mathbf{H}_3$ resulting from the computation of the neural network given in Figure~\ref{merge_example} are not obtained from an iterative process. 
Further, the input vectors $\mathbf{H}_0$ and $\mathbf{H}_2$ are never updated at any step of the procedure. Finally, the cochains $\mathbf{H}_1$ and $\mathbf{H}_3$ are never updated. From the perspective of update rules such as the ones appearing in the higher-order message passing framework (Definition~\ref{homp-definition}), this setting is unnatural in the sense that
it assumes \textit{initial cochains supported on all cells} of all dimensions as input, and in the sense that it \textit{updates all cochains supported on all cells} in the complex of the input domain at every iteration.  

In practice, such difficulties
in using the higher-order message passing framework
can be overcome with ad hoc engineering solutions
based on turning on and off iterations on certain cochains or based on introducing auxiliary cochains. The merge node is designed to overcome these limitations. Specifically, from the merge node perspective,
we can think of the first layer 
of Figure~\ref{merge_example}
as a function
$\mathcal{M}_{B_{0,1}^T,B_{1,2}}\colon \mathcal{C}^0 \times \mathcal{C}^1 \to \mathcal{C}^1$; 
see Equation~\ref{functional}.
The function
$\mathcal{M}_{B_{0,1}^T,B_{1,2}}$
takes as input the cochain vector $(\mathbf{H}_0,\mathbf{H}_2)$,
and computes the 1-chain $\mathbf{H}_1 = \mathcal{M}_{B_{0,1}^T,B_{1,2}} (\mathbf{H}_0,\mathbf{H}_2)$. Similarly,
we compute the 3-cochain
$\mathbf{H}_3=\mathcal{M}_{B_{1,3}^T, B_{2,3}}(\mathbf{H}_1,\mathbf{H}_2)$
using a merge node
$\mathcal{M}_{B_{1,3}^T, B_{2,3}} \colon
\mathcal{C}^1 \times \mathcal{C}^2 \to \mathcal{C}^3 $.

\subsection{Attention higher-order message passing and CCANNs}
\label{homp-attention}

Here, we demonstrate the connection between
higher-order message passing (Definition~\ref{homp-definition})
and CCANNs (Section~\ref{sec:hoans}).
Initially, we introduce an attention version of Definition~\ref{homp-definition}. 

\begin{tcolorbox}
[width=\linewidth, sharp corners=all, colback=white!95!black]
\begin{definition}[Attention higher-order message passing on a CC]
\label{attention_homp}
Let $\CCX$ be a CC.
Let $\mathcal{N}=\{ \mathcal{N}_1,\ldots,\mathcal{N}_n\}$ be a set of neighborhood functions defined on  $\CCX$.
Let $x$ be a cell and $y\in \mathcal{N}_k(x)$ for some $\mathcal{N}_k \in \mathcal{N}$. A \textbf{message} $m_{x,y}$ between cells $x$ and $y$
is a computation that depends on these two cells
or on the data supported on them.
Denote by $\mathcal{N}(x)$ the multi-set  $\{\!\!\{ \mathcal{N}_1(x) , \ldots ,  \mathcal{N}_n (x) \}\!\!\}$,
and by $\mathbf{h}_x^{(l)}$ some data supported on the cell $x$ at layer $l$.
\textbf{Attention higher-order message passing} on $\CCX$, 
induced by $\mathcal{N}$, is defined via the following four update rules:
\begin{align}
\label{ahomp0}
m_{x,y} &= \alpha_{\mathcal{N}_k}(\mathbf{h}_x^{(l)},\mathbf{h}_y^{(l)}), \\ \label{ahomp1}
m_{x}^k &=  \bigoplus_{y \in \mathcal{N}_k(x)} a^k(x,y)  m_{x,y}, \; 1\leq k \leq n ,   \\ \label{ahomp2}
m_{x} &=  \bigotimes_{ \mathcal{N}_k \in \mathcal{N} } b^k m_x^k ,\\ \label{ahomp3}
\mathbf{h}_x^{(l+1)} &= \beta (\mathbf{h}_x^{(l)}, m_x) .
\end{align}
Here, $a^k \colon \{x\} \times \mathcal{N}_k(x)\to [0,1] $ is a higher-order attention function  (Definition~\ref{hoa}),
$b^k$ are trainable attention weights satisfying $\sum_{k=1}^n b^k=1$,
$\bigoplus$ is a permutation-invariant aggregation function, $\bigotimes$ is an aggregation function,  $\alpha_{\mathcal{N}_k}$ and $\beta$ are differentiable functions.
\end{definition}
\end{tcolorbox}

Definition~\ref{attention_homp} distinguishes two types of attention weights. The first type is determined by the function $a^k$. The attention weight $a^k(x,y)$ of Equation~\ref{ahomp1} depends on the neighborhood function $\mathcal{N}_k$ and on cells $x$ and $y$. Further, $a^k(x,y)$ determines the attention a cell $x$ pays to its surrounding neighbors $y\in\mathcal{N}_k$,
as determined by the neighborhood function $\mathcal{N}_k$. The CC-attention push-forward operations defined in Section~\ref{sec:hoans} are a particular parameterized realization of these weights. On the other hand, the weights $b^k$ of Equation~\ref{ahomp2} are only a function of the neighborhood $\mathcal{N}_k$, and therefore determine the attention that cell $x$ pays to the information obtained from each neighborhood function $\mathcal{N}_k$. In our CC-attention push-forward operations given in Section~\ref{sec:hoans}, we set $b^k$ equal to one. However, the notion of merge node (Definition~\ref{exact_definition_merge_node})
 can be easily extended to introduce a corresponding notion of \textit{attention merge node}, which in turn can be used to realize Equation~\ref{ahomp2} in practice. Note that the attention determined by weights $b^k$ is unique to higher-order domains, and does not arise in graph-based attention models.

\section{Push-forward and higher-order (un)pooling}
\label{pooling-hoans}

This section shows how the push-forward operation of Definition \ref{pushing_exact_definition}
can be used to realize (un)pooling operations on CCs,
and subsequently introduces (un)pooling operations for $\CCN$s.
Further, this section demonstrates how CC-based pooling
provides a unifying framework for image and graph-based pooling,
and how shape-preserving pooling on CCs is related to the mapper on graphs.

In particular, we establish yet another unifying mathematical principle: pooling, as message passing, can be fundamentally built from push-forward operations. Thus, push-forward operations form the main fundamental building block from which all higher-order computations can be realized. This realization is important because it establishes a mathematical foundation for a unifying deep learning application programming interface (API) on complexes that combines pooling as well as message passing-based computations as a single operation. Indeed, in \href{https://github.com/pyt-team/TopoModelX}{TopoModelX}\footnote{https://github.com/pyt-team/TopoModelX}, one of our contributed Python packages, higher-order message passing and the pooling/unpooling operations are implemented as a single function across various topological domains.


\subsection{CC-(un)pooling}
\label{subsec:cc_pool}

\begin{figure}[!t]
\begin{center}
\includegraphics[scale = 0.064, keepaspectratio = 0.20]{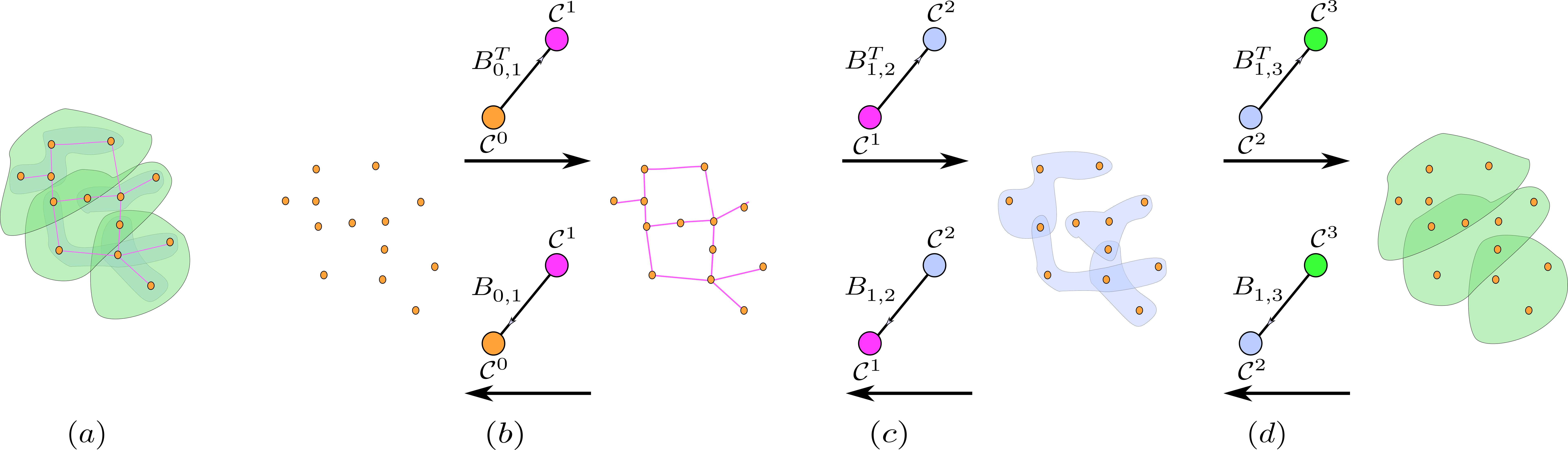}
\end{center}
\caption{An example of successive CC-(un)pooling operations.
A CC-pooling operation exploits the hierarchical structure of the underlying CC
to coarsen a lower-rank cochain by pushing it forward to higher-order cells.
CC-pooling operations improve invariance to certain distortions. 
Pink, blue, and green cells have ranks one, two, and three, respectively.
A CC $\CCX$ of dimension three is shown in (a).
On $\CCX$, we consider three cochain operators:
$B_{0,1} \colon\mathcal{C}^1\to \mathcal{C}^0  $,
$B_{1,2} \colon\mathcal{C}^2\to \mathcal{C}^1 $ and
$B_{2,3}\colon\mathcal{C}^3\to \mathcal{C}^2$.
For the top row in Figures (b), (c), and (d),
we assume that we are initially given a 0-cochain $\mathbf{H}_0$,
while for the bottom row in the same figure,
we assume that we are initially given a 3-cochain $\mathbf{H}_3$.
For instance, at the top row of Figure (b),
the input 0-cochain $\mathbf{H}_0$ gets pushed forward via the functional $\mathcal{F}_{B_{0,1}^T} \colon\mathcal{C}^0\to \mathcal{C}^1  $  to a 1-cochain $\mathbf{H}_1$. This push-forward operator induced by $B_{0,1}^T$ is a CC-pooling operator, since it sends a 0-cochain to a cochain of higher rank.
At the bottom row of Figure (b),
the push-forward operator
$\mathcal{F}_{B_{0,1}}\colon \mathcal{C}^1\to \mathcal{C}^0$
induced by $B_{0,1}$
is an unpooling operator that sends a $1$-cochain to a $0$-cochain.
Figures (c) and (d) are similar to (b),
demonstrating the pooling operators induced by $B_{1,2}^T $ and $B_{2,3}^T$ (top),
and the unpooling operators induced by $B_{1,2} $ and $B_{2,3}$ (bottom).}
\label{pooling_hoans}
\end{figure}

We define a CC-based pooling operation that extends the main characteristics of image-based and graph-based pooling operations. Specifically, we build a pooling operation that `downscales' the size of a signal supported on a CC $\CCX$. To this end, we exploit the hierarchical nature of CCs and define the pooling operation as a push-forward operation induced by a cochain map
$G\colon\mathcal{C}^{i}(\mathcal{X})\to \mathcal{C}^{j}(\mathcal{X})$
that pushes an $i$-cochain to a $j$-cochain. To obtain a useful pooling operation that downscales the size of its input $i$-cochain, we impose the constraint $j>i$.
Definition~\ref{pooling_exact_definition} realizes our idea of CC-pooling.
Figure~\ref{pooling_hoans} visualizes the intuition behind 
Definition~\ref{pooling_exact_definition}.
In particular, Figure~\ref{pooling_hoans}
shows an example of successive applications
of pooling operations on cochains supported on a CC of dimension three.

\begin{tcolorbox}
[width=\linewidth, sharp corners=all, colback=white!95!black]
\begin{definition}[CC-pooling operation]
\label{pooling_exact_definition}
Let $\CCX$ be a CC and $G\colon\mathcal{C}^{i}(\CCX)\to \mathcal{C}^{j}(\CCX)$ a cochain map. The push-forward operation induced by $G$ is called a
\textbf{CC-pooling operation} if $j>i$.
\end{definition}
\end{tcolorbox}


In Definition~\ref{unpooling_exact_definition},
we introduce an unpooling operation on CCs
that pushes forward a cochain to a lower-rank cochain.
Figure~\ref{pooling_hoans} shows as example of unpooling on CCs.

\begin{tcolorbox}
[width=\linewidth, sharp corners=all, colback=white!95!black]
\begin{definition}[CC-unpooling operation]
\label{unpooling_exact_definition}
Let $\CCX$ be a CC and $G\colon\mathcal{C}^i(\CCX)\to \mathcal{C}^j(\CCX)$ a cochain map. The push-forward operation induced by $G$ is called a \textbf{CC-unpooling operation} if $j<i$.
\end{definition}
\end{tcolorbox}

\subsection{Formulating common pooling operations as CC-pooling}

In this section, we formulate common pooling operations in terms of CC-pooling.
In particular, we demonstrate that graph and image pooling can be cast as CC-pooling.

\subsubsection{Graph pooling as CC-pooling}
\label{graph.pooling}

Here, we briefly demonstrate that the CC-pooling operation (Definition~\ref{pooling_exact_definition}) 
is consistent with a graph-based pooling algorithm.
Let $\mathbf{H}_0$ be a cochain defined on the vertices and edges of a graph $\mathcal{G}$.
Moreover, let $\mathbf{H}^{\prime}_0$ be a cochain defined on the vertices of a coarsened version $\mathcal{G}^{\prime }$ of $\mathcal{G}$.
Under such a setup,
$\mathbf{H}^{\prime}_0$ represents a coarsened version of $\mathbf{H}_0$.
A graph pooling function supported on the pair $(\mathcal{G},\mathbf{H}_0)$ is a function of the form
$\mathcal{POOL} \colon (\mathcal{G},\mathbf{H}_0) \to (\mathcal{G}^{\prime},\mathbf{H}^{\prime}_0)$ that sends every vertex in $\mathcal{G}$ to a vertex in $\mathcal{G}^{\prime}$, which corresponds to a cluster of vertices in $\mathcal{G}$. We now elucidate how the function $\mathcal{POOL}$ can be realized in terms of CC-pooling.  


\begin{proposition}
 The function $\mathcal{POOL}$ can be realized in terms of CC-pooling operations. 
\end{proposition}

\begin{proof}
Each vertex in the graph $\mathcal{G}^{\prime}$ represents a cluster of vertices in the original graph $\mathcal{G}$. Using the membership of these clusters, we construct a CC by augmenting $\mathcal{G}$ by a collection of 2-cells, so that each of these cells corresponds to a supernode of $\mathcal{G}^{\prime}$. We denote the resulting CC structure by $\mathcal{X}_{\mathcal{G}}$, consisting of $\mathcal{G}$ augmented by the 2-cells. Hence, any 0-cochain $\mathbf{H}^{\prime}_0 $ defined on $\mathcal{G}^{\prime}$ can be written as a 2-cochain $\mathbf{H}_2 \in \mathcal{C}^2(\mathcal{X}_{\mathcal{G}}) $. The relation between the vertices of the original graph $\mathcal{G}$ and the vertices of the pooled graph $\mathcal{G}^{\prime}$, or equivalently the CC $\mathcal{X}_{\mathcal{G}}$, is described via the incidence matrix $B_{0,2}^T$. Hence, learning the signal $\mathbf{H}_2$ can be realized in terms of a map $B_{0,2}^T \colon \mathcal{C}^{2} (\mathcal{X}_{\mathcal{G}}) \to \mathcal{C}^{0}(\mathcal{X}_{\mathcal{G}}) $ that pushes forward the cochain $\mathbf{H}_0$ to  $\mathbf{H}_2$.
\end{proof}

The 2-cells defined on $\mathcal{X}_{\mathcal{G}}$ can be practically constructed using the \textit{mapper on graphs}~\cite{hajij2018mog}, a classification tool in TDA.
See Section~\ref{mapper} for more details of such a construction.




\subsubsection{Image pooling as CC-pooling}
\label{image.pooling}

\begin{figure}[!t]
\begin{center}
\includegraphics[scale = 0.075, keepaspectratio = 0.20]{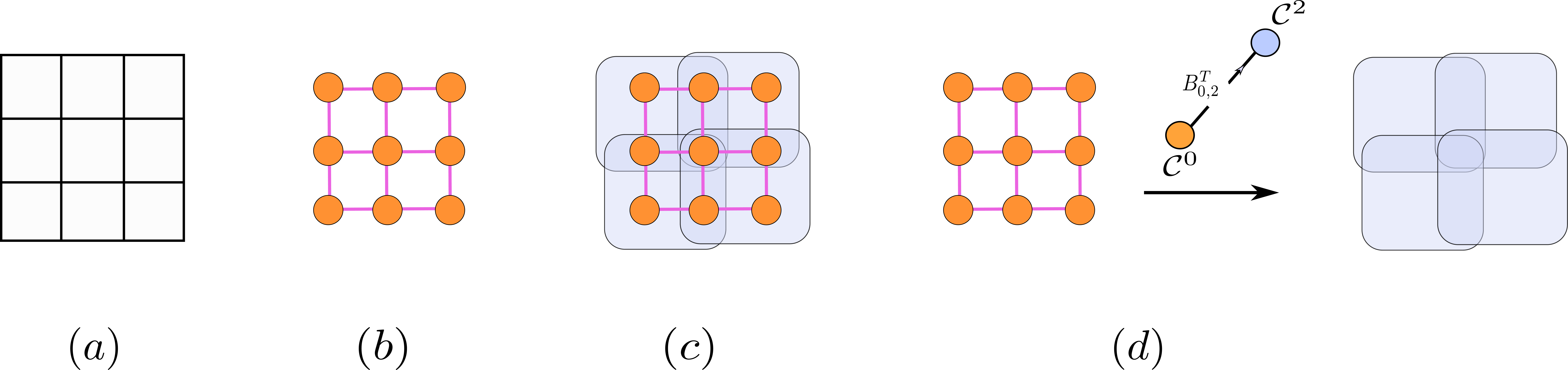}
\end{center}
\caption{ Realizing image pooling in terms of CC-pooling.
(a): An image of size $3\times3$.
(b): The lattice graph that corresponds to the image given in (a).
(c): Augmenting the lattice graph with 2-cells.
Choosing these particular cells shown in (c) is equivalent to choosing the image-pooling window size to be $2\times 2$ and the pooling stride to be one.
(d): Performing the image-pooling computation is equivalent to performing a CC-pooling operation induced by the cochain map $B_{0,2}^T \colon\mathcal{C}^0 \to \mathcal{C}^2$, which pushes forward the image signal (the $0$-cochain supported on $\CCX^2$) to a signal supported on $\CCX^2$.\vspace{-3mm}}
\label{image_pooling}
\end{figure}

Since images can be realized as lattice graphs, a signal stored on an image grid can be realized as a 0-cochain of the lattice graph that corresponds to the image. See Figures~\ref{image_pooling}(a--b) for an example. Here, we demonstrate that the CC-pooling
operation (Definition~\ref{pooling_exact_definition})
is consistent with the known image-pooling definition. Indeed, one may augment the lattice graph of Figure~\ref{image_pooling}(b)
by 2-cells, as shown in Figure~\ref{image_pooling}(c), 
to perform the image pooling operation. Usually, these cells have a regular window size. In Figure~\ref{image_pooling}(c),
we have chosen the pooling window size, or equivalently the size of the 2-cell, to be $2\times 2$, and the pooling stride to be 1. The image pooling operation in this case can be realized as a CC-pooling operation induced by the cochain map $B_{0,2}^T \colon\mathcal{C}^0 \to \mathcal{C}^2$, as visualized in Figure~\ref{image_pooling}(d). We formally record this in the following proposition.

\begin{proposition}
An image pooling operator can be realized in terms of a push-forward operator from the underlying image domain to a 2-dimensional CC obtained by augmenting the image by appropriate 2-cells where image pooling computations occur.
\end{proposition}
\begin{proof}
The proof is a straightforward conclusion from the definition of image pooling.
\end{proof}

\subsection{(Un)pooling CCNNs}

The pooling operator of Definition~\ref{pooling_exact_definition} considers only the special case in which the tensor diagram of a CCNN has a single edge. In what follows, we generalize the notion of pooling
by identifying the defining properties that characterize a CCNN as a pooling CCNN.
To this end, we start with a CCNN whose tensor diagram has a height of one.

\begin{tcolorbox}
[width=\linewidth, sharp corners=all, colback=white!95!black]
\begin{definition}[Pooling CCNN of height one]
\label{height_1_pooling}
Consider a CCNN
represented by a tensor diagram $\CCN_{\mathbf{G};\mathbf{W}}$ of height one.
Let
$\mathcal{C}^{i_1}\times\mathcal{C}^{i_2}\times \cdots \times  \mathcal{C}^{i_m}$
be the domain
and let $\mathcal{C}^{j_1}\times\mathcal{C}^{j_2}\times \cdots \times \mathcal{C}^{j_n}$
be the codomain of the CCNN.
Let $i_{min}=min(i_1,\ldots,i_m)$ and $j_{min}=min(j_1,\ldots,j_n)$. We say that the CCNN is a \textbf{pooling CCNN of height one} if
\begin{enumerate}[topsep=1pt,itemsep=1pt]
\item $i_{min}< j_{min}$, and
\item the tensor diagram $\CCN_{\mathbf{G};\mathbf{W}}$
has an edge labeled by a cochain operator
$G\colon \mathcal{C}^{i_{min}} \to \mathcal{C}^{k}$ for some $k\geq j_{min}$.
\end{enumerate}
\end{definition}
\end{tcolorbox}

\begin{figure}[!t]
\begin{center}
\includegraphics[scale = 1.2, keepaspectratio = 0.20]{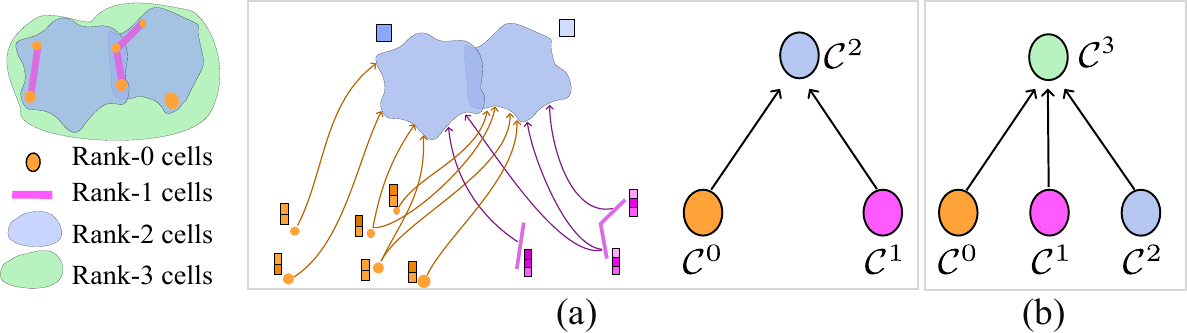}
\end{center}
\caption{Examples of pooling operations.
In this example, a CC of dimension three is displayed.
A rank cell of rank $3$ (shown in green) has the highest rank among all cells in the CC.
(a): A pooling CCNN of height one
pools a cochain vector $(\mathbf{H}_0,\mathbf{H}_1)$ to a 2-cochain $\mathbf{H}_2$.
(b): A readout operation can be realized as a pooling CCNN of height one by
encapsulating the entire CC by a single (green) cell
of rank higher than all other cells in the CC, and
by pooling (reading out) all signals of lower-order cells to the
encapsulating (green) cell.\vspace{-3mm}}
\label{readout}
\end{figure}

Intuitively, a CCNN represented by a tensor diagram of height one is a pooling CCNN of height one if it pushes forward
its lowest-rank signals to higher-rank cells. Observe that a readout operation can be realized as a pooling CCNN of height one;
see Figure~\ref{readout} for an illustration.

A CCNN may not perform a pooling operation at every layer, and it may preserve the dimensionality of the lowest-rank signal.
Before we give the general definition of pooling CCNNs, we first define lowest rank-preserving CCNNs of height one.

\begin{tcolorbox}
[width=\linewidth, sharp corners=all, colback=white!95!black]
\begin{definition}[Lowest rank-preserving CCNN of height one]
Consider a CCNN represented by a tensor diagram of height one.
Let
$\mathcal{C}^{i_1}\times\mathcal{C}^{i_2}\times \cdots \times  \mathcal{C}^{i_m}$
be the domain
and let
$\mathcal{C}^{j_1}\times\mathcal{C}^{j_2}\times \cdots \times \mathcal{C}^{j_n}$
be the codomain of the CCNN.
Let $i_{min}=min(i_1,\ldots,i_m)$ and $j_{min}=min(j_1,\ldots,i_n)$.
We say that the CCNN is a \textbf{lowest rank-preserving CCNN of height one} if $i_{min}= j_{min}$.
\end{definition}
\end{tcolorbox}

Every CCNN is a composition of CCNNs
that are represented by tensor diagrams of height one.
Hence, pooling CCNNs can be characterized in terms of tensor diagrams of height one,
as elaborated in Definition~\ref{general_pooling_hoan}.

\begin{tcolorbox}
[width=\linewidth, sharp corners=all, colback=white!95!black]
\begin{definition}[Pooling CCNN]
\label{general_pooling_hoan}
Let $\CCN_{\mathbf{G};\mathbf{W}}$
be a tensor diagram representation of a CCNN.
We decompose the CCNN as
\begin{equation*}
\CCN_{\mathbf{G};\mathbf{W}}=
\CCN_{\mathbf{G}_N;\mathbf{W}_N} \circ \cdots \circ \CCN_{\mathbf{G}_1;\mathbf{W}_1},
\end{equation*}
where
$\CCN_{\mathbf{G}_i;\mathbf{W}_i},i=1,\ldots,N$, 
is a tensor diagram of height one
representing the $i$-th layer of the CCNN,
and $\mathbf{G}_i \subseteq \mathbf{G}$.
We call the CCNN represented by $\CCN_{\mathbf{G};\mathbf{W}}$
a \textbf{pooling CCNN} if
\begin{enumerate}[topsep=1pt,itemsep=1pt]
\item every $\CCN_{\mathbf{G}_i;\mathbf{W}_i}$
is either a pooling CCNN of height one or a lowest rank-preserving CCNN of height one, and
\item at least one of the layers $\CCN_{\mathbf{G}_i;\mathbf{W}_i}$ is a pooling CCNN of height one.
\end{enumerate}
\end{definition}
\end{tcolorbox}

Intuitively, a pooling CCNN is a CCNN whose tensor diagram forms a `ladder' that pushes signals to higher-rank cells at every layer. Figure~\ref{fig:tensor}(d) gives an example of a pooling CCNN of height two. 

An unpooling CCNN of height one is defined similarly to a pooling CCNN of height one
(Definition~\ref{height_1_pooling}),
with the only difference being that
the inequality $i_{min}<j_{min}$ becomes $i_{min}>j_{min}$.
Moreover, an unpooling CCNN (Definition~\ref{general_unpooling_hoan}) is defined
analogously to a pooling CCNN (Definition~\ref{general_pooling_hoan}).

\begin{tcolorbox}
[width=\linewidth, sharp corners=all, colback=white!95!black]
\begin{definition}[Unpooling CCNN]
\label{general_unpooling_hoan}
Let $\CCN_{\mathbf{G};\mathbf{W}}$
be a tensor diagram representation of a CCNN.
We decompose the CCNN as
\begin{equation*}
\CCN_{\mathbf{G};\mathbf{W}}=
\CCN_{\mathbf{G}_N;\mathbf{W}_N} \circ \cdots \circ \CCN_{\mathbf{G}_1;\mathbf{W}_1},
\end{equation*}
where
$\CCN_{\mathbf{G}_i;\mathbf{W}_i},i=1,\ldots,N$, 
is a tensor diagram of height one
representing the $i$-th layer of the CCNN,
and $\mathbf{G}_i \subseteq \mathbf{G}$.
We call the CCNN represented by $\CCN_{\mathbf{G};\mathbf{W}}$
an \textbf{unpooling CCNN} if
\begin{enumerate}[topsep=1pt,itemsep=1pt]
\item every $\CCN_{\mathbf{G}_i;\mathbf{W}_i}$ is either an unpooling CCNN of height one or a lowest rank-preserving CCNN of height one, and
\item at least one of the layers $\CCN_{\mathbf{G}_i;\mathbf{W}_i}$ is an unpooling CCNN of height one.
\end{enumerate}
\end{definition}
\end{tcolorbox}

\subsection{Mapper and the CC-pooling operation}
\label{mapper}

In practice, constructing a useful CC-pooling operation on a higher-order domain is determined by the higher-rank cells in the input CC. Similar to image-based models, CC-pooling operations can be applied sequentially at the end of a higher-order network to provide a summary representation of the input domain; see Figure~\ref{pooling_hoans} for an example. Such a hierarchical summary might not be readily available in the input CC. For instance, if $\CCX$ is a graph, then a CC-pooling operation, as given in Definition~\ref{pooling_exact_definition}, can only push forward an input node-signal to an edge-signal,
which may not always provide a compact summary of the input signal.

\begin{figure}[!t]
\begin{center}
\includegraphics[scale = 0.033, keepaspectratio = 0.20]{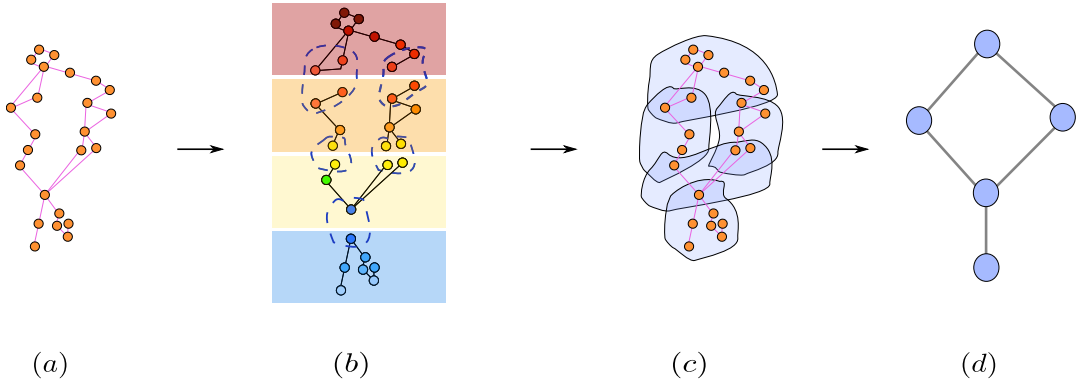}
\end{center}
\caption{An example of constructing a shape-preserving pooling operation on a CC $\CCX$. Here, we demonstrate the case in which $\CCX$ is a graph. We utilize the \textit{mapper on graphs (MOG)} construction~\cite{hajij2018mog,singh2007topological}, a graph skeletonization algorithm that can be used to augment $\CCX$ with topology-preserving cells of rank $2$.
(a): An input graph $\CCX$, which is a CC of dimension one.
(b): The MOG algorithm receives three elements as input,
namely the graph $\CCX$,
a feature-preserving scalar function $g\colon\CCX^0\to [a,b]$,
and a cover $\mathcal{U}$ of range $[a,b]$,
that is a collection of open sets that covers the closed interval $[a,b]$.
The scalar function $g$ is used to pull back a covering $\mathcal{U}$ on the range $[a,b]$ to a covering on $\CCX$. The colors of nodes in Figure (b) indicate the scalar values of $g$.
In Figure (b), $\CCX$ is split into four segments
so that each segment corresponds to a cover element of $\mathcal{U}$.
(c): The figure shows the connected components in pull-back cover elements. Each connected component is enclosed by a blue cell.
Each of these blue cells is considered as a cell of rank $2$.
We augment $\CCX$ with the blue cells to form a CC of dimension two,
consisting of the original 0-cells and 1-cells in $\CCX$
as well as the augmented 2-cells.
This augmented CC is denoted by $\CCX_{g,\mathcal{U}}$.
(d): The MOG algorithm constructs a graph, whose nodes are the connected components contained in each cover element in the pull-back of the cover $\mathcal{U}$, and whose edges are formed by the intersection between these connected components. In other words, the MOG-generated graph summarizes the connectivity between the augmented cells of rank $2$ added via the MOG algorithm. Observe that the adjacency matrix of the MOG-generated graph given in Figure (d) is equivalent to the adjacency matrix $A_{2,2}$ of $\CCX_{g,\mathcal{U}}$, since 2-cells in $\CCX$ are 2-adjacent if and only if they intersect on a node (and the latter occurs if and only if there is an edge between the nodes of the MOG-generated graph). Given this CC structure, the cochain map $B_{0,2}^T\colon\mathcal{C}^0(X_{g,\mathcal{U}})\to \mathcal{C}^2(X_{g,\mathcal{U}})$ can be used to induce a shape-preserving CC-pooling operation. Moreover, a signal $\mathbf{H}_0$ supported on the nodes of $\CCX$ can be push-forwarded and pooled to a signal $\mathbf{H}_2$ supported on the augmented 2-cells. This figure is inspired from~\cite{hajij2018mog}.}
\label{mog123}
\end{figure}

In such situations, one may choose to \textit{augment the input CC} $\CCX$ with a collection of new cells of dimension $\dim(\CCX)+1$ so that the new cells approximate the shape of the input CC $\CCX$. Figure~\ref{mog123} displays an example of augmenting a graph $\CCX$, which is a CC of dimension one, with new cells of dimension two using the \textit{mapper on graphs (MOG)} construction suggested in~\cite{hajij2018mog,dey2016multiscale,singh2007topological}\footnote{While graph skeletonization using the mapper algorithm~\cite{singh2007topological} has been studied in~\cite{dey2016multiscale}, our implementation and discussion here relies on the notions suggested in~\cite{hajij2018mog}.}.  The augmented higher-rank cells obtained from the
MOG construction summarize the shape features of the underlying graph,
which is a desirable pooling characteristic (e.g., in shape analysis). We refer to Appendix~\ref{mog_section} 
for details about the MOG construction of topology-preserving CC-pooling operations.

\section{Hasse graph interpretation and equivariances of CCNNs}

In this section, we scrutinize the properties of our topological learning machines by establishing connections to the pre-existing findings regarding GNNs. We begin our inquiry by elucidating the interpretation of CCs as specialized graphs, known as \emph{Hasse graphs}, followed by characterizing their equivariance properties against the actions of permutations and orientations. We further link our definitions of equivariances to the conventional ones under the Hasse graph representation.

\subsection{Hasse graph interpretation of CCNNs}
\label{sec:Hasse}

We first demonstrate that every CC can be reduced to a unique and special graph known as the \emph{Hasse graph}. This reduction enables us to analyze and understand various computational and conceptual aspects of CCNNs in terms of graph-based models. 
\subsubsection{CCs as Hasse graphs}
\label{hasse}
Definition~\ref{maps} implies that a CC is a poset, that is a partially ordered set whose partial order relation is the set inclusion relation. It also implies that two CCs are equivalent if and only if their posets are equivalent\footnote{For related structures (e.g., simplicial/cell complexes), this poset is typically called the \textit{face poset}~\cite{wachs2006poset}.}. 
Definition~\ref{HG} introduces the \textit{Hasse graph}~\cite{abramenko2008buildings,wachs2006poset} of a CC, which is a directed graph associated with a finite poset.

\begin{tcolorbox}
[width=\linewidth, sharp corners=all, colback=white!95!black]
\begin{definition}[Hasse graph]
\label{HG}
The \textbf{Hasse graph} of a CC $(S, \CCX,\rk)$ is a directed graph $\mathcal{H}_{\CCX}= (V (\mathcal{H}_{\CCX}), E(\mathcal{H}_{\CCX}) )$ with vertices
$V (\mathcal{H}_{\CCX})=\CCX$ and edges
$E(\mathcal{H}_{\CCX})=\{ (x,y) : x\subsetneq y, \rk(x)=\rk(y)-1 \}$.
\end{definition}
\end{tcolorbox}

\begin{figure}[!t]
\begin{center}
\includegraphics[scale = 0.14, keepaspectratio = 0.20]{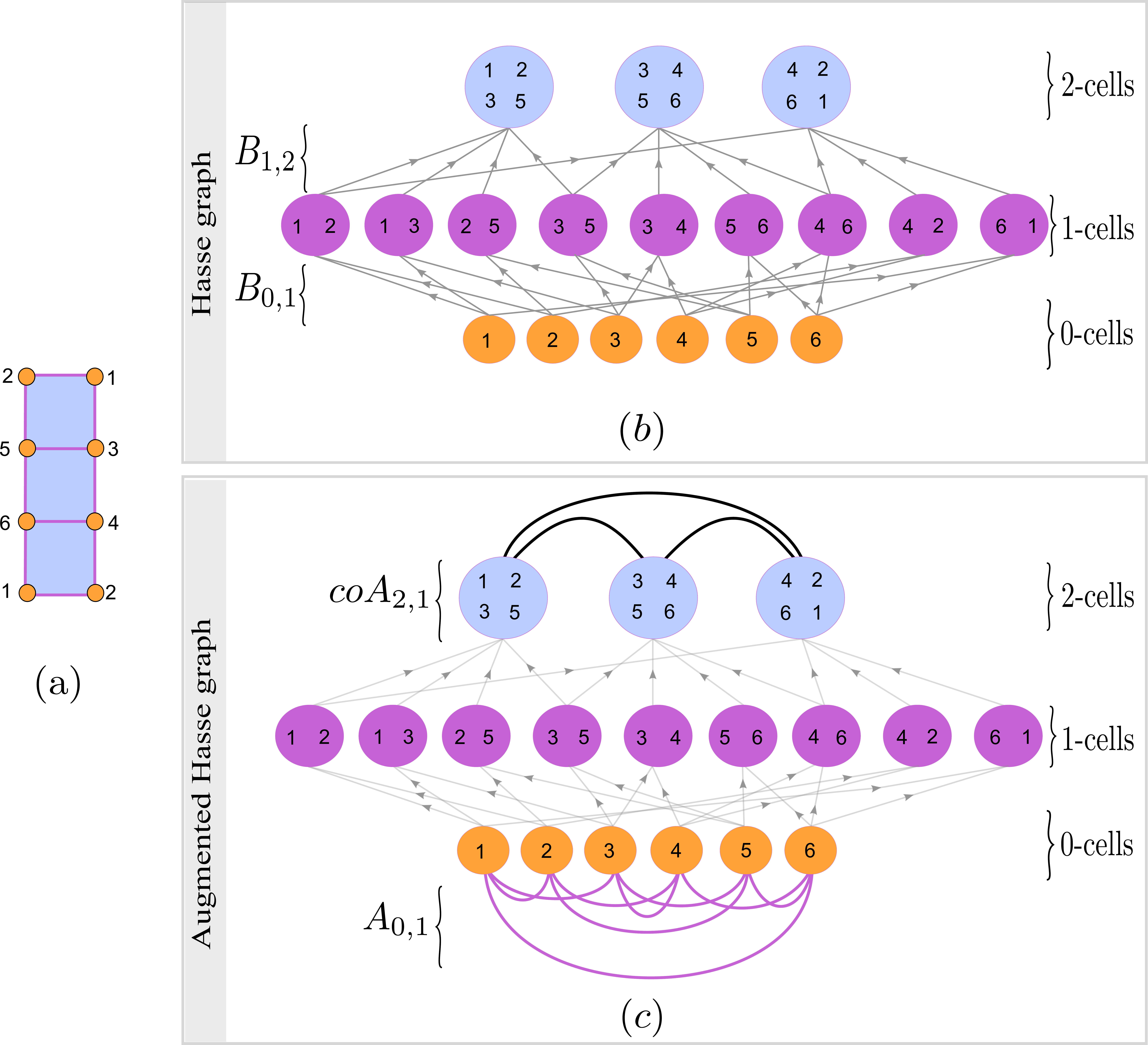}
\end{center}
\caption{Example of the Hasse graph of a CC.
(a): CC of a M\"{o}bius strip.
(b): Hasse graph of the CC, describing the poset structure between cells.
(c): Hasse graph augmented with the edges defined via $A_{0,1}$ and $coA_{2,1}$.}
\label{fig:hasse-diagram}
\end{figure}

The vertices of the Hasse graph $\mathcal{H}_{\CCX}$ of a CC $(S, \CCX,\rk)$
are the cells of $\CCX$,
while the edges of $\mathcal{H}_{\CCX}$
are determined by the immediate incidence among these cells.
Figure~\ref{fig:hasse-diagram} shows an example of the Hasse graph of a CC.

\begin{figure}[!t]
\begin{center}
\includegraphics[scale = 0.070, keepaspectratio = 0.21]{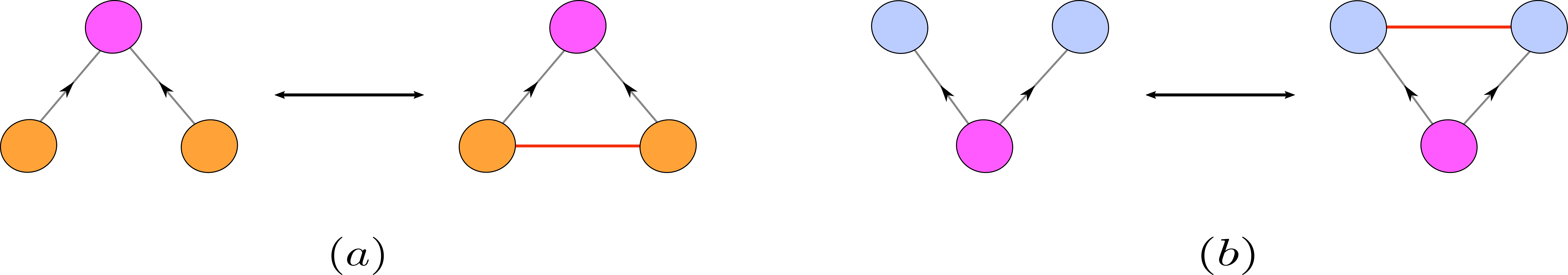}
\end{center}
\caption{Relation between immediate incidence and (co)adjacency on the Hasse graph of a CC.
This figure supports visually
the proofs of parts 2 and 3 in Proposition~\ref{prop:structure}.
(a): Two $(k-1)$ cells $x^{k-1}$ and $y^{k-1}$ (orange vertices) are 1-adjacent if and only if there exists a $k$-cell $z^k$ (pink vertex) that is incident to $x^{k-1}$ and $y^{k-1}$.
(b): Two $(k+1)$ cells $x^{k+1}$ and $y^{k+1}$ (blue vertices) are 1-coadjacent if and only if there exists a $k$-cell $z^k$ (pink vertex) that is incident to $x^{k+1}$ and $y^{k+1}$.}
\label{relation between adj/coadj/incidence}
\end{figure}

The \textit{CC structure class} is the set of CCs determined
up to isomorphism, according to Definition~\ref{maps}.
Proposition~\ref{prop:structure} provides sufficient criteria for determining CC structure classes.
The proof of Proposition~\ref{prop:structure} relies on the observations that
CC structure classes are determined by the underlying Hasse graph representation and that
the Hasse graph provides the same information as the incidence matrices $\{B_{k,k+1}\}_{k=0}^{\dim(\CCX)-1}$.
Figure~\ref{relation between adj/coadj/incidence} supports visually
the proofs of parts 2 and 3 in Proposition~\ref{prop:structure}.

\begin{proposition}
\label{prop:structure}
Let $(S, \CCX,\rk)$ be a CC. For the CC structure class indicated by $(S, \CCX,\rk)$,
the following sufficient conditions hold:
\begin{enumerate}
\item The CC structure class is determined by the incidence matrices $\{B_{k,k+1}\}_{k=0}^{ \dim(\CCX) -1}$.
\item The CC structure class is determined by the adjacency matrices $\{A_{k,1}\}_{k=0}^{\dim(\CCX)-1}$.
\item The CC structure class is determined by the coadjacency matrices $\{coA_{k,1}\}_{k=1}^{\dim(\CCX)}$.
 \end{enumerate}
\end{proposition}

\begin{proof}
The proof of the three parts of the proposition follows by noting that the structure of a CC is determined completely by its Hasse graph representation.
The first part of the proposition follows from the fact that the edges in the Hasse graph are precisely the non-zero entries of matrices $\{B_{k,k+1}\}_{k=0}^{\dim(\CCX-1)}.$ The second part follows by observing that two $(k-1)$ cells $x^{k-1}$ and $y^{k-1}$ are 1-adjacent if and only if there exists a $k$-cell $z^k$ that is incident to $x^{k-1}$ and $y^{k-1}$. The third part is confirmed by noting that two $(k+1)$-cells $x^{k+1}$ and $y^{k+1}$ are 1-coadjacent if and only if there exists a $k$-cell $z^k$ that is incident to $x^{k+1}$ and $y^{k+1}$.
\end{proof}

\subsubsection{Augmented Hasse graphs}
The Hasse graph of a CC is useful because it shows that computations for a higher-order deep learning model can be reduced to computations for a graph-based model.
Particularly, a $k$-cochain (signal) being processed on a CC $\CCX$
can be thought as a signal on the corresponding vertices of the associated Hasse graph $\mathcal{H}_{\CCX}$. The edges specified by the matrices $B_{k,k+1}$ determine the message-passing structure of a given higher-order model defined on $\CCX$. However, the message-passing structure determined via the matrices $A_{r,k}$ is not directly supported on the corresponding edges of $\mathcal{H}_{\CCX}$.
Thus, it is sometimes desirable to \textit{augment the Hasse graph} with additional edges other than the ones specified by the poset partial order relation of the CC.
Along these lines, Definition~\ref{ahg} introduces the notion of augmented Hasse graph.

\begin{tcolorbox}
[width=\linewidth, sharp corners=all, colback=white!95!black]
\begin{definition}[Augmented Hasse graph]
\label{ahg}
Let $\CCX$ be a CC, and let $\mathcal{H}_{\CCX}$ be its Hasse graph
with vertex set $V(\mathcal{H}_{\CCX})$ and edge set $E(\mathcal{H}_{\CCX})$.
Let $\mathcal{N}=\{\mathcal{N}_1,\ldots,\mathcal{N}_n\}$ be a set of neighborhood functions defined on $\CCX$. We say that $\mathcal{H}_{\CCX}$ has an augmented edge $e_{x,y}$ induced by $\mathcal{N}$ if there exist $\mathcal{N}_i \in \mathcal{N} $ such that $x \in \mathcal{N}_i(y) $ or $y \in \mathcal{N}_i(x)$. Denote by $E_{\mathcal{N}}$ the set of all augmented edges induced by  $\mathcal{N}$. The \textbf{augmented Hasse graph} of $\CCX$ induced by $\mathcal{N}$ is defined to be the graph
$\mathcal{H}_{\CCX}(\mathcal{N})= (V(\mathcal{H}_{\CCX}), E(\mathcal{H}_{\CCX}) \cup     E_{\mathcal{N}}) $. 
\end{definition}
\end{tcolorbox}

It is easier to think of the augmented Hasse graph in Definition~\ref{ahg} in terms of the matrices
$\mathbf{G}=\{G_1,\ldots,G_n\}$ associated with the neighborhood functions $\mathcal{N}=\{\mathcal{N}_1,\ldots,\mathcal{N}_n\}$. Each augmented edge in $\mathcal{H}_{\CCX}(\mathcal{N})$ corresponds to a non-zero entry in some $G_i\in \mathbf{G}$. Since $\mathcal{N}$ and $\mathbf{G}$ store equivalent information, we use $\mathcal{H}_{\CCX}(\mathbf{G})$ to denote the augmented Hasse graph induced by the edges determined by $\mathbf{G}$. For instance, the graph given in Figure~\ref{fig:hasse-diagram}(c) is denoted by $\mathcal{H}_{\CCX}( A_{0,1},coA_{2,1}).$

\subsubsection{Reducibility of CCNNs to graph-based models}

In this section, we show that any CCNN-based computational model can be realized as a message-passing scheme over a subgraph of the augmented Hasse graph of the underlying CC. 
Every CCNN is determined via a computational tensor diagram, which can be built using the elementary tensor operations, namely push-forward operations, merge nodes and split nodes. Thus, the reducibility of CCNN-based computations to message-passing schemes over graphs can be achieved by proving that these three tensor operations can be executed on an augmented Hasse graph. Proposition~\ref{Hasse-pushforward} states that push-forward operations are executable on augmented Hasse graphs.


\begin{proposition}
\label{Hasse-pushforward}
Let $\CCX$ be a CC and let  $\mathcal{F}_G \colon \mathcal{C}^i(\CCX)\to \mathcal{C}^j(\CCX)$ be a push-forward operator induced by a cochain map $G\colon\mathcal{C}^i(\CCX)\to \mathcal{C}^j(\CCX) $.
Any computation executed via $\mathcal{F}_G$ can be reduced to a corresponding computation over the augmented Hassed graph $\mathcal{H}_{\CCX}(G)$ of $\CCX$.
\end{proposition}

\begin{proof}
Let $\CCX$ be a CC. Let $\mathcal{H}_{\CCX}(G)$ be the augmented Hasse graph of $\CCX$ determined by $G$. The definition of the augmented Hasse graph implies that there is a one-to-one correspondence between the vertices $\mathcal{H}_{\CCX}(G)$ and the cells in $\mathcal{X}$. Given a cell $x\in \mathcal{X}$, let $x^{\prime}$ be the corresponding vertex in $\mathcal{H}_{\CCX}(G)$. Let $y$ be a cell in $\mathcal{X}$ with a feature vector $\mathbf{h}_y$ computed via the push-forward operation specified by Equation~\ref{functional}. Recall that the vector $\mathbf{h}_y$ is computed by aggregating all vectors $\mathbf{h}_x$ attached to
the neighbors $x \in \CCX^i$ of $y$
with respect to the neighborhood function $\mathcal{N}_{G^T}$. Let $m_{x,y}$ be a computation (message) that is executed between two cells $x$ and $y$ of $\CCX$ as a part of the computation of push-forward $\mathcal{F}_G$. It follows from the augmented Hasse graph definition that the cells $x$ and $y$ must have a corresponding non-zero entry in matrix $G$. Moreover, this non-zero entry corresponds to an edge in $\mathcal{H}_{\CCX}(G)$ between $x^{\prime}$ and $y^{\prime}$. Thus, the computation $m_{x,y}$ between the cells $x$ and $y$ of $\CCX$
can be carried out as the computation (message) $m_{x^{\prime},y^{\prime}}$
between the corresponding vertices $x^{\prime}$ and $y^{\prime}$ of $\mathcal{H}_{\CCX}(G)$.
\end{proof}

Similarly, computations on an arbitrary merge node can be characterized in terms of computations on a subgraph of the augmented Hasse graph of the underlying CC.
Proposition~\ref{Hasse} formalizes this statement.

\begin{proposition}
\label{Hasse}
Any computation executed via a merge node $\mathcal{M}_{\mathbf{G},\mathbf{W}}$ as given in Equation~\ref{sum} can be reduced to a corresponding computation over the augmented Hasse graph $\mathcal{H}_{\CCX}(\mathbf{G})$ of the underlying CC. 
\end{proposition}

\begin{proof}
Let $\CCX$ be a CC. Let $\mathbf{G}=\{ G_1,\ldots,G_n\}$ be a sequence of cochain operators defined on $\CCX$. Let $\mathcal{H}_{\CCX}(\mathbf{G})$ be the augmented Hasse graph determined by $\mathbf{G}$. By the augmented Hasse graph definition, there is a one-to-one correspondence between the vertices of $\mathcal{H}_{\CCX}(\mathbf{G})$  and the cells of $\CCX$. For each cell $x\in \CCX$, let $x^{\prime}$ be the corresponding vertex in $\mathcal{H}_{\CCX}(\mathbf{G})$. Let $m_{x,y}$ be a computation (message) that is executed between two cells $x$ and $y$ of $\CCX$ as part of the evaluation of function $\mathcal{M}_{\mathbf{G},W}$. Hence, the two cells $x$ and $y$ must have a corresponding non-zero entry in a matrix $G_i\in\mathbf{G}$. By the augmented Hasse graph definition, this non-zero entry corresponds to an edge in $\mathcal{H}_{\CCX}(\mathbf{G})$ between $x^{\prime}$ and $y^{\prime}$. Thus, the computation $m_{x,y}$ between the cells $x$ and $y$ of $\CCX$ can be carried out as the computation (message) $m_{x^{\prime},y^{\prime}}$ between the corresponding vertices $x^{\prime}$ and $y^{\prime}$ of $\mathcal{H}_{\CCX}(\mathbf{G})$.
\end{proof}

Propositions~\ref{Hasse-pushforward} and~\ref{Hasse} ensure that push-forward and merge node computations
can be realized on augmented Hasse graphs.
Theorem~\ref{Hasse-theorem} generalizes Propositions~\ref{Hasse-pushforward} and~\ref{Hasse},
stating that any computation on tensor diagrams is realizable on augmented Hasse graphs.

\begin{theorem}
\label{Hasse-theorem}
Any computation executed via a tensor diagram $\CCN_{\mathbf{G};\mathbf{W}}$ can be reduced to a corresponding computation on the augmented Hasse graph $\mathcal{H}_{\CCX}(\mathbf{G})$.
\end{theorem}

\begin{proof}
The conclusion follows directly from Propositions~\ref{hasse} and~\ref{Hasse-pushforward},
along with the fact that any tensor diagram can be realized in terms of the three elementary tensor operations.
\end{proof}

\begin{figure}[!t]
\begin{center}
\includegraphics[scale = 0.15, keepaspectratio = 0.150]{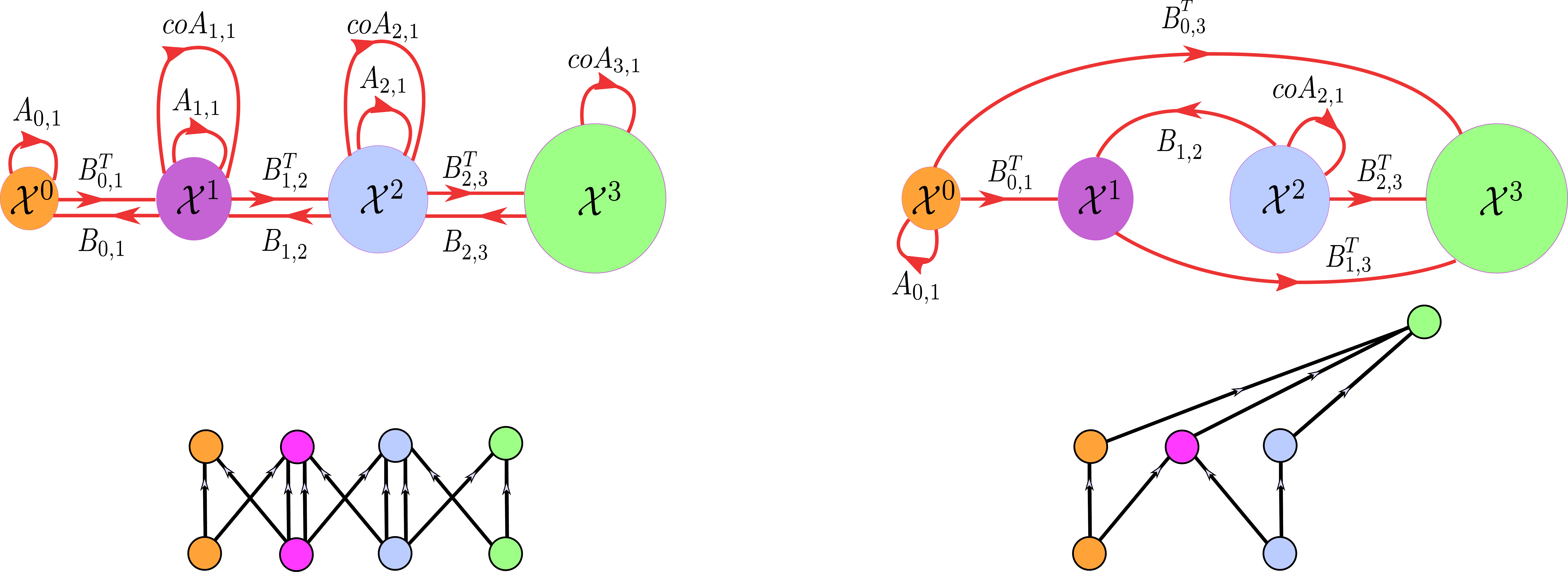}
\put(-300,-16){(a)}
\put(-100,-16){(b)}
\end{center}
\caption{Tensor diagrams of two CCNNs and their corresponding augmented Hasse graphs.
Edge labels are dropped from the tensor diagrams to avoid clutter,
as they can be inferred from the corresponding augmented Hasse graphs.
(a): A tensor diagram obtained from a higher-order message-passing scheme.
(b): A tensor diagram obtained by using the three elementary tensor operations
introduced in Section~\ref{sec:three}.}
\label{fig:hasse-diagram-examples}
\end{figure}

According to Theorem~\ref{Hasse-theorem}, a tensor diagram and its corresponding augmented Hasse graph
encode the same computations in alternative forms.
Figure~\ref{fig:hasse-diagram-examples} illustrates that the augmented Hasse graph
provides a computational summary of the associated tensor diagram representation of a CCNN.



\subsubsection{Augmented Hasse graphs and CC-pooling}

\begin{figure}[!t]
\begin{center}
\includegraphics[scale = 0.15, keepaspectratio = 0.150]{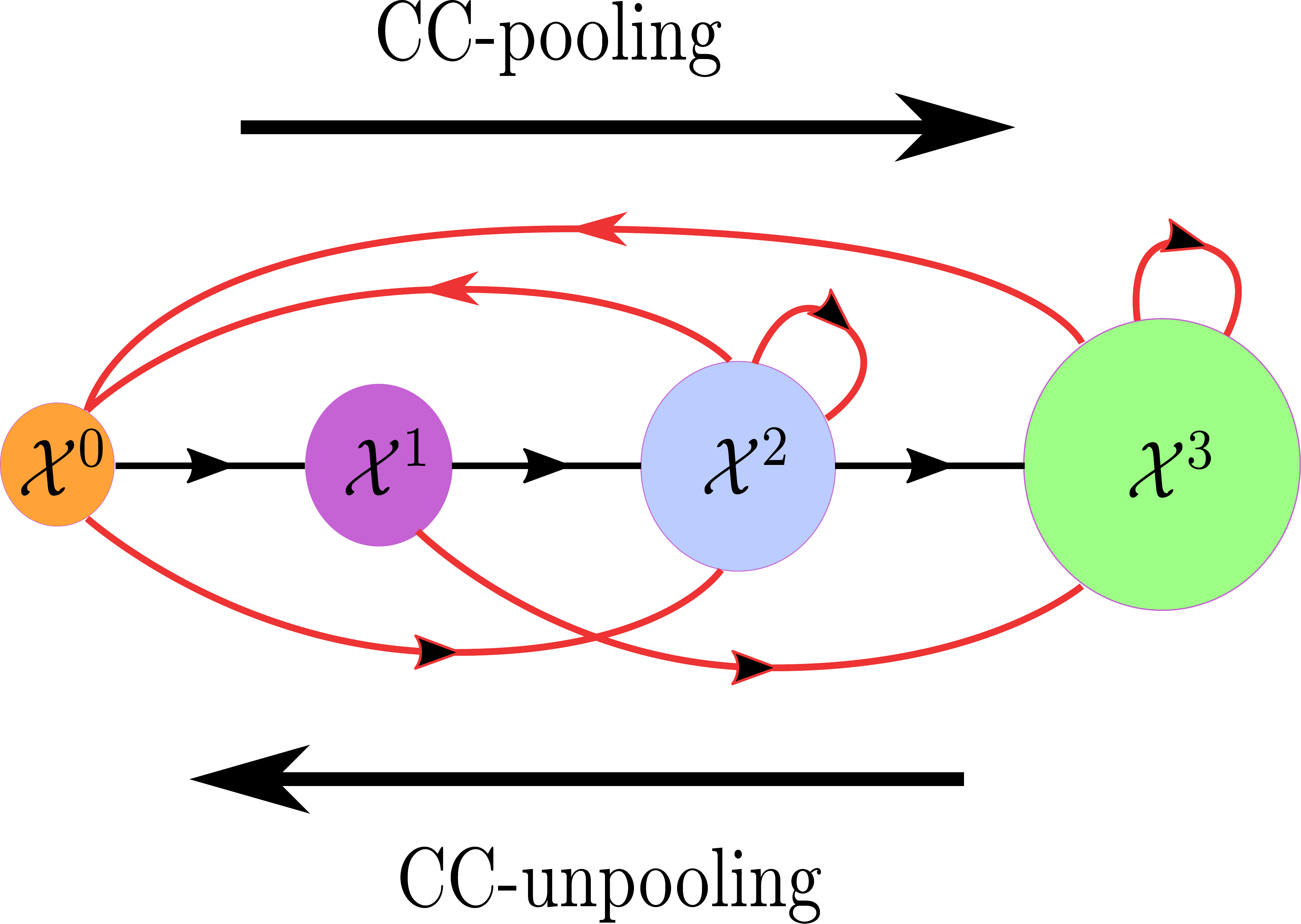}
\end{center}
\caption{CC (un)-pooling operations viewed on the augmented Hasse graph of a CC.
The vertices in the figure represent skeletons in the underlying CC. Black edges represent edges in the Hasse graph between these vertices, whereas red edges represent edges obtained from the augmented Hasse graph structure. CC-pooling corresponds to pushing a signal in the poset structure from lower-rank to higher-rank vertices, whereas CC-unpooling corresponds to pushing a signal in the poset structure from higher-rank to lower-rank vertices.\vspace{-3mm}}
\label{fig:hasse-diagram-pooling}
\end{figure}

The Hasse graph and its augmented version are graph representations of the poset structure of the underlying CC. It is instructive to interpret the (un)pooling operations (Definitions~\ref{pooling_exact_definition} and \ref{unpooling_exact_definition}) with respect to these graphs. The
CC-pooling operation of Definition~\ref{pooling_exact_definition} maps
a signal in the poset structure from lower-rank cells to higher-rank ones.
On the other hand, the CC-unpooling operation of Definition~\ref{unpooling_exact_definition} maps a signal in the opposite direction. Figure~\ref{fig:hasse-diagram-pooling} presents an example
of CC (un)-pooling operations visualized over the augmented Hasse graph of the underlying CC. 

\subsubsection{Augmented Hasse diagrams, message passing, and merge nodes}

The difference between constructing a CCNN using the higher-order message passing paradigm given in Section~\ref{sec:homp} versus using the three elementary tensor operations given in Section~\ref{sec:three} has been demonstrated in Section~\ref{merge and message passing}. In particular, Section~\ref{merge and message passing} mentions that merge nodes naturally allow for a more flexible computational framework in comparison to the higher-order message-passing paradigm. This flexibility manifests in terms of the underlying tensor diagram as well as the input for the network under consideration. The difference between tensor operations and higher-order message passing can also be highlighted with augmented Hasse graphs, as demonstrated in Figure~\ref{fig:hasse-diagram-examples}. Figure~\ref{fig:hasse-diagram-examples}(a) shows a tensor diagram obtained from a higher-order message-passing scheme on a CCNN. We observe two key properties of this CCNN: the initial input cochains are supported on all cells of all dimensions of the domain, and the CCNN updates all cochains supported on all cells of all dimensions of the domain at every iteration given a predetermined set of neighborhood functions. As a consequence, the corresponding augmented Hasse graph exhibits a uniform topological structure. In contrast, Figure~\ref{fig:hasse-diagram-examples}(b) shows a tensor diagram constructed using the three elementary tensor operations. As the higher-order message-passing rules do not impose constraints, the resulting augmented Hasse graph exhibits a more flexible structure.


\subsubsection{Higher-order representation learning}

The relation between augmented Hasse graphs and CCs given by Theorem~\ref{Hasse-theorem} suggests that many graph-based deep learning constructions have analogous constructions for CCs.
In this section, we demonstrate how \textit{higher-order representation learning} can be reduced to graph representation learning~\cite{hamilton2017representation}, as an application of certain CC computations as augmented Hasse graph computations.

The goal of graph representation is to learn a mapping that embeds the vertices, edges or subgraphs of a graph into a Euclidean space, so that the resulting embedding captures useful information about the graph. Similarly, higher-order representation learning~\cite{hajijcell} involves learning an embedding of various cells in a given topological domain into a Euclidean space, preserving the main structural properties of the topological domain. More precisely, given a complex $\mathcal{X}$, higher-order representation learning refers to learning a pair $(enc, dec)$ of functions, consisting of the \textit{encoder map} $enc \colon \mathcal{X}^k \to \mathbb{R}^d $ and the \textit{decoder map} $dec \colon \mathbb{R}^d \times \mathbb{R}^d \to \mathbb{R}$. The encoder function associates to every $k$-cell $x^k$ in $\mathcal{X}$ a feature vector $enc(x^k)$, which encodes the structure of $x^k$ with respect to the structures of other cells in $\mathcal{X}$. On the other hand, the decoder function associates to every pair of cell embeddings a measure of similarity, which quantifies some notion of relation between the corresponding cells. We optimize the trainable functions $(enc, dec)$ using a context-specific \textit{similarity measure} $sim \colon \mathcal{X}^k \times \mathcal{X}^k \to \mathbb{R} $ and an objective function
\begin{equation}
\label{loss}
\mathcal{L}_k=\sum_{ x^k \in \mathcal{X}^k     } l(  dec(  enc(x^{k}), enc(y^{k})),sim(x^{k},y^k)),
\end{equation}
where $l \colon \mathbb{R} \times \mathbb{R} \to \mathbb{R}$ is a loss function. 
The precise relation between higher-order and graph representation learning is given by Proposition~\ref{convert_graphtocc}.

\begin{proposition}
\label{convert_graphtocc}
Higher-order representation learning can be reduced to graph representation learning.
\end{proposition}
\vspace{-2.5mm}
\begin{proof}
Let $sim\colon \mathcal{X}^k \times \mathcal{X}^k \to \mathbb{R}$ be a similarity measure. The graph $\mathcal{G}_{\mathcal{X}^k}$ is defined as the graph whose vertex set corresponds to cells in $\mathcal{X}^k$ and whose edges correspond to cell pairs in $\mathcal{X}^k \times \mathcal{X}^k$ mapped to non-zero values by the function $sim$. Thus, the pair $(enc, dec)$ corresponds to a pair $(enc_{\mathcal{G}}, dec_{\mathcal{G}})$ of the form $enc_{\mathcal{G}}\colon \mathcal{G}_{\mathcal{X}^k} \to \mathbb{R}$ and $dec_{\mathcal{G}}\colon \mathbb{R}^d \times \mathbb{R}^d \to \mathbb{R}$. Thereby, learning the pair $(enc, dec)$ is reduced to learning the pair $(enc_{\mathcal{G}}, dec_{\mathcal{G}})$.
\end{proof}
\vspace{-2mm}

\href{https://github.com/pyt-team/TopoEmbedX}{TopoEmbedX}\footnote{https://github.com/pyt-team/TopoEmbedX}, one of our three contributed software packages, supports higher-order representation learning on cell complexes, simplicial complexes, and CCs. The main computational principle underlying TopoEmbedX is Proposition~\ref{convert_graphtocc}. Specifically, TopoEmbedX converts a given higher-order domain into a subgraph of the corresponding augmented Hasse graph, and then utilizes existing graph representation learning algorithms to compute the embedding of elements of this subgraph. Given the correspondence between the elements of the augmented Hasse graph and the original higher-order domain, this results in obtaining embeddings for the higher-order domain.

\begin{remark}
Following our discussion on Hasse graphs, and particularly the ability to transform computations on a CCNN to computations on a (Hasse) graph, one may argue that GNNs are sufficient and that there is no need for CCNNs. However, this is a misleading clue, in the sense that any computation can be represented by a computational graph. Applying a standard GNN over the augmented Hasse graph of a CC is not equivalent to applying a CCNN. This point will become clearer in Section~\ref{sec:equivariance}, where we introduce CCNN \emph{equivariances}.
\end{remark}

\subsection{On the equivariance of CCNNs}
\label{sec:equivariance}

Analogous to their graph counterparts, higher-order deep learning models, and CCNNs in particular, should always be considered in conjunction with their underlying \textit{equivariance}~\cite{bronstein2021geometric}. We now provide novel definitions for \emph{permutation} and \emph{orientation equivariance for CCNNs} and draw attention to their relations with conventional notions of equivariance defined for GNNs. 

\subsubsection{Permutation equivariance of CCNNs}

Motivated by Proposition~\ref{prop:structure}, which characterizes the structure of a CC, this section introduces permutation-equivariant CCNNs.
We first define the action of the permutation group on the space of cochain maps.

\begin{tcolorbox}
[width=\linewidth, sharp corners=all, colback=white!95!black]
\begin{definition}[Permutation action on space of cochain maps]
\label{perm}
Let $\CCX$ be a CC.  Define $\Sym(\CCX) = \prod_{i=0}^{\dim(\CCX)} \Sym(\CCX^k)$ the group of rank-preserving permutations of the cells of $\CCX$. Let $\mathbf{G}=\{G_k\}$ be a sequence of cochain maps defined on $\CCX$ with $G_k  \colon \mathcal{C}^{i_k}\to \mathcal{C}^{j_k}$, $0\leq i_k,j_k\leq \dim(\CCX)$.
Let $\mathcal{P}=(\mathbf{P}_i)_{i=0}^{\dim(\CCX)} \in \Sym(\CCX)$. Define the \textbf{permutation (group) action} of $\mathcal{P}$ on $\mathbf{G}$ by  
$\mathcal{P}(\mathbf{G}) = (\mathbf{P}_{j_k} G_{k} \mathbf{P}_{i_k}^T )_{i=0}^{\dim(\CCX)}$.
\end{definition}
\end{tcolorbox}

We introduce permutation-equivariant CCNNs in Definition~\ref{eqv},
using the group action given in Definition~\ref{perm}.
Definition~\ref{eqv} generalizes the relevant definitions in~\cite{roddenberry2021principled,schaub2021signal}. 
We refer the reader to~\cite{velivckovic2022message,joglwe} for a related discussion.
Hereafter, we use $\proj_k \colon \mathcal{C}^1\times \cdots \times \mathcal{C}^m \to \mathcal{C}^k$ to denote the standard $k$-th projection for $1\leq k \leq m$, defined via $\proj_k ( \mathbf{H}_{1},\ldots, \mathbf{H}_{k},\ldots,\mathbf{H}_{m})= \mathbf{H}_{k}$.

\begin{tcolorbox}
[width=\linewidth, sharp corners=all, colback=white!95!black]
\begin{definition}[Permutation-equivariant CCNN]
\label{eqv}
Let $\CCX$ be a CC and let $\mathbf{G}= \{G_k\}$ be a finite sequence of cochain maps defined on $\CCX$. Let $\mathcal{P}=(\mathbf{P}_i)_{i=0}^{\dim(\CCX)} \in \Sym(\CCX)$.
A CCNN of the form 
\begin{equation*} 
\CCN_{\mathbf{G};\mathbf{W}}\colon \mathcal{C}^{i_1}\times\mathcal{C}^{i_2}\times \cdots \times  \mathcal{C}^{i_m} \to \mathcal{C}^{j_1}\times\mathcal{C}^{j_2}\times \cdots \times \mathcal{C}^{j_n}
\end{equation*}
is called a \textbf{permutation-equivariant CCNN} if 
\begin{equation}
\label{eq:perm_equiv_ccnn}
\proj_k \circ \mathcal{\CCN}_{\mathbf{G};\mathbf{W}}(\mathbf{H}_{i_1},\ldots ,\mathbf{H}_{i_m})=\mathbf{P}_{k} \proj_k \circ \mathcal{\CCN}_{\mathcal{P}(\mathbf{G});\mathbf{W}}(\mathbf{P}_{i_1} \mathbf{H}_{i_1}, \ldots ,\mathbf{P}_{i_m} \mathbf{H}_{i_m})
\end{equation}
for all $1 \leq k\leq m $ and for any $(\mathbf{H}_{i_1},\ldots ,\mathbf{H}_{i_m}) \in \mathcal{C}^{i_1}\times\mathcal{C}^{i_2}\times \cdots \times  \mathcal{C}^{i_m}.$
\end{definition}
\end{tcolorbox}




Equation~\ref{eq:perm_equiv_ccnn} generalizes the corresponding notion of permutation equivariance of GNNs. Consider a graph with $n$ vertices and adjacency matrix $A$.  Denote a GNN on this graph by $\mathrm{GNN}_{A;W}$.  Let $H \in \mathbb{R}^{n \times k}$ be vertex features. Then $\mathrm{GNN}_{A;W}$ is permutation equivariant in the sense that for $P \in \Sym(n)$ we have 
$
P \,\mathrm{GNN}_{A;W}(H) = 
\mathrm{GNN}_{PAP^{T};W}(PH).
$

In general, working with Definition~\ref{eqv}
may be cumbersome. It is easier to characterize the equivariance in terms of merge nodes. To this end, recall that the height of a tensor diagram is the longest path from any source node to any target node. Proposition~\ref{simple} allows us to express tensor diagrams of height one in terms of merge nodes.

\begin{proposition}
	\label{simple}
	Let  $ \CCN_{\mathbf{G};\mathbf{W}}\colon \mathcal{C}^{i_1}\times\mathcal{C}^{i_2}\times \cdots \times  \mathcal{C}^{i_m} \to \mathcal{C}^{j_1}\times\mathcal{C}^{j_2}\times \cdots \times \mathcal{C}^{j_n}$ be a CCNN with a tensor diagram of height one. Then
	\begin{equation}
	\label{merge_lemma}
	\CCN_{\mathbf{G};\mathbf{W}}=(
 \mathcal{M}_{\mathbf{G}_{j_1};\mathbf{W}_1},\ldots,
 \mathcal{M}_{\mathbf{G}_{j_n};\mathbf{W}_n}),
	\end{equation}
	where $\mathbf{G}_k \subseteq \mathbf{G}$.
\end{proposition}


\begin{proof}
Let $ \CCN_{\mathbf{G};\mathbf{W}}\colon \mathcal{C}^{i_1}\times\mathcal{C}^{i_2}\times \cdots \times  \mathcal{C}^{i_m} \to \mathcal{C}^{j_1}\times\mathcal{C}^{j_2}\times \cdots \times \mathcal{C}^{j_n}$ be a CCNN with a tensor diagram of height one. Since the codomain of the function $\CCN_{\mathbf{G};\mathbf{W}}$ is $\mathcal{C}^{j_1}\times\mathcal{C}^{j_2}\times \ldots \times \mathcal{C}^{j_n}$, then $\CCN_{\mathbf{G};\mathbf{W}}$ is determined by $n$ functions $F_k\colon  \mathcal{C}^{i_1}\times\mathcal{C}^{i_2}\times \cdots \times  \mathcal{C}^{i_m} \to \mathcal{C}^{j_k}$ for $1 \leq k \leq n$. Since the height of the tensor diagram of $ \CCN_{\mathbf{G};\mathbf{W}}$ is one, then each function $F_k$ is also of height one and it is thus a merge node by definition. The result follows.
\end{proof}


Proposition~\ref{simple} states that every target node $j_k$ in a tensor diagram of height one is a merge node specified by the operators $\mathbf{G}_{j_k}$ formed by the labels of the edges with target $j_k$. 
Definition~\ref{eqv} introduces the general notion of permutation equivariance of CCNNs.
Definition~\ref{node_equivariance} introduces the notion of permutation-equivariant merge node.
Since a merge node is a CCNN, Definition~\ref{node_equivariance} is a special case of Definition~\ref{eqv}.


\begin{tcolorbox}
[width=\linewidth, sharp corners=all, colback=white!95!black]
\begin{definition}[Permutation-equivariant merge node]
	\label{node_equivariance}
	Let $\CCX$ be a CC and let $\mathbf{G}= \{G_k\}$ be a finite sequence of cochain operators defined on $\CCX$ with $G_k\colon C^{i_k}(\CCX)\to C^{j}(\CCX) $. Let $\mathcal{P}=(\mathbf{P}_i)_{i=0}^{\dim(\CCX)} \in \Sym(\CCX)$. We say that the merge node given in Equation~\ref{sum} is a \textbf{permutation-equivariant merge node} if
	\begin{equation}
	\mathcal{M}_{\mathbf{G};\mathbf{W}}(\mathbf{H}_{i_1},\ldots ,\mathbf{H}_{i_m})= \mathbf{P}_{j}  \mathcal{M}_{\mathcal{P}(\mathbf{G});\mathbf{W}}(\mathbf{P}_{i_1} \mathbf{H}_{i_1}, \ldots ,\mathbf{P}_{i_1} \mathbf{H}_{i_m})
	\end{equation}
	for any $(\mathbf{H}_{i_1},\ldots ,\mathbf{H}_{i_m}) \in \mathcal{C}^{i_1}\times\mathcal{C}^{i_2}\times \cdots \times  \mathcal{C}^{i_m}.$ 
\end{definition}
\end{tcolorbox}
\vspace{1mm}

\begin{proposition}
\label{height1}
Let $\CCN_{\mathbf{G};\mathbf{W}}\colon \mathcal{C}^{i_1}\times\mathcal{C}^{i_2}\times \cdots \times  \mathcal{C}^{i_m} \to \mathcal{C}^{j_1}\times\mathcal{C}^{j_2}\times \cdots \times \mathcal{C}^{j_n}$ be a CCNN with a tensor diagram of height one. Then $\CCN_{\mathbf{G};\mathbf{W}}$ is permutation equivariant if and only the merge nodes $\mathcal{M}_{\mathbf{G}_{j_k};\mathbf{W}_k}$ given in Equation~\ref{merge_lemma} are permutation equivariant for $ 1 \leq k \leq n$. 
\end{proposition}
\vspace{-3mm}

\begin{proof}
If a CCNN is of height one, then by Proposition~\ref{simple},
$\proj_k \circ \CCN_{\mathbf{G};\mathbf{W}}(\mathbf{H}_{i_1},\ldots ,\mathbf{H}_{i_m})= \mathcal{M}_{\mathbf{G}_{j_k};\mathbf{W}_k}.$ Hence, the result follows from the definition of merge node permutation equivariance (Definition~\ref{node_equivariance}) and the definition of CCNN permutation equivariance (Definition~\ref{eqv}).
\end{proof}
\vspace{-3mm}

Finally, Theorem~\ref{height2} characterizes the permutation equivariance of CCNNs in terms of merge nodes. From this point of view, Theorem~\ref{height2} provides a practical version of permutation equivariance for CCNNs.

\begin{theorem}
\label{height2}
A $\CCN_{\mathbf{G};\mathbf{W}}$ is permutation equivariant if and only if every merge node in $\CCN_{\mathbf{G};\mathbf{W}}$ is permutation equivariant.
\end{theorem}
\vspace{-3mm}

\begin{proof}
Proposition~\ref{height1} proves this fact for CCNNs of height one.
For CCNNs of height $n$, it is enough to observe that a CCNN of height $n$ is a composition of $n$ CCNNs of height one and that the composition of two permutation-equivariant networks is a permutation-equivariant network. 
\end{proof}

\begin{remark}
Our permutation equivariance assumes that all cells in each dimension are independently labeled with indices. However, if we label the cells in a CC with subsets of the powerset $\mathcal{P}(S)$ rather than with indices, then we only need to consider permutations of the powerset that are induced by permutations of the 0-cells in order to ensure permutation equivariance.
\end{remark}

\begin{remark}
\label{GNN_remark}
A GNN is equivariant in that a permutation of the vertex set of the graph and the input signal over the vertex set yields the same permutation of the GNN output. 
Applying a standard GNN over the augmented Hasse graph of the underlying CC is thus not equivalent to applying a CCNN. Although the message-passing structures are the same, the weight-sharing and permutation equivariance of the standard GNN and CCNN are different. In particular, Definition~\ref{maps} gives additional structure, which is not preserved by an arbitrary permutation of the vertices in the augmented Hasse graph. Thus, care is required in order to reduce message passing over a CCNN to message passing over the associated augmented Hasse graph. 
Specifically, one need only consider the subgroup of permutations of vertex labels in the augmented Hasse graph which are induced by permutations of 0-cells in the corresponding CC. Thus, there is merit in adopting the rich notions of topology to think about distributed, structured learning architectures, as topological constructions facilitate reasoning about computation in ways that are not within the scope of graph-based approaches.
\end{remark}

\begin{remark}
Note that Proposition~\ref{convert_graphtocc} does not contradict Remark~\ref{GNN_remark}. In fact, the computations described in Proposition~\ref{convert_graphtocc} are conducted on a particular subgraph of the Hasse graph whose vertices are the $k$-cells of the underlying complex. Differences between graph-based networks and TDL networks start to emerge particularly once different dimensions are considered simultaneously during computations.  
\end{remark}


\subsubsection{Orientation equivariance of CCNNs}
\label{orientation-eq}


When CC are reduced to regular cell complexes, then orientation equivariance can also be introduced to CCNNs.
Analogous to Definition~\ref{perm}, we introduce the following definition of orientation actions on CCs.

\begin{tcolorbox}
[width=\linewidth, sharp corners=all, colback=white!95!black]
\begin{definition}[Orientation action on space of diagonal-cochain maps]
\label{orientation}
Let $\CCX$ be a CC. Let $\mathbf{G}=\{G_k\}$ be a sequence of cochain operators defined on $\CCX$ with $G_k  \colon \mathcal{C}^{i_k}\to \mathcal{C}^{j_k}$, $0\leq i_k,j_k\leq \dim(\CCX)$.
Let $O(\CCX)$ be the group of tuples $\mathcal{D}=(\mathbf{D}_i)_{i=0}^{\dim(\CCX)}$ of diagonal matrices with diagonals $\pm 1$ of size $|\CCX^k| \times |\CCX^k| $ such that $\mathbf{D}_0=I$. Define the \textbf{orientation (group) action}
of $\mathcal{D}$ on $\mathbf{G}$ by
$\mathcal{D}(\mathbf{G}) = (\mathbf{D}_{j_k} G_{k} \mathbf{D}_{i_k})_{i=0}^{\dim(\CCX)}$.
\end{definition}
\end{tcolorbox}

We introduce orientation-equivariant CCNNs in Definition~\ref{OE},
using the group action given in Definition~\ref{orientation}.
Orientation equivariance for CCNNs (Definition~\ref{OE})
is put forward in analogous way
to permutation equivariance for CCNNs (Definition~\ref{eqv}).

\begin{tcolorbox}
[width=\linewidth, sharp corners=all, colback=white!95!black]
\begin{definition}[Orientation-equivariant CCNN]
\label{OE}
Let $\CCX$ be a CC and let $\mathbf{G}= \{G_k\}$ be a finite sequence of cochain operators defined on $\CCX$. Let $\mathcal{D} \in O(\CCX)$. A CCNN of the form 
	\begin{equation*} 
	\CCN_{\mathbf{G};\mathbf{W}}\colon \mathcal{C}^{i_1}\times\mathcal{C}^{i_2}\times \cdots \times  \mathcal{C}^{i_m} \to \mathcal{C}^{j_1}\times\mathcal{C}^{j_2}\times \cdots \times \mathcal{C}^{j_n}
	\end{equation*}
	is called an \textbf{orientation-equivariant CCNN} if
	\begin{equation}
	\proj_k \circ \CCN_{\mathbf{G};\mathbf{W}}(\mathbf{H}_{i_1},\ldots ,\mathbf{H}_{i_m})=\mathbf{D}_{k} \proj_k \circ \mathcal{\CCN}_{\mathcal{D}(\mathbf{G});\mathbf{W}}((\mathbf{D}_{i_1} \mathbf{H}_{i_1}, \ldots ,\mathbf{D}_{i_1} \mathbf{H}_{i_m}))
	\end{equation}
	for all $1 \leq k\leq m $ and for any $(\mathbf{H}_{i_1},\ldots ,\mathbf{H}_{i_m}) \in \mathcal{C}^{i_1}\times\mathcal{C}^{i_2}\times \cdots \times  \mathcal{C}^{i_m}.$
\end{definition}
\end{tcolorbox}

Propositions~\ref{simple} and \ref{height1} can be stated analogously for the orientation equivariance case. We skip stating these facts here, and only state the main theorem that characterizes the orientation equivariance of CCNNs in terms of merge nodes. 

\begin{theorem}
\label{height3}
A $\CCN_{\mathbf{G};\mathbf{W}}$
is orientation equivariant if and only if
every merge node in $\CCN_{\mathbf{G};\mathbf{W}}$ is orientation equivariant.
\end{theorem}
\vspace{-3mm}

\begin{proof}
The proof of Theorem~\ref{height3} is similar to the proof of Theorem~\ref{height2}.
\end{proof}
\vspace{-3mm}


\section{Implementation and numerical experiments}
\label{exp}

The proposed CCNNs can be used to construct different neural network architectures for diverse learning tasks. In this section, we demonstrate the generality and the efficacy of CCNNs by evaluating their predictive performance on shape analysis and graph learning tasks. In our geometric processing experiments, we compare CCNNs against state-of-the-art methods, which are highly engineered and trained towards specific tasks. Furthermore, we conduct our experiments on various data modalities commonly studied in geometric data processing, namely on point clouds and 3D meshes. We also perform experiments on graph data. In our experiments, we tune three main components: the choice of CCNN architecture, the learning rate, and the number of replications in data augmentation. We justify the choice of CCNN architecture for each learning task. We have implemented our pipeline in PyTorch, and ran the experiments on a single GPU NVIDIA GeForce RTX 3060 Ti using a Microsoft Windows back-end.


\subsection{Software: TopoNetX, TopoEmbedX, and TopoModelX}


All our software development and experimental analysis have been carried out using Python. We have developed three Python packages, which we have also used to run our experiments: \href{https://github.com/pyt-team/TopoNetX}{TopoNetX}\footnote{https://github.com/pyt-team/TopoNetX},\href{https://github.com/pyt-team/TopoEmbedX}{TopoEmbedX}\footnote{https://github.com/pyt-team/TopoEmbedX} and \href{https://github.com/pyt-team/TopoModelX}{TopoModelX}\footnote{https://github.com/pyt-team/TopoModelX}. TopoNetX supports the construction of several topological structures, including cell complex, simplicial complex, and combinatorial complex classes. These classes provide methods for computing boundary operators, Hodge Laplacians and higher-order adjacency operators on cell, simplicial, and combinatorial complexes, respectively. TopoEmbedX supports representation learning of higher-order relations on cell complexes, simplicial complexes, and combinatorial complexes. TopoModelX supports computing deep learning models defined on these topological domains.

In addition to our software package implementations, we have utilized PyTorch~\cite{paszke2017automatic} for training the neural networks reported in this section. Also, we have utilized Scikit-learn~\cite{scikit-learn} to compute the eigenvectors of the 1-Hodge Laplacians. The normal vectors of point clouds have been computed using the Point Cloud Utils package~\cite{point-cloud-utils}. Finally, we have used NetworkX~\cite{hagberg2008exploring} and HyperNetX~\cite{joslyn2021hypernetwork}, both in the development of our software packages as well as in our computations.


\subsection{Datasets}

In our evaluation of CCNNs, we use four datasets: the Human Body, the COSEG, the SHREC11, and a benchmark dataset for graph classification taken from~\cite{bianchi2020mincutpool}. A summary of these datasets follows.


\paragraph{Human Body segmentation dataset.}
The original Human Body segmentation dataset presented in~\cite{atzmon2018point} contains relatively large meshes with a size up to 12,000 vertices. The segmentation labels provided in this dataset are set per-face, and the segmentation accuracy is defined to be the ratio of the correctly classified faces over the total number of faces in the entire dataset. In this work, we use a simplified version of the original Human Body dataset, as provided by~\cite{hanocka2019meshcnn},
in which meshes have less than 1,000 nodes and segmentation labels are remapped to edges. We use this simplified version of the Human Body dataset for the shape analysis (i.e., mesh segmentation) task in Section~\ref{segmentation}.

\paragraph{COSEG segmentation dataset.}
The original COSEG dataset~\cite{wang2012active} contains eleven sets of shapes with ground-truth segmentation. In this work, we use a subset of the original COSEG dataset that contains relatively large sets of aliens, vases, and chairs. These three sets consist of 200, 300, and 400 shapes, respectively. We use this custom subset of the COSEG dataset for the shape analysis (i.e., mesh segmentation) task in Section~\ref{segmentation}.

\paragraph{SHREC11 classification dataset.}
SHREC 2011~\cite{lian2011shape}, abbreviated as SHREC11, is a large-scale dataset that contains 600 nonrigid deformed shapes (watertight triangle meshes) from 30 categories, where each category contains an equal number of objects. Examples of these categories include hand, lamp, woman, man, flamingo, and rabbit. 
The dataset is available as a training and test set consisting of 480 and 120 shapes, respectively.
We use the SHREC11 dataset for the shape analysis tasks in Sections~\ref{subsubsec:mesh_cl} and \ref{experiment:pooling}.

\paragraph{Benchmark dataset for graph classification.}
This dataset has graphs belonging to three different classes~\cite{bianchi2020mincutpool}. For each graph, the feature vector on each vertex (the $0$-cochain) is a one-hot vector of size five, and it stores the relative position of the vertices on the graph. This dataset has easy and hard versions. The easy version has highly connected graphs, while the hard version has sparse graphs. We use this dataset for the graph classification task in Section~\ref{subsubsec:graph_class}.

\subsection{Shape analysis: mesh segmentation and classification}

The CC structure used for the shape analysis experiments (mesh segmentation and classification) is simply induced by the triangulation of the meshes. Specifically, the $0$-, $1$-, and $2$-cells are the vertices, edges, and faces of the mesh, respectively. The matrices used for the CCNNs are $B_{0,1},~B_{0,2}$, their transpose matrices, and the (co)adjacency matrices $A_{1,1}$, $coA_{1,1}$, and $coA_{2,1}$.

A CCNN takes a vector of cochains as input features. For shape analysis tasks, we consider cochains, whose features are built directly from the vertex coordinates of the underlying mesh. We note that other choices (e.g., spectral-based cochains as in~\cite{mejia2017spectral}) can also be included. Our shape analysis tasks have three input cochains: the vertex, edge and face cochains. Each vertex cochain has two input features: the position and the normal vectors associated with the vertex. Similar to \cite{hanocka2019meshcnn},
each edge cochain consists of five features: the length of the edge, the dihedral angle, two inner angles, and two edge-length ratios for each face. Finally, each input face cochain consists of three input features: the face area, face normal, and the three face angles.

\begin{figure}[!t]
	\begin{center}
	\includegraphics[scale = 0.15, keepaspectratio = 0.20]{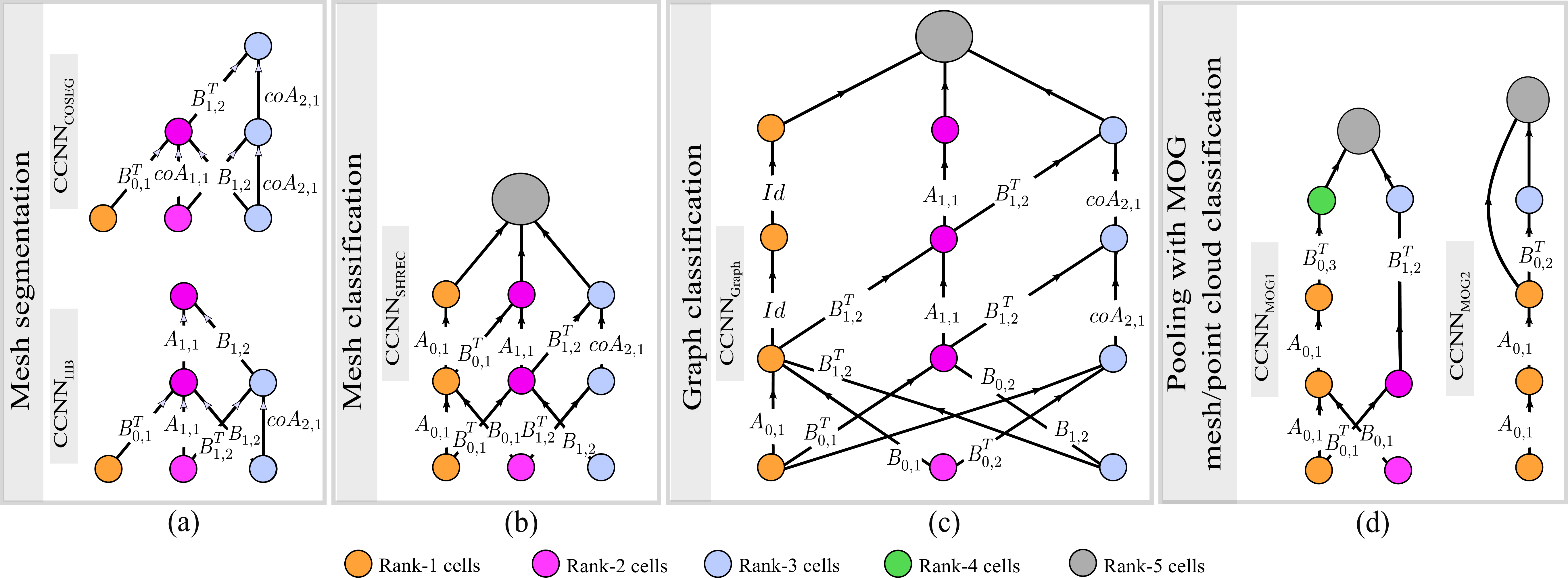}
	\end{center}
	\caption{The tensor diagrams of the CCNNs used in our experiments.
 (a): The CCNNs used in the mesh segmentation tasks. In particular, $\CCN_{HB}$ and $\CCN_{COSEG}$ are the architectures used on the Human Body dataset~\cite{atzmon2018point} and on the COSEG dataset~\cite{wang2012active}, respectively. (b): The mesh classification CCNN used on the SHREC11 dataset~\cite{lian2011shape}. (c): The graph classification CCNN used on the dataset provided in~\cite{bianchi2020mincutpool}. (d): The mesh/point cloud classification CCNNs used in conjunction with the MOG algorithm on the SHREC11 dataset.}
\label{fig:mesh_net}
\end{figure}

\subsubsection{Mesh segmentation}
\label{segmentation}
For the Human Body dataset~\cite{maron2017convolutional},
we built a CCNN that produces an edge class.
The tensor diagram of the architecture is shown in Figure~\ref{fig:mesh_net}(a). For the COSEG dataset~\cite{wang2012active},
we built a CCNN that combines our proposed feature vectors defined on vertices, edges, and faces to learn the final face class. The architecture uses incidence matrices as well as (co)adjacency matrices to construct a signal flow as demonstrated in Figure~\ref{fig:mesh_net}(b). Specifically, the tensor diagram displays three non-squared attention-blocks and three squared attention blocks. The depth of the model is chosen to be two, as indicated in Figure~\ref{fig:mesh_net}(b).

Note that the architectures chosen for the COSEG and for the Human Body datasets have the same number and types of building blocks;
compare Figures~\ref{fig:mesh_net}(a) and (b).
We use a random $85\%-15\%$ train-test split.
For both of these architectures, a softmax activation is applied to the output tensor. All our segmentation models are trained for $600$ epochs using a learning rate of $0.0001$ and the standard cross-entropy loss. These results are consistent across Human Body and Shape COSEG datasets

We test the proposed CCNNs on mesh segmentation using the Human Body~\cite{maron2017convolutional} and the Shape COSEG (vase, chair, and alien)~\cite{wang2012active} datasets. For each mesh in these datasets, the utilized CC structure is the one induced by the triangulation of the meshes, although other variations in the CC structure yield comparable results. Further, three $k$-cochains are constructed for $0\leq k \leq 2$ and are utilized in CCNN training. As shown in Table~\ref{tab:shape_analysis}, CCNNs
outperform three neural networks tailored to mesh analysis
(HodgeNet~\cite{smirnov2021hodgenet}, PD-MeshNet~\cite{milano2020primal} and MeshCCN~\cite{hanocka2019meshcnn})
on two out of four datasets, and are among the best two neural networks on all four datasets.

\paragraph{Architecture of $\CCN_{COSEG}$ and  $\CCN_{HB}$.} In $\CCN_{COSEG}$, as shown in Figure \ref{fig:mesh_net}(a), we choose a CCNN pooling architecture as given in Definition \ref{general_pooling_hoan}, which pushes signals from vertices, edges and faces, and aggregates their information towards the final face prediction class. We choose $\CCN_{HB}$ similarly, except that the predicted signal is an edge class. The reason for this choice is that the Human Body dataset~\cite{atzmon2018point} encodes the segmentation information on edges.  

\begin{table}[!t]
	\caption{Predictive accuracy on test sets related to shape analysis, namely on Human Body and COSEG (vase, chair, alien) datasets. Red and blue colors indicate best and second best results, respectively. The results reported here are based on the $\CCN_{COSEG}$ and $\CCN_{HB}$ architectures, which are visualized in Figure~\ref{fig:mesh_net}(a). In particular, the result for $\CCN_{HB}$ is reported in the first column, whereas the results for $\CCN_{COSEG}$ are reported in the second, third and forth columns.} 
	\label{tab:shape_analysis} 
	\begin{center}
	\begin{small}
		\begin{tabular}{l|cccc}
			\hline
			\multicolumn{1}{c|}{\multirow{2}{*}{Method}} &
			\multicolumn{4}{c}{Segmentation tasks} \\ \cline{2-5}
			& \multicolumn{1}{c|}{Human Body}     & \multicolumn{1}{c|}{COSEG vase}           & \multicolumn{1}{c|}{COSEG chair}          & COSEG alien        \\ \hline
			HodgeNet                & \multicolumn{1}{c|}{85.03}          & \multicolumn{1}{c|}{90.30}          & \multicolumn{1}{c|}{95.68}          & 96.03        \\ 
			PD-MeshNet              &
			\multicolumn{1}{c|}{\textcolor{blue}{85.61}}          &
			\multicolumn{1}{c|}{\textcolor{purple}{95.36}} &
			\multicolumn{1}{c|}{\textcolor{blue}{97.23}}          &
			\textcolor{purple}{98.18} \\
			MeshCNN                 & \multicolumn{1}{c|}{85.39}          & \multicolumn{1}{c|}{92.36}          & \multicolumn{1}{c|}{92.99}          & \textcolor{blue}{96.26}      \\ 
			CCNN                 &
			\multicolumn{1}{c|}{\textcolor{purple}{87.30}}
			& \multicolumn{1}{c|}{ \textcolor{blue}{93.40} }          &
			\multicolumn{1}{c|}{\textcolor{purple}{98.30}} & 93.70        \\ \hline
		\end{tabular}\vspace{-4mm}
	\end{small}
	\end{center}
\end{table}

\begin{table}[!t]
\caption{Predictive accuracy on the SHREC11 test dataset. The left and right column report the mesh and point cloud classification results, respectively. The CCNN for mesh classification is $\CCN_{SHREC}$, shown in Figure \ref{fig:mesh_net}(b), while the CCNN for point cloud classification is $\CCN_{MOG2}$, shown in Figure \ref{fig:mesh_net}(d).} 
\label{tab:shrec} 
\begin{center}
\begin{small}
\begin{tabular}{l|c|c}
\hline
\multicolumn{1}{c|}{\multirow{2}{*}{Method}} &
\multicolumn{2}{c}{Classification tasks} \\ \cline{2-3}
& \multicolumn{1}{c|}{Mesh}     & \multicolumn{1}{c}{Point cloud}        \\ \hline
HodgeNet & 99.10 &  94.70  \\ 
PD-MeshNet & \multicolumn{1}{c|}{\textcolor{purple}{99.70}}  &  \multicolumn{1}{c}{\textcolor{purple}{99.10}} \\
MeshCNN & 98.60 &  91.00 \\ 
CCNN & \multicolumn{1}{c|}{\textcolor{blue}{99.17}} & \multicolumn{1}{c}{\textcolor{blue}{95.20}} \\ \hline
\end{tabular}\vspace{-4mm}
\end{small}
\end{center}
\end{table}


\subsubsection{Mesh and point cloud classification}
\label{subsubsec:mesh_cl}

We evaluate our method on mesh classification using the SHREC11 dataset~\cite{lian2011shape} based on the same cochains and CC structure used in the segmentation experiment of Section \ref{segmentation}. The CCNN architecture for our mesh classification task, denoted by $\CCN_{SHREC}$, is demonstrated in Figure~\ref{fig:mesh_net}(b). The final layer of $\CCN_{SHREC}$, depicted as a grey node in Figure~\ref{fig:mesh_net}(b), is a simple pooling operation that sums all embeddings of the CC after mapping them to the same Euclidean space. The $\CCN_{SHREC}$ is trained for 40 epochs with both tanh and identity activation functions using a learning rate of $0.005$ and the standard cross-entropy loss. We use anisotropic scaling and random rotations for data augmentation. Each mesh is augmented 30 times, is centered around the vertex center of the mass, and is rescaled to fit inside the unit cube.

The $\CCN_{SHREC}$ with identity activations and $\tanh$ activations achieve predictive accuracies of 
$96.67\%$ and $99.17\%$, respectively. Table~\ref{tab:shrec} shows that CCNNs
outperform two neural networks tailored to mesh analysis (HodgeNet and MeshCCN),
being the second best model behind PD-MeshNet in mesh and point cloud classification.
It is worth mentioning that the mesh classification CCNN requires a significantly lower number of epochs to train (40 epochs) as compared to the mesh segmentation CCNNs (600 epochs).

\begin{table}[t!]
	\caption{Predictive accuracy on the test set of~\cite{bianchi2020mincutpool} related to graph classification; red and blue colors indicate best and second best results, respectively. All results are reported using the $\CCN_{Graph}$ architecture shown in Figure \ref{fig:mesh_net}(c).}
	\label{wrap-tab: Human Body test} 
	\begin{center}
	\begin{small}
		\begin{tabular}{c|c|c|c|c|c|c|c}
			\hline
			\multirow{2}{*}{Dataset} &
			\multicolumn{7}{c}{Method} \\ \cline{2-8}
			& Graclus & NDP & DiffPool & Top-K & SAGPool & MinCutPool & $\CCN_{Graph}$  \\ \hline
			Easy & 97.81 & 97.93 & 98.64 & 82.47 & 84.23 &
			\textcolor{purple}{99.02} & \textcolor{blue}{98.90} \\
			Hard & 69.08 & 72.67 & 69.98 & 42.80 & 37.71 & \textcolor{blue}{73.80} &
			\textcolor{purple}{75.79} \\
			\hline
		\end{tabular}\vspace{-4mm}
	\end{small}
	\end{center}
\end{table}

\paragraph{Architecture of $\CCN_{SHREC}$.} The $\CCN_{SHREC}$ has two layers and is chosen as a pooling CCNN in the sense of Definition~\ref{general_pooling_hoan}, similar to $\CCN_{COSEG}$ and $\CCN_{HB}$. The main difference is that the final layer of $\CCN_{SHREC}$, represented by the grey point in Figure~\ref{fig:mesh_net}(b), is a global pooling function that sums all embeddings of all dimensions (zero, one and two) of the underlying CC after mapping them to the same Euclidean space.

\subsubsection{Graph classification}
\label{subsubsec:graph_class}

For the graph classification task, we use the graph classification benchmark provided in~\cite{bianchi2020mincutpool}; the dataset consists of graphs with three different labels. For each graph, the feature vector on each vertex (the $0$-cochain) is a one-hot vector of size five, and it stores the relative position of the vertex on the graph. 
To construct the CC structure, we use the $2$-clique complex of the input graph. We then proceed to build the CCNN for graph classification, denoted by $\CCN_{Graph}$, which is visualized in Figure~\ref{fig:mesh_net}(c). The matrices used for the construction of $\CCN_{Graph}$ are
$B_{0,1},~B_{1,2},~B_{0,2}$, their transpose matrices, and the (co)adjacency matrices $A_{0,1},A_{1,1},~coA_{2,1}$. The cochains of $\CCN_{Graph}$ are constructed as follows. For each graph in the dataset, we set the $0$-cochain to be the one-hot vector of size $5$ provided by the dataset. This one-hot vector stores the relative position of the vertex on the graph. We also construct the $1$-cochain and $2$-cochain on the $2$-clique complex of the graph by considering the coordinate-wise max value of the one-hot vectors attached to the vertices of each cell. The input to $\CCN_{graoh}$ consists of the 0-cochain provided as a part of the dataset as well as the constructed 1 and 2-cochains. The grey node in Figure~\ref{fig:mesh_net}(c) indicates a simple mean pooling operation. We train this network with a learning rate of 0.005 and no data augmentation.

Table~\ref{wrap-tab: Human Body test} reports the results on the \textit{easy} and the \textit{hard} versions of the datasets\footnote{
The difficulty in these datasets is controlled by the compactness degree of the graph clusters; clusters in the `easy' data have more in-between cluster connections, while clusters in the `hard' data are more isolated~\cite{bianchi2020mincutpool}.},
and compares them to six state-of-the-art GNNs.
As shown in Table~\ref{wrap-tab: Human Body test},
CCNNs outperform all six GNNs on the hard dataset,
and five of the GNNs on the easy dataset.
The proposed CCNN outperforms MinCutPool on the hard dataset,
while it attains comparable performance to MinCutPool on the easy dataset.

\paragraph{Architecture of  $\CCN_{Graph}$.} In the $\CCN_{Graph}$ displayed in Figure~\ref{fig:mesh_net}(c) we choose a CCNN pooling architecture as given in Definition~\ref{general_pooling_hoan} that pushes signals from vertices, edges and faces, and aggregate their information towards the higher-order cells before making making the final prediction. For the dataset of~\cite{bianchi2020mincutpool}, we experiment with two architectures; the first one is identical to the $\CCN_{SHREC}$ shown in Figure \ref{fig:mesh_net}(b), and the second one is the $\CCN_{Graph}$ shown in Figure~\ref{fig:mesh_net}(c). We report the results for $\CCN_{Graph}$, as it provides superior performance. Note that when this neural network is conducted on an underlying simplicial complex, the neighborhood matrices $B_{0,1}$ and $B_{1,3}$ are typically not considered, hence the CC-structure equipped with these additional incidence matrices improves the generalization performance of the $\CCN_{Graph}$.

\begin{figure}[!t]
\begin{center}
\includegraphics[scale = 0.17, keepaspectratio = 0.20]{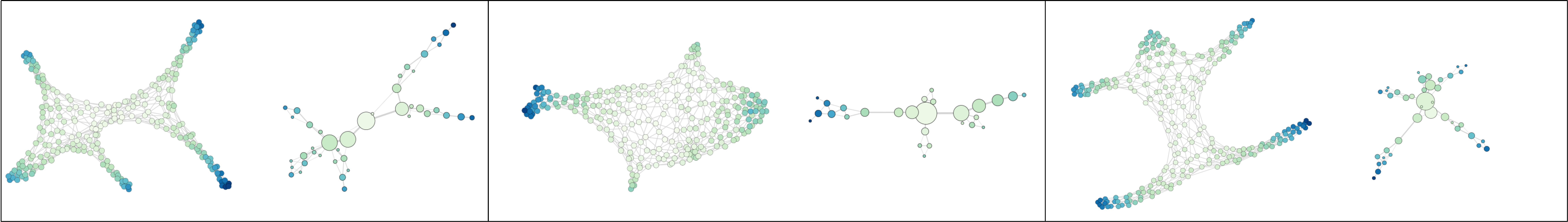}
\end{center}
\caption{Examples of applying the MOG algorithm on the SHREC11 dataset~\cite{lian2011shape}. In each figure, we show the original mesh graph on the left and the mapper graph on the right. The scalar function chosen for the MOG algorithm is the average geodesic distance (AGD). We observe that the pooled mapper graph has similar overall shape to the original graphs.}
\label{pooling_examples}
\end{figure}

\subsection{Pooling with mapper on graphs and data classification}
\label{experiment:pooling}

We perform experiments to measure the effectiveness of the MOG pooling strategy discussed in Section~\ref{mapper}. Recall that the MOG algorithm requires two pieces of input: the 1-skeleton of a CC $\mathcal{X}$, and a scalar function on the vertices of $\mathcal{X}$. Our choice for the input scalar function is the average geodesic distance (AGD)~\cite{KimLipmanChen2010}, which is suitable for shape detection as it is invariant to reflection and rotation. For two entities $u$ and $v$ on a graph, the geodesic distance between $u$ and $v$, denoted by $d(v,u)$, is computed using Dijkstra's shortest path algorithm. The AGD is given by the following equation:
\begin{equation}
\label{agd}
AGD(v)=\frac{1}{|V|}\sum_{u\in V}d(v,u).
\end{equation}

From Equation~\ref{agd}, it is immediate that the vertices near the center of the graph are likely to have low function values, while points on the periphery are likely to have high values. This observation has been utilized to study graph symmetry~\cite{KimLipmanChen2010}, and it provides a justification for selecting the AGD for the MOG pooling strategy. Figure~\ref{pooling_examples} presents a few examples of applying the MOG pooling strategy using AGD on the SHREC11 dataset.

In order to demonstrate the effectiveness of our MOG pooling approach, we conduct three experiments on the SHREC11 dataset:
mesh classification based on CC-pooling with input vertex and edge features (Section~\ref{pooling_mapper1}),
mesh classification based on CC-pooling with input vertex features only (Section~\ref{mog2}),
and point cloud classification based on CC-pooling with input vertex features only (Section~\ref{mog3}).
The experiments in Sections~\ref{pooling_mapper1} and~\ref{mog2} utilize the mesh structure in the SHREC11 dataset,
whereas the experiment in Section~\ref{mog3} utilizes its own point cloud version. In particular, we choose two simple CCNN architectures shown in Figure~\ref{fig:mesh_net}(d), denoted by $\CCN_{MOG1}$ and $\CCN_{MOG2}$, as opposed to the more complicated architecture of $\CCN_{SHREC}$ in Figure \ref{fig:mesh_net}(b). The main difference between $\CCN_{MOG1}$ and $\CCN_{MOG2}$ is the choice of the input feature vectors as described next.

\subsubsection{Mesh classification: CC-pooling with input vertex and edge features}
 \label{pooling_mapper1}
 
 In this experiment, we consider the vertex feature vector to be the position concatenated with the normal vectors for each vertex in the underlying mesh. For the edge features, we compute the first ten eigenvectors of the 1-Hodge Laplacian~\cite{eckmann1944harmonische,dodziuk1976finite} and attach a 10-dimensional feature vector to the edges of the underlying mesh. The CC that we consider here is 3-dimensional, as it consists of the triangular mesh (vertices, edges and faces) and 
 of 3-cells. The 3-cells are obtained using the MOG algorithm, and are used for augmenting each mesh. We calculate the 3-cells via the MOG algorithm using the AGD scalar function as input. We conduct this experiment using the CCNN defined via the tensor diagram $\CCN_{MOG1}$ given in Figure \ref{fig:mesh_net}(d). During training, we augment each mesh with ten additional meshes, with each of these additional meshes being obtained by a random rotation as well as $0.1\%$ noise perturbation to the vertex positions. We train $\CCN_{MOG1}$ for 100 epochs using a learning rate of $0.0002$  and the standard cross-entropy loss, and obtain an accuracy of $98.1 \%$. While the accuracy of $\CCN_{MOG1}$ is lower than the one we report for $\CCN_{SHREC}$ ($99.17\%$) in Table~\ref{tab:shrec}, we note that $\CCN_{MOG1}$ requires a significantly smaller number of replications for mesh augmentation to achieve a similar accuracy
 ($\CCN_{MOG1}$ requires 10, whereas $\CCN_{SHREC}$ required 30 replications).

\paragraph{Architecture of $\CCN_{MOG1}$.} The tensor diagram $\CCN_{MOG1}$ of Figure~\ref{fig:mesh_net}(d) corresponds to a pooling CCNN. In particular, $\CCN_{MOG1}$ pushes forward the signal towards two different higher-order cells: the faces of the mesh as well as the 3-cells obtained from the MOG algorithm.

\subsubsection{Mesh classification: CC-pooling with input vertex features only}
\label{mog2}

In this experiment, we consider the position and the normal vectors of the input vertices. The CC structure that we consider is the underlying graph structure obtained from each mesh; i.e., we only use the vertices and the edges, and ignore the faces. We augment this structure by 2-cells obtained via the MOG algorithm using the AGD scalar function as input.
We choose the network architecture to be relatively simpler than $\CCN_{MOG1}$, and report it in Figure \ref{fig:mesh_net}(d) as $\CCN_{MOG2}$. During training we augment each mesh with $10$ additional meshes, with each of these additional meshes being obtained by a random rotation as well as $0.05\%$ noise perturbation to the vertex positions. We train $\CCN_{MOG2}$ for 100 epochs using a learning rate of $0.0003$ and the standard cross-entropy loss, and obtain an accuracy of $97.1 \%$.

\paragraph{Architecture of  $\CCN_{MOG2}$ for mesh classification.} The tensor diagram $\CCN_{MOG2}$ of Figure~\ref{fig:mesh_net}(d) corresponds to a pooling CCNN. In particular, $\CCN_{MOG2}$ pushes forward the signal towards a single 2-cell obtained from the MOG algorithm. Observe that the overall architecture of $\CCN_{MOG2}$ is similar in principle to AlexNet~\cite{krizhevsky2017imagenet}, where convolutional layers are followed by pooling layers.

\subsubsection{Point cloud classification: CC-pooling with input vertex features only}
\label{mog3}

In this experiment, we consider point cloud classification on the SHREC11 dataset. The setup is similar in principle to the one studied in Section~\ref{mog2} where we consider only the features supported on the vertices of the point cloud as input. Specifically, for each mesh in the SHREC11 dataset, we sample $1,000$ points from the surface of the mesh. Additionally, we estimate the normal vectors of the resulting point clouds using the Point Cloud Utils package~\cite{point-cloud-utils}. To build the CC structure, we first consider the $k$-nearest neighborhood graph obtained from each point cloud using $k=7$. We then augment this graph by 2-cells obtained via the MOG algorithm using the AGD scalar function as input.
We train the $\CCN_{MOG2}$ shown in Figure \ref{fig:mesh_net}(d). During training, we augment each point cloud with $12$ additional instances, each one of these instances being obtained by random rotation. We train $\CCN_{MOG2}$ for 100 epochs using a learning rate of $0.0003$ and the standard cross-entropy loss, and obtain an accuracy of $95.2\%$ (see Table~\ref{tab:shrec}).


\subsection{Ablation studies}

In this section, we perform two ablation studies. The first ablation study reveals that pooling strategies in CCNNs have a crucial effect on predictive performance. The second ablation study demonstrates that CCNNs have better predictive capacity than GNNs; the advantage of CCNNs arises from their topological pooling operations and from their ability to learn from topological features.

\paragraph{Pooling strategies in CCNNs.} To evaluate the impact of the choice of pooling strategy on predictive performance, we experiment with two pooling strategies using the SHREC11 classification dataset. The first pooling strategy is the MOG algorithm described in Section~\ref{experiment:pooling}; the results of this pooling strategy based on $\CCN_{MOG2}$ are discussed in Section~\ref{mog2} ($97.1\%$). The second pooling strategy is briefly described as follows. For each mesh, we consider the 2-dimensional CC obtained by considering each 1-hop neighborhood to be the 1-cells in the CC and each 2-hop neighborhood to be the 2-cells in the CC. We train $\CCN_{MOG2}$, and obtain an accuracy of $89.2\%$, which is lower than $97.1\%$. These experiments suggest that the choice of pooling strategy has a crucial effect on predictive performance.

\paragraph{Comparing CCNNs to GNNs in terms of predictive performance.} Observe that $\CCN_{SHREC}$ has topological features of dimension one and two as inputs. On the other hand, $\CCN_{MOG2}$ has only vertex features as input, but it learns the higher-order cell latent features by using the push-forward operation that pushes the signal from 0-cells to the 2-cells obtained from the MOG algorithm. In both cases, using a higher-order structure is essential for improving predictive performance, even though two different strategies towards exploiting the higher-order structures are utilized. To support our claim, we run an experiment in which we replace the pooling layer in $\CCN_{MOG2}$ by the cochain operator induced by $A_{0,1}$, effectively rendering the neural network as a GNN. In this setting, using the same setup as in experiment~\ref{mog2}, we obtain an accuracy of $84.56\%$. This experiment reveals the performance advantages of employing higher-order structures, either by utilizing the input topological features supported on higher-order cells or via pooling strategies that augment higher-order cells.

\section{Related work}
\label{related}

Topological deep learning (TDL) has recently emerged as a new research frontier that lies at the intersection of several areas, including geometric and topological machine learning, and network science. To demonstrate where TDL fits in the existing literature, we review a broad spectrum of prior works, and categorize them into graph-based models, higher-order deep learning models, graph-based pooling, attention-based models, and applied algebraic topology.

\subsection{Graph-based models} 

Graph-based models have been widely used for modeling pairwise interactions (edges) between elements (vertices) of different systems, including social systems (e.g., social network analysis) and biological systems (e.g., protein-protein interactions), see~\cite{knoke2019social,jha2022prediction}. Based on their edge or vertex properties, graphs can be classified as unweighted graphs (unweighted edges), weighted graphs (weighted edges), signed graphs (signed edges), undirected or directed graphs (undirected or directed edges), and spatio-temporal graphs (spatio-temporal vertices), as discussed in~\cite{wu2020comprehensive,goyal2018graph}. Each of these graph types can be combined with neural networks to form graph neural networks and model different interactions in various systems~\cite{wu2020comprehensive,goyal2018graph}. For example, unweighted and undirected graph-based models have been used for omic data mapping~\cite{amar2014constructing} and mutual friendship detection in social networks~\cite{tabassum2018social}; weighted graph-based models have been widely used with systems related to traffic forecasting~\cite{zhang2018kernel,halaoui2010smart} and epidemiological modeling/forecasting~\cite{linka2020outbreak,manriquez2021protection}; signed graph-based models are suitable for tasks such as segmentation~\cite{bailoni2022gasp} and clustering~\cite{kunegis2010spectral, gallier2016spectral}; spatio-temporal graph-based models can describe systems that are spatio-temporal in nature, such as human activity and different types of motion~\cite{yan2018spatial, plizzari2021spatial, bhattacharya2020step}. 

As graph-based approaches that utilize single-layer or monolayer graphs cannot model multiple types of relations between vertices in a network~\cite{wu2020comprehensive,goyal2018graph}, multilayer or multiplex networks have been proposed~\cite{chang2022graphrr,zhang2020multiplex,kivela2014multilayer}. Similar to monolayer graphs, multiplex networks contain vertices and edges, but the edges exist in separate layers, where each layer represents a specific type of interaction or relation. Multiplex networks have been used in various applications, including multilayer modeling of the human brain~\cite{de2017multilayer,anand2023hodge} and online gaming~\cite{chang2022graphrr}. All these types of networks can only model pairwise relations between vertices, motivating the need for higher-order networks, as discussed in Section~\ref{subsec:utility}.

\subsection{Higher-order deep learning models}


In recent years, there has been an increasing interest in higher-order networks~\cite{mendel1991tutorial,battiston2020networks,bick2021higher} due to the ability of these networks to adequately capture higher-order interactions. Hodge-theoretic approaches, message passing schemes, and skip connections have been developed for higher-order networks in the signal processing and deep learning literature.

A Hodge-theoretic approach~\cite{lim2020hodge}
over simplicial complexes has been introduced
by~\cite{schaub2021signal,barbarossa2020topological}. This effort has been extended to hypergraphs by~\cite{barbarossa2016introduction,schaub2021signal} and to cell complexes by~\cite{sardellitti2021topological,roddenberry2021signal}. The work of~\cite{roddenberry2019hodgenet} has defined an edge-based convolutional neural network by exploiting the 1-Hodge Laplacian operator for linear filtering~\cite{barbarossa2018learning,schaub2018denoising,barbarossa2020topological,barbarossa2020topologicalmag,schaub2021signal}.


Convolutional operators and message-passing algorithms have been developed for higher-order neural networks. For example, a convolutional operator on hypergraphs has been proposed by~\cite{jiang2019dynamic,feng2019hypergraph,arya2018exploiting} and
has been investigated further
by~\cite{wu2022hypergraph,bai2021multi,jiang2019dynamic,bai2021hypergraph,gao2020hypergraph,gong2023generative,giusti2022cell}. A unifying framework for learning on graphs and hypergraphs has been proposed recently in~\cite{huang2021unignn}. The authors in~\cite{gao2022hgnn} have introduced the so-called general hypergraph neural networks, which constitute a multi-modal/multi-type data correlation modeling framework. As for message passing on complexes, the work of~\cite{hajijcell} has introduced a higher-order message-passing framework that encompasses those proposed by~\cite{ebli2020simplicial,bunch2020simplicial,gilmer2017neural,hayhoe2022stable} and has utilized various local neighborhood aggregation schemes. In~\cite{mitchell2022topological}, recurrent simplicial neural networks have been proposed and applied to trajectory prediction. The authors in~\cite{calmon2022higher} have addressed the challenge of processing signals supported on multiple cell dimensions concurrently, by introducing a coupling multi-signal approach on higher-order networks that utilizes the Dirac operator.  Several simplicial and cellular neural networks have been introduced recently, including~\cite{burnssimplicial,sardellitti2021topological,bodnar2021weisfeiler,roddenberry2021signal,sardellitti2022topological,battiloro2023topological,yang2023convolutional}. For more details, the reader is referred to the recent survey of~\cite{mathilde2023} on TDL.


A generalization of skip connections~\cite{he2016deep,ronneberger2015u} to simplicial complexes has been introduced by~\cite{hajij2022high}, which allows the training of higher-order deep neural networks. The authors in~\cite{morris2019weisfeiler} have proposed a higher-order graph neural network that takes into account higher-order graph structures at multiple scales. While these methods allow for multi-way hierarchical coupling, the coupling is isotropic and weight differences within a particular multi-way connection can not be learned. These limitations can be alleviated by attention-based models.

Higher-order models have achieved promising performance in several real-world applications, including link prediction~\cite{chen2021bscnets,hajij2022high,piaggesi2022effective}, action recognition~\cite{wang2023survey}, visual classification~\cite{shi2018hypergraph}, optimal homology generator detection~\cite{keros2021dist2cycle}, time series~\cite{santoro2023higher}, dynamical systems~\cite{majhi2022dynamics}, spectral clustering~\cite{reddy2023clustering}, node classification \cite{hajij2022high}, and trajectory prediction~\cite{roddenberry2021principled,benson2018simplicial}.


\vspace{-1mm}
\subsection{Attention-based models}
Real-world relational data is large, unstructured, sparse and noisy.
As a result, graph neural networks (GNNs)
may learn suboptimal data representations, and therefore may exhibit compromised performance~\cite{wu2020comprehensive,dai2021nrgnn,asif2021graph}. To address these issues, various attention mechanisms~\cite{chaudhari2021attentive} have been incorporated in GNNs, which allow to learn neural architectures that detect the most relevant parts of a given graph while ignoring irrelevant parts. Based on the used attention mechanism, existing graph attention approaches can be divided into weight-based attention, similarity-based attention,
and attention-guided walk~\cite{boaz2019}.

The majority of attention-based mechanisms, with the exception of~\cite{giusti2022cell,goh2022simplicial,giusti2022simplicial,bai2021hypergraph,kim2020hypergraph,georgiev2022heat}, are designed for graphs. For example, the attention model proposed by~\cite{goh2022simplicial} is a 
generalization of the graph attention model of~\cite{velickovic2017graph}. In~\cite{giusti2022simplicial}, the authors have utilized
a model based on Hodge decomposition, similar to the one suggested in~\cite{roddenberry2021principled}, to introduce an attention model for simplicial complexes. The hypergraph attention models introduced in~\cite{bai2021hypergraph,kim2020hypergraph} provide alternative generalizations
of the graph attention model of~\cite{velickovic2017graph}. 
The aforementioned attention models
neither allow nor combine higher-order attention blocks of entities of different dimensions.
This limits the space of neural architectures and the scope of applications of existing attention models.

\vspace{-1mm}
\subsection{Graph-based pooling}\vspace{-1mm}
Several attempts have been made to emulate the success of image-based pooling layers in the context of graphs. Some of the early work employs popular graph clustering algorithms~\cite{kushnir2006fast,dhillon2007weighted} to achieve graph-based pooling architectures~\cite{bruna2013spectral}. Coarsening operations have been applied to graphs
to attain the invariance properties needed in learning tasks~\cite{ying2018hierarchical,mesquita2020rethinking,gao2021topology,mesquita2020rethinking}. The current state-of-the-art graph-based pooling approaches mostly rely on dynamically learning the pooling needed for the learning task~\cite{grattarola2022understanding}. This includes spectral methods~\cite{ma2019graph}, clustering methods such as DiffPool~\cite{ying2018hierarchical} and MinCut~\cite{bianchi2020spectral}, top-K methods~\cite{gao2019graph,lee2019self,zhang2021hierarchical}, and hierarchical graph pooling~\cite{zhang2019hierarchical,huang2019attpool,lee2019self,zhang2021hierarchical,li2020graph,pang2021graph}. Pooling on higher-order networks remains unstudied, with the exception of a general simplicial complex pooling strategy developed by~\cite{cinque2022pooling} along the lines of the proposal made by~\cite{grattarola2022understanding}.



\subsection{Applied algebraic topology}\vspace{-1mm}
Although algebraic topology~\cite{hatcher2005algebraic} is a relatively old field, applications of this field have only recently started to crystallize~\cite{edelsbrunner2010computational,carlsson2009topology}. Indeed, topological constructions have been found to be natural tools for the formulation of longstanding problems in many fields. For instance, persistent homology~\cite{edelsbrunner2010computational} has been successful at finding solutions to various complex data problems~\cite{attene2003shape, bajaj1997contour, boyell1963hybrid, carr2004simplifying, curto2017can, DabaghianMemoliFrank2012, giusti2016two, kweon1994extracting, LeeChungKang2011b, LeeChungKang2011, LeeKangChung2012, LeeKangChung2012b, lum2013extracting, nicolau2011topology, rosen2017using}. Recent years have witnessed increased interest in the role of topology in machine learning and data science~\cite{DW22,hensel2021survey}.
Topology-based machine learning models have been applied in many areas, including topological signatures of data~\cite{biasotti2008describing, carlsson2005persistence, rieck2015persistent}, neuroscience~\cite{curto2017can, DabaghianMemoliFrank2012, giusti2016two, LeeChungKang2011b, LeeChungKang2011, LeeKangChung2012, LeeKangChung2012b}, bioscience~\cite{chan2013topology, dewoskin2010applications, lo2016modeling, nicolau2011topology, taylor2015topological, topaz2015topological}, the study of graphs~\cite{BampasidouGentimis2014, CarstensHoradam2013, ELuYao2012, HorakMaleticRajkovic2009, PetriScolamieroDonato2013, PetriScolamieroDonato2013b,hajij2020efficient,rieck2019persistent}, time series forcasting~\cite{zeng_topological_2021}, Trojan detection~\cite{hu_trigger_2022}, image segmentation~\cite{hu_topology-preserving_2019}, 3D reconstruction~\cite{waibel_capturing_2022},
and time-varying setups~\cite{edelsbrunner2004time, maletic2016persistent, perea2015sw1pers,rieck_uncovering_2020}.

Topological data analysis (TDA)~\cite{edelsbrunner2010computational,carlsson2009topology,DW22,love2020topological,ghrist2014elementary} has emerged as a scientific area that harnesses topological tools
to analyze data and develop machine learning algorithms.
TDA has found many applications in machine learning, including enhancing existing machine learning models~\cite{hofer2017deep,bruel2019topology,wangtopogan,leventhal2023exploring,bentaieb2016topology,clough2019explicit}, improving the explainability of deep learning models~\cite{elhamdadi2021affectivetda,carlsson2020topological,love2021topological}, dimensionality reduction~\cite{moor2020topological}, filtration learning~\cite{hofer2020graph}, and topological layers constructions~\cite{kim2020pllay}. A notable research trend has been the vectorization of persistence diagrams. Vector representations of persistence diagrams are constructed in order to be utilized in downstream machine learning tasks. These methods include Betti curves~\cite{umeda2017time}, persistence landscapes~\cite{bubenik2015statistical}, persistence images~\cite{adams2017persistence}, and other vectorization constructions~\cite{chen2015statistical,berry2020functional,kusano2016persistence}. A unification of these methods has been proposed recently in~\cite{carriere_perslay_2020}.   

Our work introduces combinatorial complexes (CCs) as a generalized higher-order network on which deep learning models can be defined and studied in a unifying manner. Hence, our work expands TDA by formalizing deep learning notions in topological terms and by realizing constructions in TDA, e.g., mapper~\cite{singh2007topological}, in terms of our TDL framework. The construction of CCs and of combinatorial complex neural networks (CCNNs), which are neural networks defined on CCs, is inspired by classical notions in algebraic topology~\cite{hatcher2005algebraic} and in topological quantum field theory~\cite{turaev2016quantum}, and by recent advances in TDA~\cite{carlsson2006algebraic, carlsson2009topology, carlsson2008local, carlsson2008persistent, carlsson2009theory, carlsson2005persistence, collins2004barcode} as applied to machine learning~\cite{pun2018persistent,DW22}.

\vspace{-1mm}
\section{Conclusions}
\vspace{-1mm}
We have established a topological deep learning (TDL) framework that enables the learning of representations for data supported on topological domains. To this end, we have introduced combinatorial complexes (CCs) as a new topological domain to model and characterize the main components of TDL. Our framework provides a unification for many concepts that may be perceived as separate in the current literature. Specifically, we can reduce most deep learning architectures presented thus far in the literature to particular instances of combinatorial complex neural networks (CCNNs), based on computational operations defined on CCs. Our framework thus provides a platform for a more systematic exploration and comparison of the large space of deep learning protocols on topological spaces.

\paragraph{Limitations.} This work has laid out the foundations of a novel TDL framework. While TDL has great potential, similar to other novel learning frameworks, there are limitations, many of which are still not well-understood. Specifically, some known limitations involve:
\begin{itemize}[leftmargin=*,noitemsep,topsep=0.1em]
    \item \textit{Computational complexity}: The primary challenge for moving from graphs to richer topological domains is the combinatorial increase in the complexity to store and process data defined on such domains. Training a TDL network can be a computationally intensive task, requiring careful consideration of neighborhood functions and generalization performance. TDL networks also require a large amount of memory, especially when working with a large number of matrices during network construction. The topology of the network can also increase the computational complexity of training.
    \item \textit{The choice of the neural network architecture}: Choosing the appropriate neural network architecture for a given dataset and a given learning task can be challenging. The performance of a TDL network can be highly dependent on the choice of architecture and its hyperparameters. 

    \item \textit{Interpretability and explainability}: The architecture of a TDL network can make it difficult to interpret the learnt representations and understand how the network is making predictions. 
    \item \textit{Limited availability of datasets}: TDL networks require topological data, which we have found to be of limited availability. Ad hoc conversions of existing data to include higher-order relations may not always be ideal.
\end{itemize}

\paragraph{Future work.} Aforementioned limitations leave ample room for future studies, making the realization of the full potential of TDL an interesting endeavour. While the flexible definition of CCs, as compared to, e.g., simplicial complexes, already provides some mitigation to the associated computational challenges, improving the scaling of CCNNs even further will require the exploration of sparsification techniques, randomization and other algorithmic improvements. Besides addressing the aforementioned limitations, promising directions not treated within this paper include explorations of directed~\cite{ausiello2017directed}, weighted~\cite{battiloro2023topological}, multilayer~\cite{menichetti2016control}, and time-varying dynamic topological domains~\cite{torres2021and,anwar2022synchronization,yin2022dynamic}. There are also several issues related to the selection of the most appropriate topological domain for a given dataset in the first place, which need further exploration in the future. Additionally, there is a need, but also a research opportunity, to better understand CCNN architectures from a theoretical perspective. This could in turn lead to better architectures. To illustrate this point, consider graph neural networks (GNNs) based on message passing~\cite{gilmer2017neural}, which have recently been shown to be as powerful as the Weisfeiler--Lehman isomorphism test\footnote{The Weisfeiler--Lehman isomorphism test~\cite{weisfeiler1968reduction} is a widely-used graph isomorphism algorithm that provides a coloring of a graph's vertices, and this coloring gives a necessary condition for two graphs to be isomorphic.}~\cite{xu2018powerful,maron2019provably, morris2019weisfeiler}. 
This connection between GNNs and classical graph-based combinatorial invariants has driven theoretical developments of the graph isomorphism problem and has inspired new architectures~\cite{xu2018powerful,maron2019provably,arvind2020weisfeiler,bouritsas2020improving}.
We expect that connecting similar developments will also be important for TDL.

The topological viewpoint we adopt brings about many interesting properties. For example, we are able to model other types of \emph{topological inductive biases} in our computational architectures, such as the properties that do not change under different discretizations of the underlying domain, e.g., the Euler characteristic that is commonly used to distinguish topological spaces. 
While isomorphisms are the primary \textit{equivalence relation} in graph theory, \textit{homeomorphisms} and \textit{topological equivalence} are more relevant for data defined on topological spaces, and invariants under homeomorphisms have different machine learning applications\footnote{Intuitively, two topological spaces are equivalent if one of them can be deformed to the other via a continuous transformation.}. 
Homeomorphism equivalence is more relevant in various applications in which domain discretization is an artifact of data processing, and not an intrinsic part of the data~\cite{sharp2022diffusionnet}. 
Further, homeomorphism equivalence translates to a similarity question between two structures. 
Indeed, topological data analysis has been extensively utilized towards addressing the problem of similarity between meshes~\cite{dey2010persistent,hajij2018visual,rieck2019persistent}. 
In geometric data processing, neural network architectures that are agnostic to mesh discretization are often desirable and perform better in practice~\cite{sharp2022diffusionnet}. 
We anticipate that the development of TDL models will open up new avenues to explore topological invariants across topological domains.



\subsubsection*{Acknowledgments}

M.~H. acknowledges support from the National Science Foundation, award DMS-2134231.
G.~Z. is currently affiliated with National Institutes of Health (NIH), but the core of this research was done while being associated with the University of South Florida (USF). This article reports contributions of the authors and does not represent the views of NIH, or the United States Government.
N.~M. acknowledges support from the National Science Foundation, Award DMS-2134241.
T.~B. acknowledges support from the Engineering and Physical Sciences Research Council [grant EP/X011364/1].
T.~K.~D. acknowledges support from the National Science Foundation, Award CCF 2049010.
N.~L. acknowledges support from the Roux Institute and the Harold Alfond Foundation.
R.~W. acknowledges support from the National Science Foundation, Award DMS-2134178. 
P.~R. acknowledges support from the National Science Foundation, Award IIS-2316496.
M.~T.~S. acknowledges funding by the Ministry of Culture and Science (MKW) of the German State of North Rhine-Westphalia (NRW R\"uckkehrprogramm) and the European Union (ERC, HIGH-HOPeS, 101039827). Views and opinions expressed are however those of M.~T.~S. only and do not necessarily reflect those of the European Union or the European Research Council Executive Agency; neither the European Union nor the granting authority can be held responsible for them.

The authors would like to thank Mathilde Papillon and Sophia Sanborn for helping improve Figure~\ref{fig:HON} and for the insightful discussions on the development of tensor diagrams.

\printbibliography[heading=bibintoc, title={References}] 


\newpage

\appendix

\section{Glossary}
\label{appendix_glossary}

Tables~\ref{tab:gloss} and~\ref{tab:term} summarize the paper's notations and acronyms, respectively.
\vspace{3mm}

\small
\begin{longtable}{|l l|}
\caption{Tabulation of the notations used in this paper.\vspace{-1mm}}
\label{tab:gloss}\\
\hline
\multicolumn{1}{|l}{\cellcolor{gray!35}\textbf{Notation}} &
\multicolumn{1}{l|}{\cellcolor{gray!35}\textbf{Description}} \\
\hline
\multicolumn{2}{|c|}{\cellcolor{gray!10} Set notations}  \\
\hline
			$S$ & Non-empty finite set of abstract entities \\
           $P_{S}$ &  Index set \\
			$\mathcal{P}(S)$ & Power set of a set $S$\\
            $(S,\mathcal{N})$ & Topological space of a nonempty set $S$ and a neighborhood topology $\mathcal{N}$ \\
			$\mathcal{N}_{a}(x)$ & Adjacency set of a cell $x$ \\
			$\mathcal{N}_{co}(x)$ & Coadjacency set of a cell $x$ \\
			$\mathcal{N}_{\{ G_1,...,G_n \}  }(x)$ & Neighbors of $x$ 
			specified by the neighborhood matrices $\{ G_1,\dots,G_n \}$\\
			$\mathcal{N}_{\searrow}(x)$ &  Set of down-incidence of a cell $x$  \\
			$\mathcal{N}_{\nearrow}(x)$ & Set of up-incidence of a cell $x$\\ 	
			$\mathcal{N}_{\searrow,k}(x)$ & Set of $k$-down incidence of a cell $x$  \\
			$\mathcal{N}_{\nearrow,k}(x)$ & Set of $k$-up incidence of a cell $x$\\ 				
			$\N$ and $\Znon$ & Set of positive integers and non-negative integers, respectively \\

			\hline
		\multicolumn{2}{|c|}{\cellcolor{gray!10}Domains}  \\
			\hline
			$\mathcal{G}$ &  Graph \\
			$x^k$ & Cell $x$ of rank $k$ \\
            $\rk$ & Rank function \\
            $(S, \CCX, \rk)$ & CC, consisting of a set $S$, a subset $\CCX$ of $\mathcal{P}(S)\setminus\{\emptyset\}$, and a rank function $\rk$ \\ 
			$\dim (\mathcal{X})$ & Dimension of a CC $\mathcal{X}$ \\
			$\{c_\alpha\}_{\alpha \in I }$ &  Partition into subspaces (cells) indexed by an index set $I$ \\
			$\Int(x)$  &  Interior of a cell $x$ in a regular cell complex \\
			$n_\alpha\in \mathbb{N}$ & Dimension of a cell in a regular cell complex \\
			$0$-cells & Vertices of a CC \\
			$1$-cells & Edges of a CC \\
			$k$-\textit{cells} & Cells with rank $k$ \\ 
            $\mathcal{X}^{(k)}$ & $k$-skeleton of $\mathcal{X}$,
            formed by $i$-cells in $\mathcal{X}$ with $i\leq k$\\
			$\mathcal{X}^k$ & Set of k-cells of $\mathcal{X}$\\
			$|\mathcal{X}^k|$ & Cardinality of $\mathcal{X}^k$,
   that is number of $k$-cells of $\mathcal{X}$ \\
			$\CC_{n-hop}(G)$ & $n$-hop CC of a graph $G$ \\
			$\CC_p(G)$ & Path-based CC of a graph $G$ \\ 
			$\CC_{loop}(G)$ & Loop-based CC of a graph $G$  \\
			$\CC_{SC}(\mathcal{Y})$ & Coface CC of a simplicial complex/CC $\mathcal{Y}$  \\
   			\hline
			\multicolumn{2}{|c|}{\cellcolor{gray!10}Matrix notations}  \\
			\hline
			$B_{r,k}$ & Incidence matrices between $r$-cells and $k$-cells \\ 
   	$A_{r,k}$ & Adjacency matrices among the cells of $\CCX^{r}$
with respect to the cells of $\CCX^{k}$ \\
$coA_{r,k}$ & Coadjacency matrices among the cells of $\CCX^{r}$
with respect to the cells of $\CCX^{k}$ \\
			\hline
			\multicolumn{2}{|c|}{\cellcolor{gray!10}CCNNs}  \\
			\hline
			$\mathbf{W}$  & Trainable parameter \\
			$\mathcal{C}^k(\mathcal{X},\mathbb{R}^d)$ & $k$-cochain space with features in $\mathbb{R}^d$ \\
			$\mathcal{C}^k$ &  $k$-cochain space with features in some Euclidean space \\  
			$\mathbf{G}= \{ G_1,\ldots ,G_m \} $ & Set of cochain maps $G_i$ defined on defined on a complex \\
   			$\mathcal{M}_{ \mathbf{G};\mathbf{W}}$ & Merge node \\
			$G:C^{s}(\mathcal{X})\to C^{t}(\mathcal{X})$ & Cochain map  \\
			$(\mathbf{x}_{i_1},\ldots, \mathbf{x}_{i_m})$ & Vector of cochains \\
			$att^{l}: C^{s}(\mathcal{X})\to C^{s}(\mathcal{X}) $ &  Higher-order attention matrix  \\
			$\mathcal{N}_{\mathcal{Y}_0}=\{ \mathcal{Y}_1,\ldots, \sigma_{|\mathcal{N}_{\mathcal{Y}_0}|} \}$ & Set of a complex object in the vicinity of  $\mathcal{Y}_0$  \\ 
			$a: {\mathcal{Y}_0}\times \mathcal{N}_{\mathcal{Y}_0}\to [0,1] $ &  Higher-order attention function \\
			$\CCN_{\mathbf{G};\mathbf{W}}$ & CCNN or its tensor diagram representation\\
			$\mathcal{H}_{\mathcal{X}}= (V (\mathcal{H}_{\mathcal{X}}), E(\mathcal{H}_{\mathcal{X}}) )$ & Hasse graph with vertices
			$V (\mathcal{H}_{\mathcal{X}})$
                and edges
                $E(\mathcal{H}_{\mathcal{X}})$;
                see Definition~\ref{HG}
			 \\ 
			\hline
\end{longtable}
\normalsize

\small
\begin{longtable}{|l l|}
\caption{Tabulation of the acronyms used in this paper.}
\label{tab:term}\\
\hline
\multicolumn{1}{|l}{\cellcolor{gray!35}\textbf{Acronym}} &
\multicolumn{1}{l|}{\cellcolor{gray!35}\textbf{Description}} \\
\hline
AGD & Average geodesic distance \\ 
CC & Combinatorial complex \\
CCANN & Combinatorial complex attention neural network \\
CCCNN & Combinatorial complex convolutional neural network \\ 
CCNN & Combinatorial complex neural network \\
CNN & Convolutional neural network \\
DEC & Discrete exterior calculus \\
GDL & Geometric deep learning \\ 
GNN & Graph neural network \\
MOG & Mapper on graphs \\ 
RNN & Recurrent neural network \\
SCoNe & Simplicial complex network \\
sub-CC & sub-combinatorial complex \\
TDA & Topological data analysis \\
TDL & Topological deep learning \\
TQFT & Topological quantum field theory \\
\hline
\end{longtable}
\normalsize





\section{Lifting maps}
\label{appdx:liftings}

Lifting refers to the process of mapping a featured domain to another featured domain via a well-defined procedure. This section shows how we can lift a given domain to a CC or cell complex.
Such lifting is useful as it allows CCNNs to be applied
to common topological domains, including graphs and cell/simplicial complexes. This section only scratches the surface, as there remain many lifting constructions to be explored. We refer the reader to~\cite{ferri2018simplicial} for examples of lifting graphs to simplicial complexes.

\subsection{n-hop CC of a graph}
\label{example1}

\begin{figure}[!t]
\begin{center}
\includegraphics[scale = 0.13, keepaspectratio = 0.20]{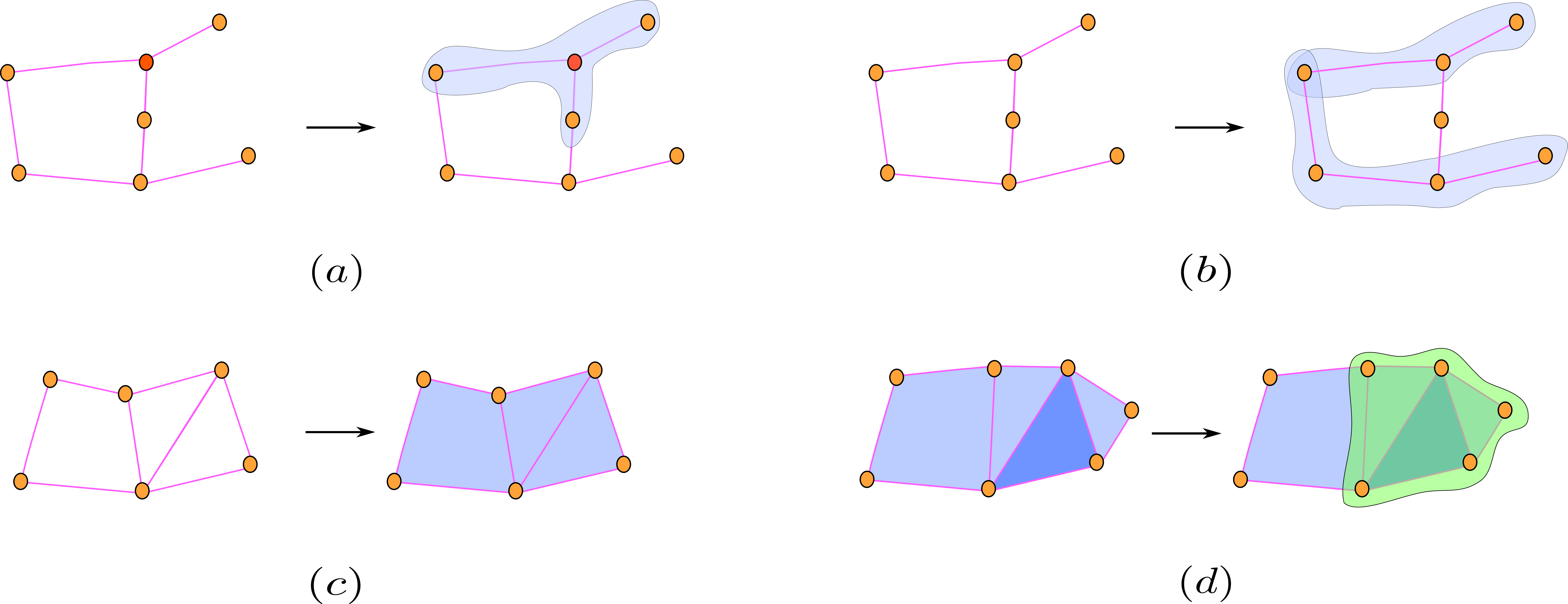}
\end{center}
\caption{Examples of lifting domains to CCs and cell complexes.
(a): The $1$-hop neighborhood of the red node can be considered as a 2-cell that we can augment to the graph. Adding such 2-cells to a graph yields a CC called the 1-hop neighborhood of the graph (see Section~\ref{example1}).
(b): A path on a graph of length more than two can be considered as a 2-cell that we can augment to the graph. Adding such 2-cells to a graph yields a CC called a path-based CC of the graph (see Section~\ref{example2}).
(c): A loop in a graph (i.e., a closed path with no repeating edges) can be considered as a 2-cell that we can augment to the graph. Adding such 2-cells to a graph yields a CC called a loop-based CC of the graph (see Section~\ref{example3}). 
(d): For every blue 2-cell of a simplicial complex,
we introduce a green 3-cell obtained by considering the 1-coface of the 2-cell.
Adding such 3-cells to a simplicial complex yields a CC of dimension three called
the coface CC of the simplicial complex (see Section~\ref{example4}).}
\label{fig:liftings}
\end{figure}

Let $\mathcal{G}=(V(\mathcal{G}),E(\mathcal{G}))$ be a graph and $n\geq 2$ an integer.
The $n$\textit{-hop CC} of $\mathcal{G}$, denoted by $\CCHOP{n}(\mathcal{G})$,
is the CC whose $0$-cells, $1$-cells, and $n$-cells are
the nodes of $\mathcal{G}$,
edges of $\mathcal{G}$, and set nodes in $n$-hop neighborhoods of the nodes in $\mathcal{G}$, 
respectively. It is easy to verify that $\CCHOP{n}(\mathcal{G})$ is a CC of dimension $n$.
Figure~\ref{fig:liftings}(a) visualizes the $1$-hop CC of a graph.

\subsection{Path-based and subgraph-based CC of a graph}
\label{example2}

Let $\mathcal{G}=(V(\mathcal{G}),E(\mathcal{G}))$ be a graph.
A natural CC structure on $\mathcal{G}$ considers paths of $\mathcal{G}$.
We define a \textit{path-based CC} of $\mathcal{G}$,
denoted by $\CC_p(\mathcal{G})$,
to be a CC consisting of $0$-cells, $1$-cells and $2$-cells
specified as follows.
First, $\CCX^0$ and $\CCX^1$ in $\CC_p(\mathcal{G})$
are the sets of nodes and edges of $\mathcal{G}$, respectively.
We now explain how to construct a $2$-cell in $\CC_p(\mathcal{G})$.
Let $P$ be a path in $\mathcal{G}$ with length larger than or equal to two
(i.e., with two or more edges). An element $x_P$ in $\CCX^2$
induced by $P$ is defined to be $x_P=\cup_{v\in P }\{v\} $.
The set $\CCX^2$ in $\CC_p(\mathcal{G})$ is a non-empty collection of elements $x_P$.
It is easy to verify that $\CC_p(\mathcal{G})$ is a CC with $\dim(\CC_p(\mathcal{G}))=2$.
Note that we may replace the path $P$ by a tree/subgraph
of graph $\mathcal{G}$ and obtain a similar CC structure
induced by the tree/subgraph of $\mathcal{G}$.
Figure~\ref{fig:liftings}(b) shows an example of a path-based CC of a graph.


\subsection{Loop-based CC of a graph}
\label{example3}

Let $\mathcal{G}=(V(\mathcal{G}),E(\mathcal{G}))$ be a graph. We associate a CC structure with $\mathcal{G}$ that considers loops in $\mathcal{G}$.
We define a \textit{loop-based CC} of $\mathcal{G}$, denoted by $\CC_{loop}(\mathcal{G})$,
to be a CC consisting of $0$-cells, $1$-cells and $2$-cells
specified as follows.
First, we set $\CCX^0$ and $\CCX^1$ in $\CC_{loop}(\mathcal{G})$ to be the nodes and edges of $\mathcal{G}$, respectively.
We now explain how to construct a $2$-cell in $\CC_{loop}(\mathcal{G})$.
A 2-cell in $\CC_{loop}(\mathcal{G})$
is a set $C=\{x^0_1, \ldots , x^0_k\} \subset \CCX^0$
such that $\{x^0_i,x^0_{i+1}\}$, $1 \leq  i \leq k - 1$, and $\{x^0_k, x^0_1\}$ are the only edges in $\CCX^1 \cap C$.  The set $\CCX^2$ in $\CC_{loop}(\mathcal{G})$ is a nonempty collection of elements $C$. It is easy to verify that $\CC_{loop}(\mathcal{G})$ is a CC with $\dim(\CC_{loop}(\mathcal{G}))=2$. Note that the sequence $(x^0_1, \ldots , x^0_k)$ defines a loop in $\mathcal{G}$. This loop is called the loop that characterizes the 2-cell $C=\{x^0_1, \ldots , x^0_k\}$.
Similar constructions are suggested in~\cite{aschbacher1996combinatorial,basak2010combinatorial,savoy2022combinatorial,roddenberry2021signal}.
In fact, it is easy to confirm that every 2-dimensional regular cell complex can be constructed in this manner~\cite{roddenberry2021signal}.
Figure~\ref{fig:liftings}(c) shows an example of a loop-based CC of a graph.

\subsection{Coface CC of a simplicial complex/CC}
\label{example4}

Here, we describe a method to lift a simplicial complex of dimension two to a CC of dimension three. This method can be easily generalized to other dimensions. For a simplicial complex $\CCY$ of dimension two, the \textit{coface CC} of $\CCY$, denoted by $\CC_{SC}( \CCY)$, is defined as follows. $\CCY^0$, $\CCY^1$, and $\CCY^2$ in $\CC_{SC}( \CCY)$ are the nodes, the edges, and the triangles in $\CCY$, respectively. We now explain how to construct a $3$-cell in $\CC_{SC}( \CCY)$. Let $x^2$ be a 2-cell in $\CCY$. The 3-cell in $\CC_{SC}( \CCY)$ associated with $x^2$ is the union of all 0-cells in $\mathcal{N}_{co,1}(x^2) \cup x^2$. The set $\CCY^3$ in $\CC_{SC}( \CCY)$ is defined as the set of all 
3-cells associated with all 2-cells $x^2$ in $\CCY$. It is easy to verify that $\CC_{SC}(\CCY)$ is a CC with $\dim(\CC_{SC}(\CCY) )=3$. A similar lifting construction can be defined to augment any CC of dimension $n$ with $(n+1)$-cells in order to obtain a CC of dimension $n+1$.

\subsection{Augmentation of CCs by higher-rank cells}

The lifting methods proposed in Sections~\ref{example2},~\ref{example3}, 
and~\ref{example4} can be described abstractly under a single general lifting construction. Specifically, the essence of all these lifting methods is to \textit{augment the underlying CC $\CCX$ with new cells that have a rank of $\dim(\CCX)+1$}.
Proposition~\ref{augmented_CC} formalizes the general lifting construction.

\begin{proposition}
\label{augmented_CC}
Let $S$ be a nonempty set and $(S,\CCX,\rk)$ a CC of dimension $n$ defined on $S$. Consider a set $\CCX^{n+1} \subset \mathcal{P}(S)$
such that if $x\in\CCX$ and $y \in \CCX^{n+1}$ with
$x\subseteq y $, then $x\subsetneq y$.
Further, consider a map
$\hat{\rk}\colon \CCX\cup \CCX^{n+1}\to \Znon$ that satisfies
$\hat{\rk}(x)= \rk(x)$ for all $x\in\CCX$ and $\hat{\rk}(x)=n+1$ for all $x \in \CCX^{n+1} $.
For $\CCX^{n+1}$ and $\hat{\rk}$ satisfying such conditions,
$(S,\CCX\cup \CCX^{n+1},\hat{\rk} )$ is a CC of dimension $n+1$.
\end{proposition}

\begin{proof}
The proof follows directly from Definition~\ref{def:cc}.
\end{proof}

Given a CC $\CCX$, we call a CC of the form $(S,\CCX\cup \CCX^{n+1},\hat{\rk} )$, as constructed in Proposition~\ref{augmented_CC}, a \textit{highest-rank augmented CC of $\CCX$}. Note that Proposition~\ref{augmented_CC} provides a constructive and iterative method to build a CC of arbitrary dimension 
from a nonempty set $S$ of abstract points. 



\section{CCNN architecture search and topological quantum field theories}
\label{tqft}

The problem of CCNN architecture search for a given TDL task can be cast as a hyperparameter optimization problem over the space of CCNNs. More precisely, consider the query of searching for an optimal $\CCN$ between two fixed Cartesian products
$ \mathcal{C}^{i_1}\times\mathcal{C}^{i_2}\times \cdots \times  \mathcal{C}^{i_m}$
and $ \mathcal{C}^{j_1}\times\mathcal{C}^{j_2}\times \cdots \times \mathcal{C}^{j_n}$
of cochain spaces.
In practice, this query poses a challenging problem.
Rather than performing a computationally expensive CCNN search directly,
an alternative approach is to conduct a search in the simpler space of
\textit{marked trivalent graphs} between marked points $\{i_1,\ldots,i_m\}$ and  $\{j_1,\ldots,j_n\}$,
and then map the resulting marked trivalent graph to a corresponding CCNN architecture.
In the present section, we briefly sketch such a graph-based search method
using tools from \textit{topological quantum field theory (TQFT)}~\cite{turaev2016quantum},
and accordingly we assume some familiarity with the basics of \textit{category theory}.

\begin{figure}[!t]
\begin{center}
\includegraphics[scale = 0.17, keepaspectratio = 0.10]{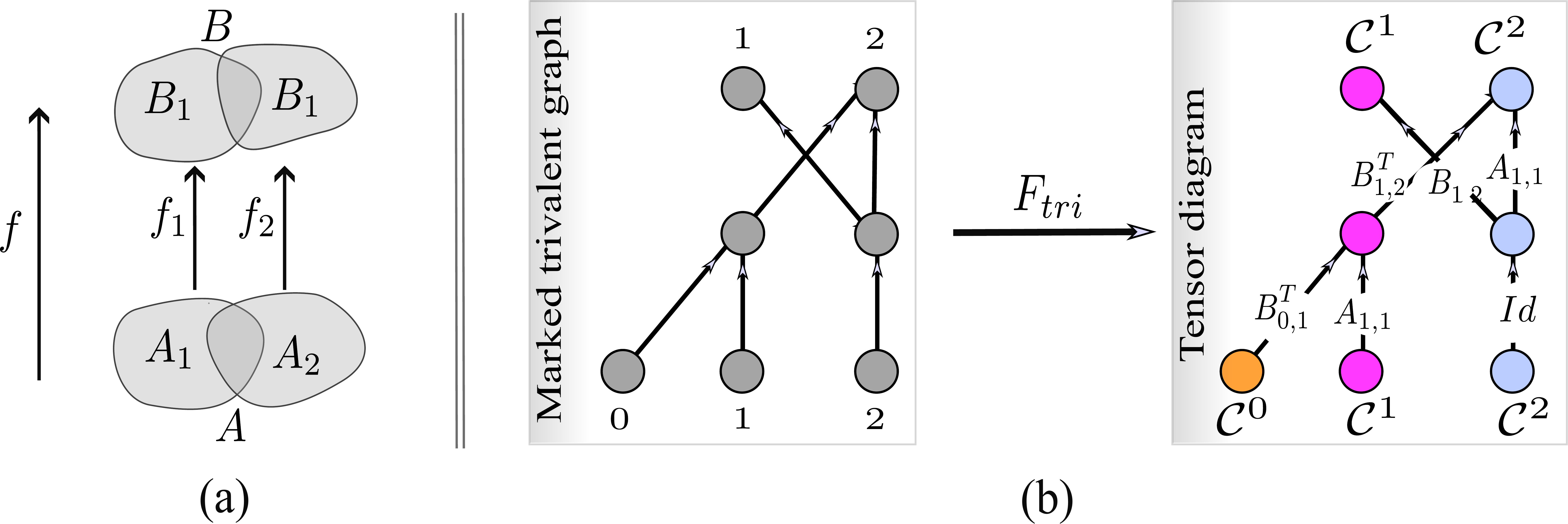}
\end{center}
\caption{A sketch of the main idea of a TQFT construction in (a)
and a functor $F_{tri}$ for constructing tensor diagrams in (b).
(a): In the depicted example, the goal is to construct a map $f\colon A\to B$ between two topological spaces $A$ and $B$. We first decompose $A$ and $B$ into simpler sub-spaces, say $A = A_1 \cup A_2$ and $B = B_1 \cup B_2$, so that ${A_1,A_2}$ and ${B_1,B_2}$ are `more elementary spaces' than $A$ and $B$, respectively. We then construct two maps $f_1 \colon A_1 \to B_1$ and $f_2 \colon A_2 \to B_2$, and use them to construct $f$.
(b): A visual example of a functor $F_{tri}$ that pairs each marked trivalent graph with a corresponding tensor diagram. This example sends a marked trivalent graph with marked points $\{0,1,2\}$ at the bottom and marked points $\{1,2\}$ at the top to a tensor diagram with domain $\mathcal{C}^{0} \times \mathcal{C}^{1}\times \mathcal{C}^{2}$ and codomain $\mathcal{C}^{1}\times \mathcal{C}^{2}$. Once the push-forward, the merge and the split nodes are defined, the functor $F_{tri}$ attaches a well-defined tensor diagram to each trivalent graph. A CCNN can be then constructed from the tensor diagram.}
\label{functor}
\end{figure}

Tensor diagrams draw their inspiration from TQFT, in which arbitrary maps between topological spaces are constructed from simpler and more manageable building blocks. The main workflow in TQFT constructions involves breaking down the topological spaces under consideration into simpler subspaces, and subsequently utilizing maps between the subspaces to construct maps between the initial topological spaces; see Figure~\ref{functor}(a).
Thus, the topological properties of maps between topological spaces are better understood
via the topological properties of simpler constituent maps between respective subspaces.
This can be especially useful in applications such as knot theory or the study of three-dimensional manifolds, where understanding the topological properties of involved maps is the main interest~\cite{turaev2016quantum}.  

Before we introduce the relation between tensor diagrams and TQFTs,
we sketch two required preliminaries,
namely marked trivalent graphs and TQFTs. First, we define the objects and morphisms of the category of \textit{marked trivalent graphs} $Tri$. A marked trivalent graph is intuitively described via its layout. Consider the 2d-disk $[0,1]\times [0,1]$ with a collection of marked points $\{i_1,\ldots,i_m\}$ at the bottom and a collection of marked points $\{j_1,\ldots,j_n\}$ at the top. A marked trivalent graph is a graph that is drawn inside the disk in such a way that all edges flow within the disk, connecting the marked points at the top with the marked points at the bottom, and meeting at each vertex with exactly three edges. In the category $Tri$, a trivalent graph with marked points $\{i_1,\ldots,i_m\}$ and $\{j_1,\ldots,j_n\}$ at the bottom and top, respectively, plays the role of a morphism between the object $\{i_1,\ldots,i_m\}$ and the object $\{j_1,\ldots,j_n\}$. The composition of two such morphisms, when admissible, is defined to be the vertical concatenation of their corresponding marked trivalent graphs. The vertical concatenation of marked trivalent graphs yields a marked trivalent graph. 
Finally, any trivalent graph can be built by concatenating horizontally and vertically three elementary trivalent graphs
similar to the graph shown in Figure~\ref{functor}(b), ignoring the labels in the figure.

 
At a high level, a TQFT is a functor $F\colon Bord_{n}\to Vec_{K}$ that assigns a $K$-vector space $F(x)\in Vec_{K}$ to each $n$-manifold $x \in Bord_{n}$, and a linear map $F(f)$ to each $(n+1)$-cobordism $f \in Bord_{n}$. It is noted that a $(n+1)$-cobordism is an $(n+1)$-manifold representing a morphism between two $n$-manifolds. The functor $F$ is typically defined on a few elementary morphisms in $Bord_{n}$, but can be extended to arbitrary morphisms in $Bord_{n}$ by preserving the structure of the elementary maps when mapped via $F$. Such a functor enables the study of $n$-manifolds and $(n+1)$-cobordisms defined between them by mapping $n$-manifolds to their corresponding simpler vector spaces in $Vec_{K}$ and by mapping $(n+1)$-cobordisms to linear maps between the corresponding vector spaces. 

To establish the relation between marked trivalent graphs and tensor diagrams, we introduce a new TQFT $F_{tri}$ that sends
each morphism (marked trivalent graph) in $Tri$
to a corresponding tensor diagram
by labeling marked trivalent graphs with appropriate cochain maps. Figure~\ref{functor}(b) shows an illustration of such a functor $F_{tri}$.
Through the lens of $F_{tri}$,
the relation between tensor diagrams and marked trivalent graphs becomes clear;
a tensor diagram is a marked trivalent graph whose edges are labeled via cochain maps.
The functor $F_{tri}$ allows us to search the space of marked trivalent graphs
as an equivalent way of conducting CCNN architecture search,
since CCNNs are represented by tensor diagrams.




\section{Learning discrete exterior calculus operators with CCANNs}
\label{linear}

The operator $G_{tr}=G\odot att $
of Equations~\ref{attention1} and~\ref{attention2} has an advantageous cross-cutting interpretation. First, recall that $G_{tr}$ has the same shape as the original operator $G$. More importantly, $G_{tr}$ can be viewed as a learnt version of $G$. For instance, if $G$ is the $k$-Hodge Laplacian $\mathbf{L}_k$, then the learnt attention version $G_{tr}$ of it represents a $k$-Hodge Laplacian that is adapted to the domain $\CCX$ for the learning task at hand. This perspective converts our attention framework to a tool for learning \textit{discrete exterior calculus (DEC) operators}~\cite{desbrun2008discrete}. We refer the interested reader to recent works along these lines~\cite{smirnov2021hodgenet,trask2022enforcing}, where neural networks are used to learn Laplacian operators in various shape analysis tasks.

Concretely, one of the main building blocks of DEC is a collection of linear operators of the form
$\mathcal{A} \colon \mathcal{C}^i(\CCX) \to \mathcal{C}^j(\CCX)$
that act on a cochain $\mathbf{H}$ to produce another cochain $\mathcal{A}(\mathbf{H})$. An example of an operator $\mathcal{A}$ is the graph Laplacian. There are seven primitive DEC operators, including the discrete exterior derivative, the hodge star and the wedge product. These seven primitive operators can be combined together to form other operators. In our setting, the discrete exterior derivatives are precisely a signed version of the incidence matrices defined in the context of cell/simplicial complexes. We denote the $k$-signed incidence matrix defined on a cell/simplicial complex by $\mathbf{B}_k$. It is common in the context of discrete exterior calculus~\cite{desbrun2008discrete}
to refer to $\mathbf{B}_k^T$ as the $k^{th}$ \textit{discrete exterior derivative} $d^k$.
So, from a DEC point of view, the matrices $\mathbf{B}_0^T, \mathbf{B}_1^T$ and $\mathbf{B}_2^T$
are regarded as the discrete exterior derivatives
$d^0(\mathbf{H})$, $d^1 (\mathbf{H})$, and $d^2 (\mathbf{H})$ of some 0-, 1-, and 2-cochains defined on $\mathcal{\CCX}$,
which in turn are the discrete analogs of the gradient $\nabla \mathbf{H}$, curl $ \nabla\times \mathbf{H} $ and divergence $ \nabla \cdot \mathbf{H} $ of a smooth function defined on a smooth surface. We refer the reader to~\cite{desbrun2008discrete} for a coherent list of DEC operators and their interpretation.
Together, cochains and the operators that act on them provide a concrete framework that facilitates
computing a cochain of interest, such as a cochain obtained by solving a partial differential
equation on a discrete surface.

Our attention framework can be viewed as a non-linear version of the DEC based on linear operators $\mathcal{A}$, and can be used to learn the DEC operators on a domain $\CCX$ for a particular learning task.
Specifically, a linear operator $\mathcal{A}$, as it appears in classical DEC,
can be considered as a special case of Equations~\ref{attention1} and~\ref{attention2}. Unlike existing work~\cite{smirnov2021hodgenet,trask2022enforcing}, our DEC learning approach based on CCANNs generalizes and applies to all domains in which DEC is typically applicable; examples of such domains include triangular and polygonal meshes~\cite{crane2013digital}. In contrast, existing operator learning methods are defined only for particular types of DEC operators, and therefore cannot be used to learn arbitrary types of DEC operators.

\section{A mapper-induced topology-preserving CC-pooling operation}
\label{mog_section}

In this section, we give an example of constructing a shape-preserving pooling operation on a CC $\CCX$. Specifically, we demonstrate the case in which $\CCX$ is a graph, and utilize the \textit{mapper on graphs (MOG)} construction~\cite{hajij2018mog}, which is a graph skeletonization algorithm that can be used to augment $\CCX$ with topology-preserving higher-rank cells, as demonstrated in Figure~\ref{fig:hs}. Although we only demonstrate the shape-preserving pooling construction on graphs, the method suggested herein can be easily extended to CCs.

Let $\CCX$ be a connected graph and $g\colon\CCX^{0} \to [0,1]$ a scalar function. Let $\mathcal{U}=\{U_\alpha\}_{\alpha \in I}$ be a finite collection of open sets that covers the interval $[0,1]$. The MOG construction of a graph $MOG(V_{MOG},E_{MOG})$
based on the triplet ($\CCX$, $g$, $\mathcal{U}$) consists of the following steps: 
\begin{enumerate}
    \item We first use the cover $\mathcal{U}$ to construct the pull-back cover $g^{\star}(\mathcal{U})=\{g^{-1}(U_{\alpha})\}_{\alpha \in I}$.
    \item The vertex set $V_{MOG}$ is formed by considering the connected components (i.e., maximal connected subgraphs) induced by $g^{-1}(U_\alpha)$ for each $\alpha$. 
    \item The edge set $E_{MOG}$ is formed by considering the intersection among the connected components computed in step 2.
\end{enumerate}

Figure~\ref{mog} shows an illustrative example of
applying the MOG algorithm to a graph.
Figure~\ref{mog}(a) shows the graph on which the MOG algorithm is applied,
while Figure~\ref{mog}(b) visualizes the scalar function $g$ and
the covering $\mathcal{U}$,
which consists of four covering elements
depicted in red, orange, yellow and blue colors.

\begin{figure}[!t]
\begin{center}
\includegraphics[scale = 0.035, keepaspectratio = 0.20]{graph_tda_1.pdf}
\end{center}
\caption{An illustration of the MOG algorithm. The input to the MOG algorithm
is a triplet $(\CCX$, $g$, $\mathcal{U})$, where $\CCX$ is a graph,
$g\colon\CCX^0\to [0,1]$ is a scalar function defined on the vertex set of $\CCX$,
and  $\mathcal{U}$ is a cover of $[0,1]$.
(a): An input graph $\CCX$.
(b): The scalar function $g\colon\CCX\to [0,1]$ is visualized
by color-mapping its scalar values according to the displayed color bar.
Figure (b) also shows the covering $\mathcal{U}= \{U_1,U_2, U_3, U_4\}$ depicted in red, orange, yellow and blue colors.
(c): We pull back via $g$ each cover element $U_i$ in $\mathcal{U}$, and we compute the connected components in $g^{-1}(U_i)$.
(d): The vertex set in the graph generated by the MOG algorithm
consists of the connected components induced by $g^{-1}(U_i)$,
while the edge set is formed by considering the intersection
among the connected components.
Note that the graph generated by the MOG algorithm
approximates the shape of the input graph.}
\label{mog}
\end{figure}

The connected components obtained from the MOG algorithm can be used to augment a graph $\CCX$ with a 
skeleton $\CCX^2$ of dimension 2, thus constructing a CC, as described in Proposition~\ref{augmented_CC}.
We denote the resulting CC by $\CCX_{g,\mathcal{U}}$. 
Figure~\ref{mapper_as_pooling_operator} demonstrates 
the construction of a CC using
this augmentation process
based on the MOG algorithm. 

\begin{figure}[!t]
\begin{center}
\includegraphics[scale = 0.035, keepaspectratio = 0.20]{augmented_mapper_CC.pdf}
\end{center}
\caption{A visual example of obtaining a CC from a graph via the MOG algorithm.
(a): An input graph $\CCX$.
(b): The graph $\CCX$ is augmented by the $2$-cells formed via
the connected components obtained from applying the MOG algorithm
to $\CCX$, as described in Figure~\ref{mog123}.
(c): The CC $X_{g,\mathcal{U}}$ obtained by augmenting $\CCX$ with these $2$-cells.}
\label{mapper_as_pooling_operator}
\end{figure}

\begin{figure}[!t]
\begin{center}
\includegraphics[scale = 0.35, keepaspectratio = 0.20]{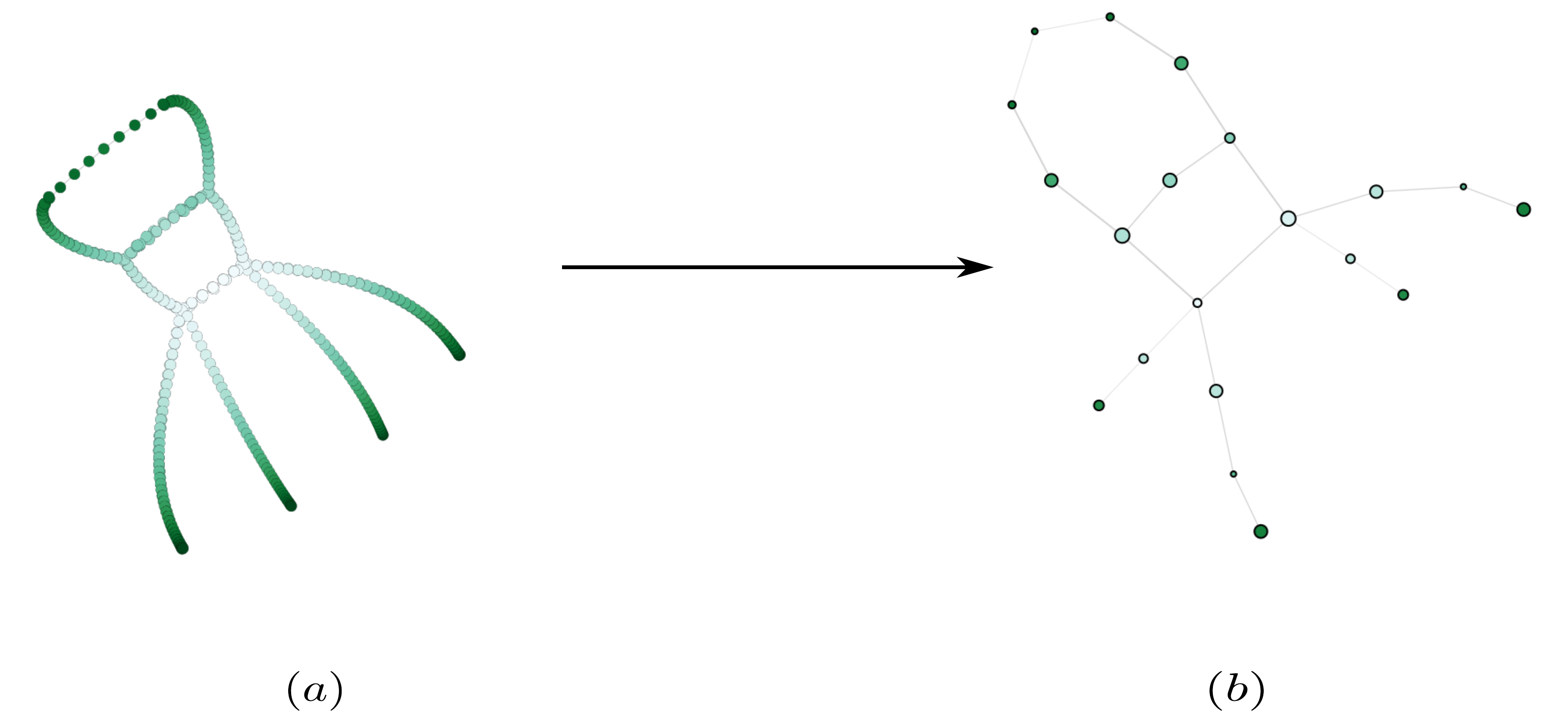}
\end{center}
\caption{Illustration of coarsening a graph via the MOG algorithm.
(a): Visualization of a graph and of a scalar function defined on it,
which are used as inputs to the MOG algorithm.
The scalar function is visualized by color-mapping its scalar values.
(b): The resulting MOG graph.
Notice how the MOG construction preserves the overall shape of the original graph.}
\label{read_data}
\end{figure}

MOG is a topology-preserving graph skeletonization algorithm,
which can be used to coarsen the size of an input graph $\CCX$.
Such coarsening typically occurs when constructing a pooling layer.
So, the skeleton $\CCX^2$ generated by the MOG algorithm
can be utilized to obtain a feature-preserving pooling operation.
Figure~\ref{read_data} illustrates that the MOG algorithm
coarsens an input graph
to produce a graph that preserves topological features of the input graph.

\begin{figure}[!t]
\begin{center}
\includegraphics[scale = 0.035, keepaspectratio = 0.20]{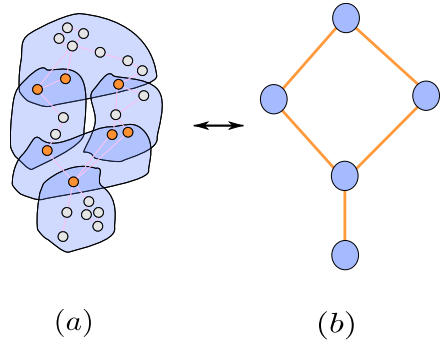}
\end{center}
\caption{Visual demonstration of Proposition~\ref{connection_to_tda}.
(a): Visualization of a CC $\CCX_{g,\mathcal{U}}$,
which highlights the intersection between blue 2-cells.
(b): The corresponding $MOG(\CCX, g, \mathcal{U})$ graph.
There is a one-to-one correspondence between the blue 2-cells of $\CCX_{g,\mathcal{U}}$ on the left-hand side
and the blue vertices of $MOG(\CCX, g, \mathcal{U})$ on the right-hand side. Further, two blue 2-cells of 
$\CCX_{g,\mathcal{U}}$ (left) intersect on an orange vertex if and only if there is an orange edge (right)
that connects the corresponding two vertices of $MOG(\CCX$, $g$, $\mathcal{U})$.}
\label{adjadj}
\end{figure}

To recap, the MOG algorithm receives an input graph,
and outputs a coarsened graph along with connected components.
The connected components yield in turn a CC.
The coarsened graph and the adjacency relation between the
$2$-cells of the associated CC
are related, as elaborated in Proposition~\ref{connection_to_tda}.
Figure~\ref{adjadj} conveys visually Proposition~\ref{connection_to_tda}.

\begin{proposition}
\label{connection_to_tda}
Let $(\CCX$, $g$, $\mathcal{U})$ be a triplet
consisting of a graph $\CCX$,
a scalar function $g\colon\CCX^0\to [0,1]$ defined on the vertex set of $\CCX$,
and a cover $\mathcal{U}$ of $[0,1]$.
Let $MOG(\CCX$, $g$, $\mathcal{U})$ be the graph generated by the MOG algorithm
upon receiving the triplet $(\CCX$, $g$, $\mathcal{U})$ as input.
Furthermore, let $\CCX_{g,\mathcal{U}}$ be the CC constructed
from the connected components that are generated by the MOG algorithm.
The adjacency matrix of the graph $MOG(\CCX$, $g$, $\mathcal{U})$
is equivalent to the adjacency matrix $A_{2,2}$ of the CC $\CCX_{g,\mathcal{U}}$.
\end{proposition}

\begin{proof}
The proof follows by observing that the $2$-cells,
which are augmented to $\CCX$ to generate the CC $\CCX_{g,\mathcal{U}}$,
are 2-adjacent if and only if they intersect on a vertex.
Moreover, two cells intersect on a vertex if and only if there is an edge between them
in the graph $MOG(\CCX$, $g$, $\mathcal{U})$ outputted by the MOG algorithm.
\end{proof}

Given a CC $X_{g,\mathcal{U}}$
obtained from a graph $\CCX$
via the MOG algorithm, the map $B_{0,2}^T\colon\mathcal{C}^0(X_{g,\mathcal{U}})\to \mathcal{C}^2(X_{g,\mathcal{U}})$ can be used to induce a shape-preserving CC-pooling operation
(Definition~\ref{pooling_exact_definition}).
More specifically, a signal $\mathbf{H}_0$ supported on the vertices of $\CCX$ can be push-forwarded and pooled to a signal $\mathbf{H}_2$ supported on the 2-cells of $X_{g,\mathcal{U}}$.

\end{document}